%% file: main.tex
\documentclass{egpubl}
\usepackage{egsgp2025}

\usepackage{dfadobe}  

\usepackage[utf8]{inputenc}

\usepackage[backend=biber,backend=biber,bibstyle=EG,citestyle=alphabetic,backref=true]{biblatex} 
\usepackage[T1]{fontenc}  

\usepackage{egweblnk} 
\usepackage{url}            
\usepackage{booktabs}      
\usepackage{tabularx}
\usepackage{amsfonts}      
\usepackage{nicefrac}       
\usepackage{microtype}     
\usepackage{xcolor}         
\usepackage{graphicx}
\usepackage{adjustbox}
\usepackage{comment}
\usepackage{amsmath}
\usepackage{rotating}
\usepackage{multirow}
\usepackage{array}
\usepackage{caption}
\usepackage{subcaption}
\usepackage{pgfplots}
\usepackage{tikzsymbols}
\pgfplotsset{compat=1.18} 
\usepackage[11pt]{moresize}
\usepackage[percentage]{overpic}
\usepackage{pifont}
\usepackage{bbm}

\usepackage{orcidlink}

\usepackage{dsfont}
\usetikzlibrary{external}
\definecolor{forestgreen}{rgb}{0.13, 0.55, 0.13}
\definecolor{cPLOT1}{RGB}{214,113,176}
\definecolor{cPLOT3}{RGB}{80,150,80}
\definecolor{cPLOT4}{RGB}{165, 124, 27}
\definecolor{cPLOT2}{RGB}{68, 33, 175}
\definecolor{cPLOT5}{RGB}{39, 174, 239}
\definecolor{cPLOT6}{RGB}{179,0,0}
\definecolor{cPLOT7}{RGB}{199,21,133}
\definecolor{cPLOT8}{RGB}{30,144,255}

\newcommand{\bm}{\texttt{BeCoS}}%

\newcommand{\cmark}{{\color{forestgreen} \ding{51}}}%
\newcommand{\xmark}{{\color{red} \ding{55}}}%
\newcolumntype{?}{!{\vrule width 1pt}}

\title[BeCoS: Beyond Complete Shapes]%
      {Beyond Complete Shapes: A Benchmark for Quantitative Evaluation of 3D
      Shape Surface Matching Algorithms}

\author[V. Ehm, N. El Amrani et al.]
{\parbox{\textwidth}{\centering V.~Ehm\thanks{Authors contributed equally.}$^{1,3}$\orcid{0009-0009-0142-5442},
        N.~El~Amrani$^{\dagger 2,4}$\orcid{0009-0004-9961-2855},
        Y.~Xie$^{1}$\orcid{0009-0003-8670-6003},
        L.~Bastian$^{1,3}$\orcid{0000-0001-8088-3920},
        M.~Gao$^{1,3}$\orcid{0009-0008-4303-2893},
        W.~Wang$^{2,4}$\orcid{0009-0003-6637-6860},\\
        L.~Sang$^{1,3}$\orcid{0009-0007-1158-5584},
        D.~Cao$^{2,4}$\orcid{0000-0002-6505-6465},
        T.~Weißberg$^{2,4}$\orcid{0009-0005-4589-6909},
        Z.~Lähner$^{2,4}$\orcid{0000-0003-0599-094X},
        D.~Cremers$^{1,3}$\orcid{0000-0002-3079-7984},
        and F.~Bernard$^{2,4}$\orcid{0009-0008-1137-0003}
    }
    \\
{\parbox{\textwidth}{\centering 
         $^1$Technical University of Munich, Germany, $\quad$ 
         $^2$University of Bonn, Germany \\
         $^3$Munich Center for Machine Learning, Germany $\quad$ 
         $^4$Lamarr Institute, Germany 
       }
}
}

\begin{document}

\teaser{
 \includegraphics[width=\linewidth]{vis/teaser_high_res.pdf}
 \centering
  \caption{Our benchmark \bm{} provides \textbf{correspondences between thousands of shapes} (visualised via colour transfer).
  As a special focus, we consider the generation of partial shapes, where we can generate partial shape matching instances across diverse four-legged animals and humanoid categories (grey parts indicate the missing parts of partial shapes).\\
  }
\label{fig:teaser}
}

\maketitle

\input{sec/00_abstract}

\input{sec/01_introduction}

\input{sec/02_related}
\input{sec/03_data_generation}

\input{sec/04_benchmark}

\input{sec/05_limitations}
\input{sec/06_conclusion}


\printbibliography

\input{supplementary/main_supplementary}

\end{document}

%% file: sec/00_abstract.tex
\begin{abstract}
Finding correspondences between 3D deformable shapes is an important and long-standing problem in geometry processing, computer vision, graphics, and beyond. While various shape matching datasets exist, they are mostly static or limited in size, restricting their adaptation to different problem settings, including both full and partial shape matching. In particular the existing partial shape matching datasets are small (fewer than 100 shapes) and thus unsuitable for data-hungry machine learning approaches. Moreover, the type of partiality present in existing datasets is often artificial and far from realistic. To address these limitations, we introduce a generic and flexible framework for the procedural generation of challenging full and partial shape matching datasets. Our framework allows the propagation of custom annotations across shapes, making it useful for various applications.  
By utilising our framework and manually creating cross-dataset correspondences between seven existing (complete geometry) shape matching datasets, we propose a new large benchmark \textbf{\bm{}} with a total of 2543 shapes. Based on this, we offer several challenging benchmark settings, covering both full and partial matching, for which we evaluate respective state-of-the-art methods as baselines. 
Visualisations and code of our benchmark can be found at:  \url{https://nafieamrani.github.io/BeCoS/.}
\end{abstract}

%% file: sec/01_introduction.tex
\section{Introduction}
Finding correspondences between deformable shapes is a fundamental problem in geometry processing, with various applications such as texture and deformation transfer~\cite{dinh2005texture,ovsjanikov2012functional}, or statistical shape analysis \cite{hasler2009statistical,bogo2014faust, elamrani2024fuss}. The problem has been widely studied in the literature ~\cite{ovsjanikov2012functional,eisenberger2020,abdelreheem2023zero,wimmer2024back,cao2023unsupervised}, and can vary significantly in complexity. For example, current methods still struggle when estimating severe \textit{non-isometric} deformations for shapes of different categories~\cite{dyke2020shrec,panine2022non,eisenberger2020,bastian2023hybrid,cao2024spectral}, or when they are only partially observed~\cite{roetzer2022scalable,ehm2024partial}.
Current non-isometric shape datasets are often limited in size and flexibility~\cite{Zuffi:CVPR:2017,melzi2019shrec,dyke2020shrec,magnet2022smooth,li20214dcomplete}, restricting their ability to support dense correspondence propagation or custom annotations, which reduces their applicability to broader tasks.

In real-world scenarios, it is particularly common to encounter shapes with missing parts due to occlusions, limited sensor coverage, or acquisition errors during 3D scanning.
The setting in which at least part of one shape does not exist on the other is called partial shape matching~\cite{rodola2017partial,attaiki2021dpfm,roetzer2022scalable,ehm2023geometrically,ehm2024partial}. Despite its practical relevance, partial shape matching has received limited attention in the context of machine learning-based approaches, partly because of the increased difficulty, but largely due to the lack of suitable and realistic large-scale datasets, a requirement for data-hungry learning algorithms. The full-to-full matching setting is not only simpler but also benefits from wider data availability which has supported the development of data-driven methods that already display convincing performance, even without supervision~\cite{halimi2019unsupervised,roufosse2019unsupervised,cao2023unsupervised}. 
In contrast, existing datasets for partial shape correspondence often suffer from a small number of shapes, limited realism of generated partiality (e.g., cuts or holes)~\cite{cosmo2016shrec}, and a lack of diversity in shape categories~\cite{bracha2023partial,attaiki2021dpfm}. This limited data availability hampers the progress of data-driven methods for partial shape matching; recent approaches have thus required pre-training with large full-geometry datasets~\cite{cao2023unsupervised}, or rely on ground-truth correspondences~\cite{attaiki2021dpfm} which are challenging to obtain in practice. 

To address the absence of large datasets and foster research on (partial) shape correspondence, we introduce a procedural data generation framework to generate diverse shape pairs from a shape network.
By simulating  more realistic partiality patterns using ray casting, we bridge the gap between existing synthetic datasets and real-world partial shape correspondence problems. Furthermore, we propose \bm{}, a carefully curated benchmark based on existing datasets comprising diverse shape categories, thus enabling
to assess the performance of existing and foster the development of novel methods. Through extensive experiments of current state-of-the-art partial and full shape matching methods, we demonstrate the limitations of partial
methods and highlight the need for further research in this area. \\
\paragraph*{Contributions.}
We summarise our main contributions as follows:
\begin{itemize}
\item \textbf{Shape Network for Correspondence Propagation:} We unify multiple existing mesh-based datasets through manual cross-category annotations, constructing a large and diverse shape network that indicates between which shapes correspondences exist. This shape network facilitates the propagation of dense correspondences and custom annotations across shapes, enhancing its applicability to various shape analysis tasks.
\item \textbf{Scalable Framework for Partial Shape Correspondence:} We introduce a flexible and scalable framework that enables the generation of
training examples for partial shape correspondence tasks that is applicable to new datasets.
\item \textbf{Benchmark and Performance Analysis:} 
As a concrete realization of our framework, we present a challenging non-rigid shape dataset, serving as a standardised benchmark for shape matching methods. We further conduct an extensive evaluation of state-of-the-art techniques on this benchmark, offering insights into their strengths, limitations, and key challenges. Our results indicate that partial shape matching remains an open problem, motivating future research in the field.
\end{itemize}

%% file: sec/02_related.tex
\section{Related Work}
\label{sec:rw}
\subsection{Datasets}
Although several 3D shape datasets
exist, we align with established deformable
shape matching literature by focusing on humanoid and four-legged animal shapes. These categories support continuous deformations and offer well-defined dense correspondences, making them particularly suitable for our study.
In the following, we provide an overview of existing datasets for full-to-full (F2F), partial-to-full (P2F), and partial-to-partial (P2P) shape matching datasets.
Table~\ref{tab:datasets} summarises existing deformable shape matching datasets.

\noindent \textbf{Full-to-Full Datasets.} 
Full-to-full shape correspondence datasets, such as FAUST~\cite{bogo2014faust} and SCAPE~\cite{anguelov2005scape}, have been essential in methodological progress on shape correspondence. They encompass humans in various poses, initially emphasising point-wise correspondence recovery under isometric deformations. 
SHREC'19~\cite{melzi2019shrec} was introduced to evaluate the generalisation capabilities of methods to different mesh connectivity on human scans.
Datasets like TOSCA~\cite{bronstein2008numerical}, SMAL~\cite{Zuffi:CVPR:2017}, SHREC'20~\cite{dyke2020shrec}, and DeformingThings4D~\cite{magnet2022smooth, li20214dcomplete} have shifted the emphasis towards non-isometric correspondence between various animal categories and humanoid figurines.
Although deep learning methods have achieved state-of-the-art performance in shape correspondence, non-isometric datasets remain relatively small and static, limiting their applicability in training and evaluating deep learning-based correspondence methods.
While full-to-full shape matching has received considerable attention, the partial setting has been explored less (see Table~\ref{tab:datasets}).

\input{tab/tab_related_partial_dataset}

\noindent \textbf{Partial-to-Full Datasets.}
In this setting, a partial shape is deformably matched to a shape template with complete geometry. SHREC'16~\cite{cosmo2016shrec}, based on the TOSCA~\cite{bronstein2008numerical} dataset, was the first benchmark for this problem setting. PFAUST~\cite{bracha2023partial} and PFARM~\cite{attaiki2021dpfm} were later introduced using a similar approach based on the FAUST~\cite{bogo2014faust} and FARM~\cite{kirgo2021wavelet} datasets, respectively. SHREC'20~\cite{dyke2020shrec} also contains a subset of full-to-partial correspondence pairs.

\noindent \textbf{Partial-to-Partial Datasets.}
Partial-to-partial shape matching involves matching two partial shapes with unknown overlap, making it considerably more challenging than partial-to-full matching. %
In~\cite{attaiki2021dpfm} the authors proposed the first benchmark for this setting with CP2P~\cite{attaiki2021dpfm} based on the SHREC'16 CUTS~\cite{cosmo2016shrec} benchmark. Recently, PSMAL~\cite{ehm2024partial} was introduced, extending the SMAL~\cite{Zuffi:CVPR:2017} dataset to include partial-to-partial correspondence pairs.

As can be seen in Table ~\ref{tab:datasets}, most existing datasets for partial shape matching (P2F and P2P) focus only on near-isometric problem settings. Furthermore, they contain only a limited number of shapes, impeding the capabilities of learning-based approaches, particularly when considering unsupervised settings for partial non-isometric correspondence settings.

\subsection{Methods for Shape Correspondence}
We proceed by discussing a selection of shape matching methods and refer to recent surveys~\cite{sahillioglu_recent_2020,deng_survey_2022} for a more thorough treatment.
While point-cloud matching methods also exist~\cite{li2022lepard,bai2020d3feat,huang2021predator, yu2023rotation}, we focus on mesh-based methods, which are better aligned with the surface representations emphasised in our dataset.
Methods addressing shape correspondence can be categorised into axiomatic and learning-based. Among the axiomatic approaches are discrete optimization methods~\cite{bernard2020mina} and spectral methods utilising the powerful framework of functional maps~\cite{ovsjanikov2012functional,rodola2017partial,melzi2019zoomout}. However, the performance of the latter class typically degrades as assumptions of isometry are violated, such as in partial correspondence settings.

\paragraph*{Full-to-Full Shape Matching.}
Spectral methods based on the eigenfunctions of the Laplace-Beltrami Operator (LBO) have been successful in full-to-full matching settings~\cite{ovsjanikov2012functional, litany2017deep, donati2020deepGeoMaps, bastian2023hybrid, cao2023unsupervised}, yielding a compact solution.
In contrast to spectral approaches, discrete optimization formulations have also been introduced for full-to-full matching~\cite{bernard2020mina, roetzer2022scalable, roetzer2023fast, roetzer2024spidermatch}. These algorithms often come with guaranteed bounds on optimality, sometimes incorporating geometric consistency constraints.

\paragraph*{Partial-to-Full Shape Matching.}
Rodola et al.~\cite{rodola2017partial} demonstrate that for partial-to-full matching, the functional map has a slanted diagonal structure, which has, for example, been used in \cite{litany2016non} to solve multi-part partial-to-partial correspondence; however, under the assumption that a full template is given. Several methods also operate in the spectral domain in learning-based settings~\cite{attaiki2021dpfm,cao2022unsupervised,cao2023unsupervised}.
Other works do not use functional map layer~\cite{bracha2023partial, bracha2024wormhole} and explicitly tackle the partial-to-full case.

\paragraph*{Partial-to-Partial Shape Matching.}
Few works thus far explicitly address the partial-to-partial matching setting, which is particularly challenging as the overlapping region between shapes is unknown. As such,  the overlapping region must simultaneously be identified and matched under an unknown deformation. DPFM~\cite{attaiki2021dpfm} demonstrates that spectral methods are applicable in this domain using an additional attention mechanism, albeit using ground-truth supervision for feature learning. In~\cite{windheuser2011geometrically} shape matching is formulated as an integer linear program (ILP) over the space of orientation-preserving diffeomorphisms. SM-COMB~\cite{roetzer2022scalable} demonstrated that the latter constitutes a strong prior to guide partial-to-partial matching. However, holes must be synthetically filled to obtain a closed manifold. Recently, GC-PPSM~\cite{ehm2024partial} formulated a geometrically consistent partial-to-partial shape matching non-linear integer program. While both SM-COMB and GC-PPSM are compatible with various surface descriptors, they achieve SOTA results only with high-quality pre-trained input features (e.g. from ULRSSM~\cite{cao2023unsupervised}), which can be prohibitive when these are not available~\cite{roetzer2022scalable,ehm2024partial}.

\subsection{Feature Descriptors for Shape Correspondence}
\label{sub:features}

Many axiomatic and learning-based methods for shape matching rely on surface feature descriptors as guidance for correspondence computation.
Feature descriptors, like SHOT~\cite{tombari2010unique}, can be handcrafted based on extrinsic quantities like XYZ-coordinates or surface normals.
Many surface descriptors are sensitive to surface degradation, such as holes, or are not invariant under rotations, as is the case for XYZ coordinates.
The common intrinsic heat kernel (HKS)~\cite{sun_concise_2009} and wave kernel signatures (WKS)~\cite{aubry_wave_2011} are invariant under isometric deformations; however, they are not effective under partiality as the isometry assumption can be heavily violated.
On the other hand, methods like DiffusionNet~\cite{sharp2022diffusionnet} aim to learn robust surface descriptors from data while still using some features like XYZ coordinates or WKS descriptors as input.
Considering that 3D shapes of many datasets are typically oriented in an upright manner~\cite{bogo2014faust,anguelov2005scape,bronstein2008numerical,rodola2014dense,dyke2020shrec}, or generally positioned in a canonical pose~\cite{eisenberger2021neuromorph} this can lead to biased predictions.
We discuss these biases in detail in Sec.~\ref{sec:challenges}.
Moreover, learned features require large training datasets and a high-quality (self-)supervision signals which is challenging to achieve under severe partiality. 
Since obtaining per-vertex feature descriptors directly on 3D shapes is challenging, \cite{wei2016dense} addresses this by establishing correspondences between 3D shapes through correspondences between their 2D rendered images.
Recent methods \cite{dutt2024diff3f, decatur20233d, kobayashi2022decomposing, fischer2024nerf} leverage 2D image foundation model features \cite{oquab2023dinov2, rombach2022high}
from multiple rendered views of a 3D shape to aggregate per-vertex features. 
However, such an approach can suffer from significant bias due to missing semantic information in the rendered views. 

%% file: tab/tab_related_partial_dataset.tex
\newcommand{\HumanoidShape}{\Strichmaxerl[1.3]}
\begin{table}[!t]
\centering
\footnotesize
\resizebox{\columnwidth}{!}{%
    \begin{tabular}{@{}cl@{}ccc@{}}
    \toprule
    \multirow{2}{*}{\textbf{Setting}} & \multirow{2}{*}{\textbf{Dataset}} & \textbf{\# Unique} & \textbf{Shape} & \textbf{Non}\\
    & \textbf{} & \textbf{Shapes} & \textbf{Type} & \textbf{Isometry}\\
        \midrule
        \multirow{9}{*}{F2F} &FAUST~\cite{bogo2014faust}                                                              & 100 & \HumanoidShape & \xmark \\
&SCAPE~\cite{anguelov2005scape}                                                          & 71  & \HumanoidShape & \xmark \\
&SMAL~\cite{Zuffi:CVPR:2017}                                          & 49 & \Cat & \cmark\\
&DT4D Human~\cite{magnet2022smooth, li20214dcomplete}                 & 248 & \HumanoidShape & \cmark\\
&DT4D Animal~\cite{magnet2022smooth, li20214dcomplete}               & 1950 & \Cat & \xmark\\
&TOSCA Human~\cite{bronstein2008numerical}                                               & 45 & \HumanoidShape & \xmark\\
&TOSCA Four-legged~\cite{bronstein2008numerical}                                         & 37 & \Cat & \xmark\\
&KIDS~\cite{rodola2014dense}                                                             & 32    & \HumanoidShape & \xmark\\
&SHREC'20~\cite{dyke2020shrec}                                        & 11 & \Cat & \cmark\\
&SHREC'19~\cite{melzi2019shrec}                                     & 44 & \HumanoidShape & \cmark\\
    \midrule
    \multirow{6}{*}{P2F} & SHREC'16~\cite{bronstein2008numerical,cosmo2016shrec} & 76 
    & \HumanoidShape/\Cat
    & \xmark
    \\
    & PFAUST~\cite{bracha2023partial, bogo2014faust} & 10 &
    \HumanoidShape
    & \xmark
    \\
    & FARM~\cite{kirgo2021wavelet} & 
    5 & 
    \HumanoidShape 
    & \xmark
    \\
    & PFARM~\cite{attaiki2021dpfm,kirgo2021wavelet} & 28 
    & \HumanoidShape
    & \xmark
    \\
    & SHREC'20~\cite{dyke2020shrec} & 14
    & \Cat 
    & \cmark
    \\
    \addlinespace
    \midrule
    \multirow{2}{*}{P2P}
    & CP2P~\cite{attaiki2021dpfm, cosmo2016shrec} & 76 & 
    \HumanoidShape/\Cat
    & \xmark
    \\
    & PSMAL~\cite{ehm2024partial, Zuffi:CVPR:2017} & 43 &
    \Cat 
    & \cmark
    \\
    \midrule
    Both & \textbf{\bm{}} (ours) & \textbf{2543} & \HumanoidShape/\Cat
    & \cmark
    \\
    \bottomrule
    \end{tabular}
    }
    
\caption{\textbf{An Overview of 3D Deformable Shape Datasets}, categorised as full-to-full (F2F), partial-to-full (P2F) and partial-to-partial (P2P), including the number of available shapes, shape type (Humanoid \HumanoidShape or Four-legged \Cat), and deformation class (Near-Isometric (\xmark) and Non-Isometric (\cmark)). Due to limited correspondence pairs, FARM and PFARM are typically only used for evaluation.
Note that in previous datasets, not all unique shapes have annotated correspondences, limiting the number of possible correspondence pairs, e.g., DT4D Animal (maximum 62~452 correspondence pairs) compared to our benchmark (maximum 1~414~179 correspondence pairs).
As SHREC20~\cite{dyke2020shrec} contains three partial shapes, it can be used as a partial-to-full or full-to-full dataset.
} 
\label{tab:datasets}
\end{table}

%% file: sec/03_data_generation.tex
\input{fig/fig_pipeline}

\section{Our Cross-Dataset Shape Matching Network 
}
\label{sec:generation}

We propose a novel framework enabling the procedural generation of shape matching instances with ground-truth correspondences based on a given collection of 3D shapes. Our framework supports shapes from multiple categories to generate shape matching instances of non-isometric shape pairs. Furthermore, we manually link correspondences between multiple datasets to vastly increase the diversity and representativeness.
Our framework provides researchers with the possibility to propagate dense correspondences or custom annotations (e.g.~left/right annotations, which we later consider as demonstrative example) through the entire collection of 3D shapes.
Our pipeline supports re-meshing while preserving shape correspondences within a collection and thus fosters discretisation-agnostic method development.
To overcome the limitations in realism and sample size of existing partial shape benchmarks, we specifically focus on partial shape generation.
We use ray casting to generate more realistic partial shapes than existing benchmarks based on synthetic planar cuts and holes.
We summarise the main ideas in Fig.~\ref{fig:method_overview}.
A detailed description of our data generation can be found in the supp.~mat.~Sec.~\ref{sec:procedure_data}.
We apply our framework to construct a new large benchmark combining different shape correspondence datasets (Section~\ref{sec:benchmark}) and show open challenges in the field through an evaluation of existing methods on this benchmark, as detailed in Sec.~\ref{sec:evaluation}.

\subsection{Cross-Dataset/Category Correspondences} \label{sub:crossdataset}
Current datasets commonly lack correspondences between different categories, limiting data availability and diversity~\cite{bronstein2008numerical, li20214dcomplete}.
We address this shortcoming by manually establishing correspondences across categories \emph{and} across different datasets.
To this end, we treat each category for which a cross-category correspondence is not available due to missing annotation
as a separate shape group.
Then, we select a template shape for each group and manually choose a set of corresponding landmarks on pairs of templates using FaceForm \cite{faceform} (see supp.~mat.~Sec.~\ref{sec:procedure_data}). In the case of non-isometry, more manual landmarks are needed to guide the deformation, see Fig.~\ref{fig:face_form} in supp. mat. for an example. We then morph one template into another to establish dense point-to-surface correspondences (represented in barycentric coordinates to account for different discretisations). Using the tightly aligned shapes, dense correspondences are obtained by projecting points onto triangles of the other surface. After the templates are in dense correspondence, we propagate them to the entire shape category using the existing intra-category correspondences. Overall, we build a shape network
of pairwise dense correspondences that cover all possible shape templates (cross-category and cross-dataset). The detailed shape network is presented on our \href{https://nafieamrani.github.io/BeCoS}{project page}.
A colourmap visualization of the correspondences can be seen in Fig.~\ref{fig:method_overview}.

\subsection{Scaling of Shapes}
\label{sec:scaling_shapes}
Our framework integrates multiple mesh-based datasets, each with varying shape scales. To harmonise these datasets, we manually define scaling factors for each dataset, ensuring shapes are scaled according to semantic criteria (e.g., the same species across datasets have consistent sizes). Consequently, our framework offers diverse non-isometrically deformed shapes with varying scaling factors (e.g., a cat is smaller than an elephant; see Fig.~\ref{fig:teaser} for examples).

\subsection{Alignment of Shapes}
\label{sec:alignment_shapes}
Many correspondence datasets, specifically partial ones, such as CP2P~\cite{attaiki2021dpfm} and SHREC'16~\cite{cosmo2016shrec}, keep shapes in the same canonical pose during training and testing. This bias can prevent methods from generalising to shapes under random rigid transformations, see also~\cite{attaiki2023generalizable}. As this assumption is often unrealistic in real-world settings, we randomly rotate the shapes around the z-axis (with respect to their canonical pose as present in the original dataset) and thus generate shapes in random upright positions.
We show an ablation study about rotation in Sec.~\ref{sec:challenges}.

\subsection{Custom Annotation Propagation}
\label{sec:custom_annotations}
Shape annotations like movable parts labels used in robotics \cite{qiu2025articulate} or part segmentation labels~\cite{kalogerakis2010learn, chen2024DAE-Net} are often indispensable for data-driven training or evaluation. Our framework allows for the propagation of any type of information defined on the vertices from a few annotated examples to all shapes in the dataset. By leveraging the shape network, we can find a path to the closest annotated example from each shape, establishing correspondences to a shape that offers ground-truth information. To demonstrate the effectiveness of our method, we choose to propagate left/right annotations from only $12$ manually annotated shapes to the whole dataset (see Figure~\ref{fig:custom_annotations}). 
While in theory, even fewer annotations would suffice, we selected $12$ shapes to reduce propagation error (see the examples in the supplementary material in Figure~\ref{fig:additional_custom_annotations}).

\input{fig/fig_custom_annotations}

\subsection{Partiality Generation}
\label{sub:partiality_generation}

\noindent \textbf{Partiality via Ray Casting.}
Given a shape $\mathcal{X}$, we introduce partiality by simulating a simple unidirectional 3D scanning procedure based on ray casting, as shown in Fig.~\ref{fig:method_overview} right. Doing so leads to a variety of different partial shape boundaries and overlapping behaviour (see Fig.~\ref{fig:teaser} for examples). To perform ray casting, we first centre and scale each full shape $\mathcal{X}$ to be within a fixed unit bounding box and then choose random camera poses on the unit sphere. Partiality is obtained by casting $R$ rays from the sampled camera pose onto the full shape $\mathcal{X}$. Rays stop upon hitting a triangle. All hit triangles that form the largest connected component (by area) are kept to be part of the partial shape $\hat{\mathcal{X}}$. Finally, each partial shape $\hat{\mathcal{X}}$ is scaled and translated back to its original size and location. 
To generate a shape pair for partial-to-partial shape matching we get a partial shape $\hat{\mathcal{Y}}$ from a second full shape $\mathcal{Y}$.

Random camera poses could lead to non-existent or prohibitively small overlap in the partial-to-partial matching setting. Therefore, we sample two camera poses with the constraint that their maximum angular disparity in both azimuth and elevation does not exceed $\alpha$ degrees. Through this, we attain an overlap of a certain size to ensure meaningful matching instances. 
For the overlap computation, we rigidly align the full shapes $\mathcal{X}$ and $\mathcal{Y}$ 
with Procrustes analysis
using the given correspondences to ensure that both shapes are in a similar position i.e.~both shapes are up-right and facing the same direction. Then, we perform the previously described partiality generation on both shapes to generate  $\hat{\mathcal{X}}$ and $\hat{\mathcal{Y}}$.
Subsequently, we quantify the overlap as the percentage of vertices from $\hat{\mathcal{X}}$ that are mapped to $\hat{\mathcal{Y}}$ and vice-versa. If neither of these quantities falls within a predefined range (see Sec.~\ref{sec:benchmark} for details), the process proceeds with new random camera poses until the overlap meets the specified criteria, or until a maximum of $m$ iterations is reached. In the latter case, the partial shapes whose overlap is the closest to the desired range are selected (see Sec.~\ref{sec:p2p_overlapping_region} in supp.~mat.~for more details). 

\noindent \textbf{Flexible Generation of Shape Matching Instances.}
Existing partial shape datasets provide a precomputed and fixed number of holes/cuts~\cite{cosmo2016shrec, bracha2023partial, kirgo2021wavelet, attaiki2021dpfm, dyke2020shrec, ehm2024partial}, creating partial shape pairs from identical meshes in varying poses. Instead, our proposed framework can generate partial shapes from a larger and more diverse pool of shapes, allowing for pairing compatible shapes from different categories, potentially originating from different datasets. This creates a more challenging scenario encompassing simultaneous non-isometric deformations and partiality, representing an important step towards more realistic settings. Furthermore, our framework introduces more variety by using different camera poses for ray casting and re-meshing shapes to different resolutions.

\subsection{Semantic Missing Parts}
By combining multiple categories and datasets, we encounter cases where dense correspondences cannot be reliably established between certain parts of shapes without introducing inaccuracies (e.g., matching a deer's horn to a horse’s head). 
While such ambiguities can be managed when considering semantic correspondences~\cite{abdelreheem2023zero}, they remain problematic for dense correspondences.
To address this, for the first time we create large cross-dataset correspondences with semantic missing parts annotations, as shown in Fig.~\ref{fig:f2f_partiality}. 
To this end, we modify our pipeline described in Sec.~\ref{sub:crossdataset} by manually removing the semantic missing parts from the shapes and match the rest of the shapes to get the dense correspondences. We show an example in Fig.~\ref{fig:face_form} in supp. mat.
This creates a novel objective for shape matching methods that warrants future research.
\input{vis/partial_matchings_f2f/partial_f2f}

%% file: fig/fig_pipeline.tex
\begin{figure*}[ht!]
    \centering
\includegraphics[width=0.965\textwidth]{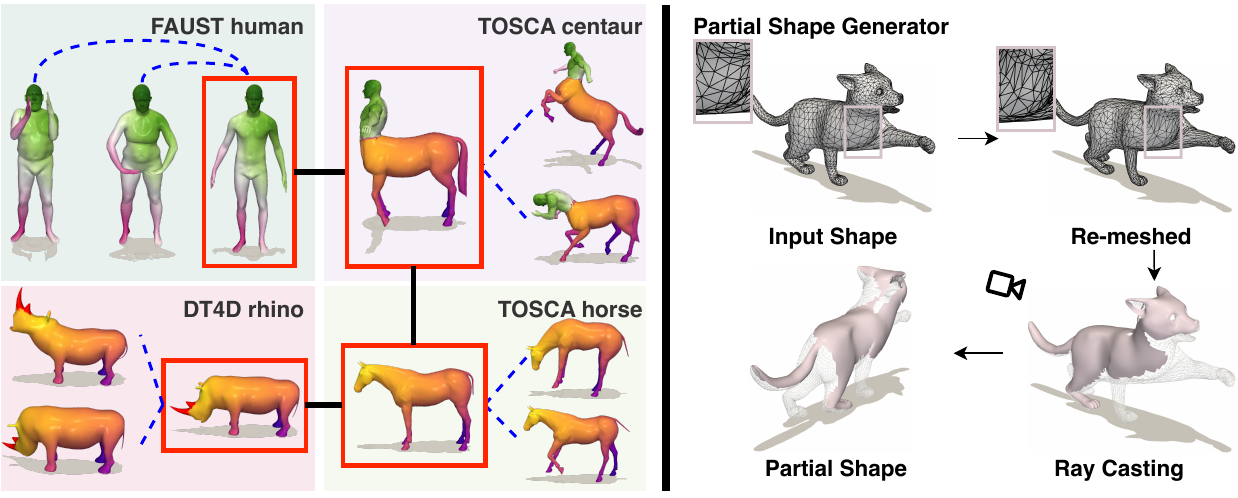}
    \caption{
    \textbf{Left:} We use a \textbf{collection of multiple complete geometry datasets} to generate novel shape pairs.
    Using FaceForm (see supplementary material), we manually annotate cross-category and cross-dataset correspondences (black lines, see Sec.~\ref{sub:crossdataset}) between (manually selected) templates (shapes in red boxes). These are then combined with existing intra-dataset correspondences (blue dashed lines) to construct a shape network, through which dense correspondences can be propagated. \textbf{Right:} Given an input shape with full geometry from our shape collection, we first \textbf{re-mesh} the shape, then \textbf{apply ray casting} to simulate a unidirectional 3D scanning procedure to introduce partiality (see Sec.~\ref{sub:partiality_generation}), and finally apply a deterministic random \textbf{rotation}.} 
    \label{fig:method_overview}
\end{figure*}

%% file: fig/fig_custom_annotations.tex
\begin{figure}[!ht]
    \centering
    \begin{tabular}{@{}c|ccc@{}}
    \setlength{\tabcolsep}{0pt} 
        \adjustbox{valign=m}{\includegraphics[height=0.11\textheight]{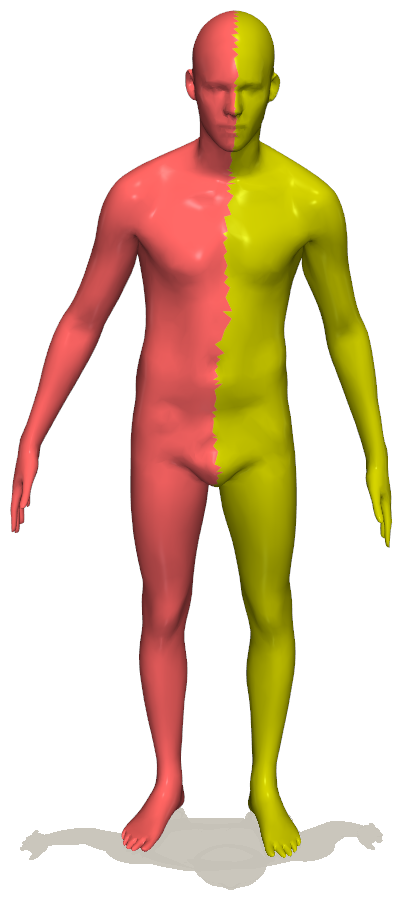}} &
        \adjustbox{valign=m}{\includegraphics[height=0.11\textheight]{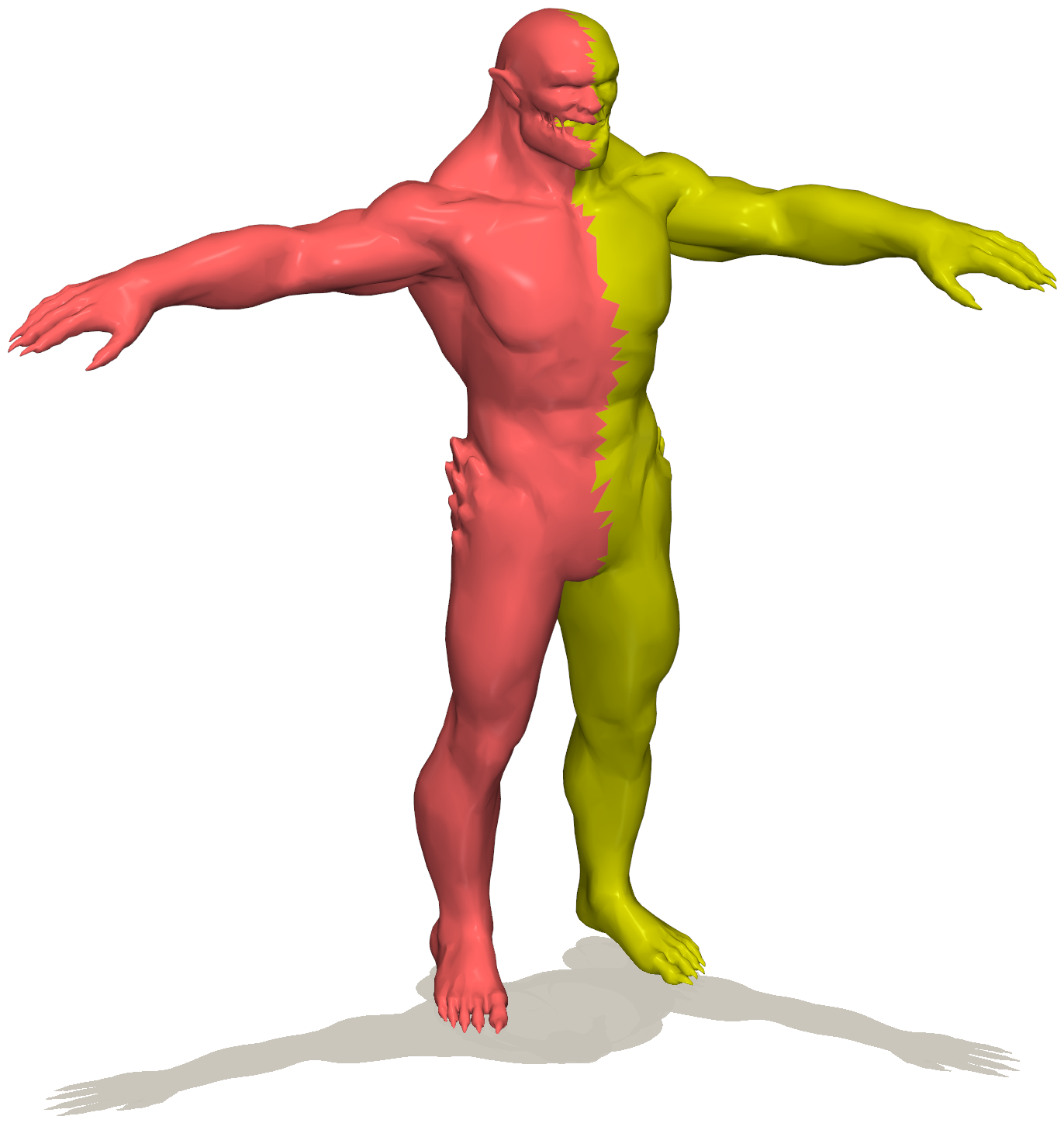}} &
        \adjustbox{valign=m}{\includegraphics[height=0.11\textheight]{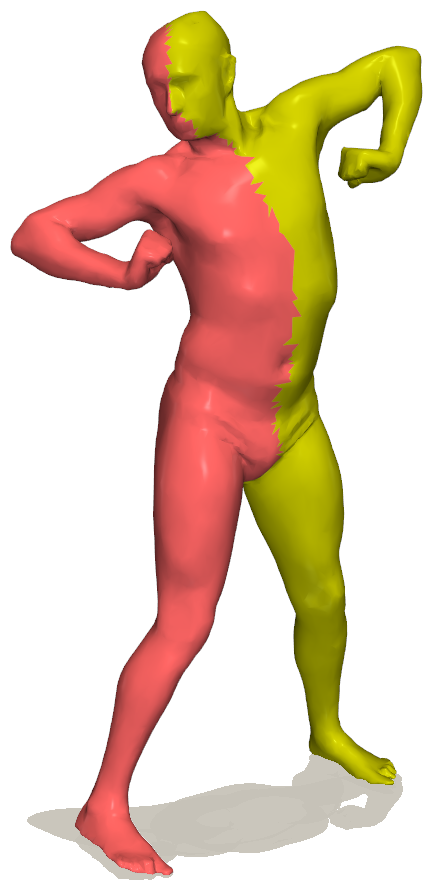}} &
        \adjustbox{valign=m}{\includegraphics[height=0.11\textheight]{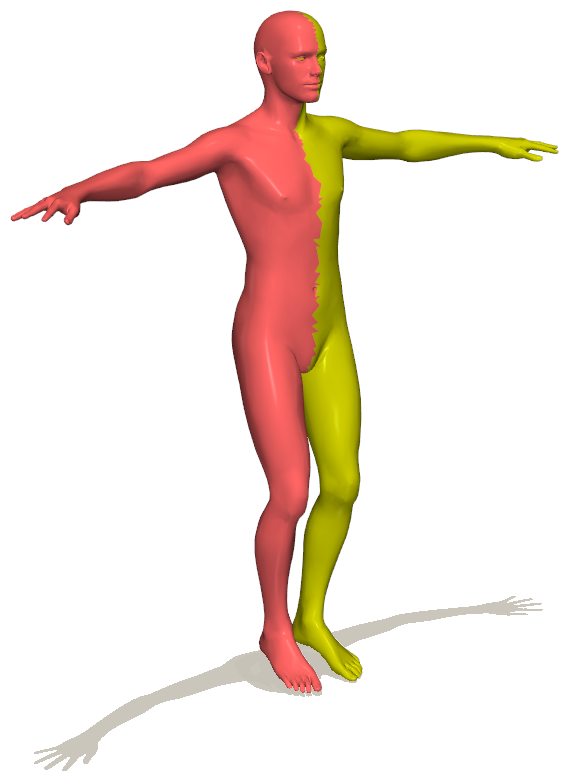}} 
        \\
        \small{\textbf{(a)}} & \multicolumn{3}{c}{\textbf{(b)}} \\
    \end{tabular}
    \caption{Our framework allows for the \textbf{propagation of custom annotations} throughout all shapes. We show the case of left/right annotations. \textbf{(a):} Left/Right annotations are manually annotated on the shape considered as template. \textbf{(b):} Left/Right annotations are propagated from the closest template in the shape network (see Sec.~\ref{sec:supp_custom_annotation} in supp. mat. for more examples).}
    \label{fig:custom_annotations}
\end{figure}

%% file: vis/partial_matchings_f2f/partial_f2f.tex
\begin{figure}[!ht]
    \centering
    \begin{tabular}{@{}c@{}@{}c|c@{}@{}c@{}}
    \setlength{\tabcolsep}{0pt}  
        

        \adjustbox{valign=m}{\includegraphics[height=0.1\textheight]{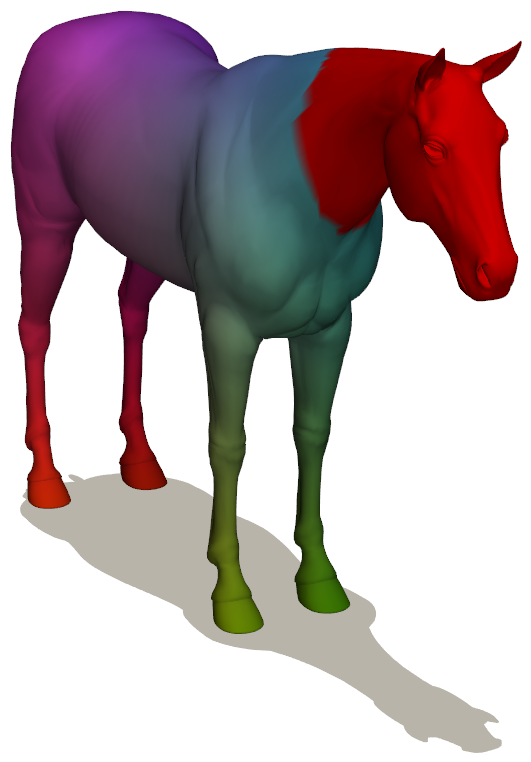}} & 
        \adjustbox{valign=m}{\includegraphics[height=0.12\textheight]{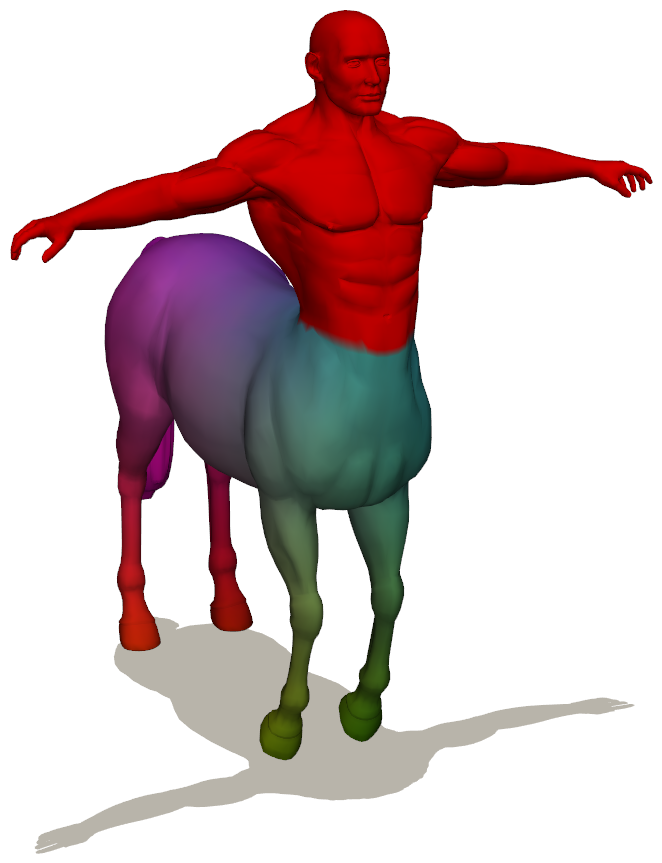}} &
        \adjustbox{valign=m}{\includegraphics[height=0.09\textheight]{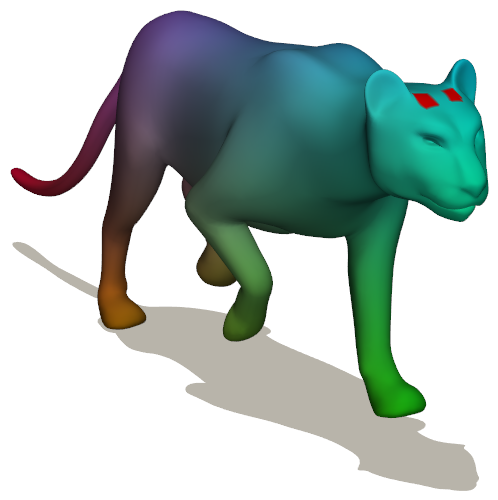}} &
        \adjustbox{valign=m}{\includegraphics[height=0.09\textheight]{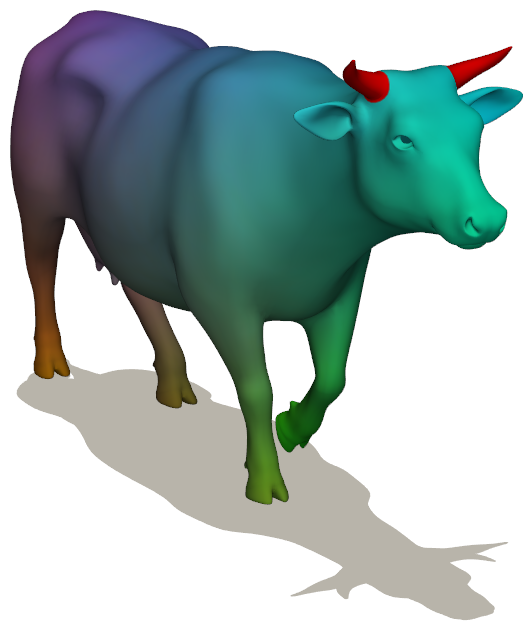}} \\
    
        \small{Source} & \small{Target} &  \small{Source} & \small{Target}

    \end{tabular}
    \caption{For the first time we define \textbf{semantic missing parts in full shapes.} The horse and puma do not possess a human upper body or horns, making these features unmatchable in the centaur and cow, respectively.}
    \label{fig:f2f_partiality}
\end{figure}

%% file: sec/04_benchmark.tex
\section{The \bm{} Benchmark}
\label{sec:benchmark}

In addition to the procedural data generation framework, we propose the \bm{} benchmark, which is a specific instantiation of our general framework introduced in Sec.~\ref{sec:generation}.
\bm{} contains
realistic full and partial shapes from established datasets with cross-category and cross-dataset correspondences.
The purpose of \bm{} is to challenge current state-of-the-art shape matching methods, specifically in the partial case, and facilitate progress in the field through a standardised evaluation.
To create our benchmark, we apply the pipeline introduced in Sec.~\ref{sec:generation} on a combination of several existing datasets.
We train and evaluate various existing shape correspondence methods on \bm{} (see Sec.~\ref{sec:baselines} for an overview), demonstrating the shortcomings of existing methods, particularly for partial-to-partial matching.

\noindent \textbf{Datasets.}
\bm{} is composed of seven different datasets
with four-legged creatures and humanoid shapes encompassing challenging and diverse shape matching scenarios for both isometric and non-isometric deformations: TOSCA~\cite{bronstein2008numerical}, FAUST~\cite{bogo2014faust}, 
SCAPE~\cite{anguelov2005scape}, KIDS~\cite{rodola2014dense}, DT4D~\cite{magnet2022smooth}, SMAL~\cite{Zuffi:CVPR:2017} and SHREC'20~\cite{dyke2020shrec}. 
See Table~\ref{tab:datasets} for more information about the used datasets.
From these datasets, we use a total of 2543 shapes (496 are humanoid shapes and 2047 are four-legged shapes) to generate pairwise matching problems. We partition our dataset into predefined training/validation/test splits, ensuring evaluation is performed on unseen categories. To combat dataset imbalance, we oversample the smaller datasets in our training set.
The number of shape matching instances in the train/validation/test splits is 10185/137/142. We limit the test set size to easily compare to axiomatic methods, which are often computationally heavy.

\noindent \textbf{Predefined Benchmark Parameters.}
We choose the following parameters in the data generation pipeline to create a challenging but reasonable benchmark. For re-meshing, we use quadric decimation~\cite{garland1997surface} and we fix the range of vertices to a random number between $9000{-}10{,}000$ to facilitate training of current supervised/unsupervised methods and limit memory consumption. Note that re-meshing the same shape multiple times results in different re-meshed versions of the shape due to the random selection of the number of vertices. Our main setting introduces a random rotation around the z-axis, assuming each shape has a natural `upright' direction. The maximum camera angle for the partial-to-partial setting is set to $\alpha=\frac{\pi}{4}$. Similar to~\cite{attaiki2021dpfm} and~\cite{ehm2024partial}, these parameters are chosen to maintain an overlap of $10\% - 90\%$, using $m=10$ maximum iterations (see supp.~mat.~Sec.~\ref{sec:p2p_overlapping_region} for distribution of overlap).
All default parameters to generate the BeCoS benchmark are explicitly stated and explained in details in our implementation repository.

The parameter configuration of \bm{} is chosen to significantly challenge state-of-the-art methods while still being attainable.
In the future, if the benchmark is saturated due to the development of more accurate shape matching methods, the customisable nature of our framework enables a straightforward construction of more challenging \bm{} variants (e.g. random rotation around all axes).

\input{tab/tab_benchmark_list_methods}
\input{vis/qualitative_results_all/qualitative_all_double_column}
\input{fig/fig_benchmark_results}

\section{Evaluation}
\label{sec:evaluation}
We evaluate different shape matching methods, specifically showcasing the difficulty posed by partiality.
We additionally present numerous ablation studies and further experiments with custom annotations (left/right annotations), showcasing the challenges and possibilities of \bm{}.

\subsection{Evaluation Metrics}
\label{sec:eval_metrics}
For quantitative evaluation, we consider well-established quantitative measures: we use the \textbf{Geodesic Error} to quantify the correspondence accuracy, as well as the \textbf{Intersection over Union} (IoU) and the \textbf{F1 Score} to quantify the overlap.
For the ablation study on custom annotation with left/right prediction, we use a \textbf{Left/Right Accuracy} metric.

\noindent \textbf{Geodesic Error.}
We evaluate the geodesic error of correspondences using the Princeton Protocol~\cite{kim2011blended}. This evaluation entails normalising the geodesic error between the ground truth and the computed correspondences by the shape diameter, defined as the square root of the area of the target shape. For partial shapes, we use the square root of the area of their corresponding full shape. To handle the partial-to-partial matching paradigm, we follow~\cite{ehm2024partial} and set an infinite geodesic error for unmatched vertices that should be matched and for vertices that are matched but should not be. In contrast to the partial-to-full (P2F) and full-to-full (F2F) settings, where no overlapping region is determined, the overlapping region prediction affects the geodesic error curve for the partial-to-partial setting. 

\noindent \textbf{Intersection over Union.}
To evaluate the quality of the overlapping areas induced by a matching, we consider the Intersection over Union (IoU). To this end, we define $P \in \{0,1\}^{(|V| \times 1)}$ and $G \in \{0,1\}^{(|V| \times 1)}$ as vectors (predicted and ground truth, respectively), indicating which vertices are part of the matching.
We quantitatively evaluate the overlapping region in the partial-to-partial setting using the intersection over union (IoU), calculated as $\text{IoU} = \frac{|P \cap G|}{|P \cup G|}$.

\noindent \textbf{F1 Score.}
To give further insight into the overlap prediction, we use the F1 score.
It integrates precision ($\frac{\sum P \cap G}{\sum P}$) and recall ($\frac{\sum P \cap G}{\sum G}$) through the harmonic mean:
\text{F1 score}~=~$\frac{2\cdot \text{precision}\cdot \text{recall}}{\text{precision}+\text{recall}}$. Its possible value spans $[0,1]$ with~$1$ indicating a perfect precision and recall.

\noindent \textbf{Left/Right Accuracy.}
For a shape $\mathcal{X}$ with $N$ vertices, the left/right accuracy is defined as
\begin{equation}
    acc_{\chi} = \frac{1}{N}  \sum_{i \in N}  \mathds{1}(\chi_i = \chi_i^{gt})
\end{equation}
where $\chi_i$ is the left/right prediction for the $i$-th vertex in $\mathcal{X}$, $\mathds{1}$ is the indicator function,  $\chi_i^{gt}$
is the ground truth of the left/right annotation of the $i$-th vertex in $\mathcal{X}$.
We use a similar approach for the partial-to-partial case as when using the geodesic error. 
For unmatched vertices that should be matched and vertices that are matched but should not be, we set an accuracy of $0$.

\subsection{Baselines}
\label{sec:baselines}
We choose a mix of learning-based and axiomatic methods for each of the three settings (full-to-full, partial-to-full and partial-to-partial), including SOTA methods in the supervised, unsupervised and axiomatic case. 
We train all learning-based methods on \bm{}.
We show an overview of the methods in Table~\ref{tab:benchmarks}. For the partial-to-partial problem setting, an unsupervised approach has not yet been proposed in the literature. Therefore, we include two axiomatic methods instead. One reason that learning-based approaches are not extensively used for partial-to-partial shape matching is the lack of sufficient training data, which we explicitly address in this work. The severe difficulty of partial-to-partial shape matching is manifested by a significant performance degradation compared to other matching settings, see Fig.~\ref{fig:geo_errors}.

\subsection{Results}
We show the results on full-to-full, partial-to-full and partial-to-partial settings in Fig.~\ref{fig:geo_errors}. We additionally show the quality of the overlapping region of the partial-to-partial case in Table~\ref{tab:mIoU}. Qualitative results can be found in Fig.~\ref{fig:all_qualitative_results}.

\noindent \textbf{Discussion.}
While the learning-based methods yield mostly promising results on the full geometries~(see Fig.~\ref{fig:all_qualitative_results} F2F), existing methods struggle with the challenging partial scenarios (P2F and P2P) presented in our benchmark, demonstrating a need for further methodological advances: In the partial-to-partial case DPFM returns noisy overlapping region predictions (see Fig.~\ref{fig:all_qualitative_results} P2P, third column). Both SM-COMB and GC-PPSM perform poorly, probably because they rely on informative and reliable feature descriptors for computing the matching energy. Since these are not available for \bm{}, we resort to XYZ coordinates for computing respective energies (see supp.~mat.~Sec.~\ref{sec:p2p_alternative_energies} for a justification and analysis of this choice). In the partial-to-full case, both axiomatic and unsupervised methods return bad correspondences, both qualitatively (see Fig.~\ref{fig:all_qualitative_results} P2F, third and fifth column) and quantitatively (see Fig.~\ref{fig:geo_errors} middle). We discuss further in Sec.~\ref{sec:challenges}, why our benchmark is so challenging for these settings. We believe that our work will contribute to the development of sophisticated and reliable machine learning approaches for these partial shape matching settings.
\input{tab/tab_benchmark_overlap_region}

\subsection{Ablation Studies}
\label{sec:challenges}
Multiple challenges can be simultaneously present in the shape matching problem: partiality, varying scale, non-isometry, different orientations/poses, and non-alignment of pairs of shapes. 
Previous datasets,  especially partial shape matching datasets, simplify these challenges (e.g.\ by aligning the shapes). In contrast, we include all the above mentioned challenges in our benchmark, which makes it significantly more challenging.

In order to separate the effects of partiality, scaling and rotation on existing methods we conduct two ablation studies. First, we present the effects of the scaling challenge. Then, we demonstrate how random rotations adversely affect the performance of existing SOTA methods for partial-to-full shape matching.

\input{tab/tab_benchmark_scale_analysis}

\noindent \textbf{Scaling of Shapes.}
To illustrate the effects of scaling, we compare the performance of multiple shape matching methods using the \textit{default} setting of \bm{} (as discussed in Sec.~\ref{sec:scaling_shapes}) and a \textit{normalised} setting, where the surface area of each (full) shape in the benchmark is normalised to~$1$. The results in Table~\ref{tab:p2p_normalized} show that the performance drops when using the \textit{default} setting for most methods. While full-to-full methods can easily handle the full-to-full variant of \bm{} under the \textit{default} setting by normalising the surface area to~$1$, resulting in similarly scaled shapes. This is non-trivial for the partial-to-full and partial-to-partial settings, as each partial shape represents an unknown percentage of its full counterpart, making the normalisation non-trivial. Using the normalised surface area under these two partial settings thus increases performance, especially for the supervised learning-based methods DPFM and GeomFmaps. On the other hand, the axiomatic and unsupervised methods generally perform poorly on \bm{}, making it hard to draw a conclusion about the methods' performance in the \textit{normalised} and \textit{default} settings.

\noindent \textbf{Alignment of Shapes.}
To demonstrate the impact of shape alignment on the performance of shape matching methods, we compare two different settings to probe existing methods for robustness to shape rotation on the partial-to-full setting: 
\begin{enumerate}
    \item \textit{With rotation}: Our default realistic benchmark setting \textit{with rotation}, where all shapes in the train and test sets have a random rotation around the $z$-axis (i.e.~all pairs of shapes are in different poses).
    \item \textit{Without rotation}: The train and test are not rotated, and shapes are kept as present in their respective datasets (most shapes are in a canonical pose).
\end{enumerate}

We select the partial-to-full methods for this ablation, as the partial-to-partial methods perform prohibitively poorly, resulting in a mostly random prediction accuracy. Many full-to-full methods use wave kernel signature features \cite{aubry_wave_2011} as input, making them independent of rotations.

We show the results of our ablation in Fig.~\ref{fig:ablation_rot}.
A noticeable decline in performance is observed when rotation is introduced compared to when the shapes are kept as present in their datasets. This indicates that current methods benefit from biased datasets.
The bias errantly favours XYZ features as input features when shapes are all aligned in a canonical pose.
While reliable intrinsic (and therefore rotation-invariant) features such as wave kernel signature (wks) exist, these are not robust to partiality.
This presents a challenge for future research in the domain of partial shape matching involving rotated partial shapes.
One important question is how to define rotation-invariant feature descriptors that are robust to partiality.
We believe our benchmark may serve as an important contribution to tackling this challenge.

\input{fig/fig_benchmark_rotation}

\subsection{Custom Annotations: Left/Right Predictions}
\label{sec:chirality_benchmark}
As a proof of concept, we show our framework's ability to propagate custom annotations throughout the dataset. To this end, we provide a quantitative evaluation of the performance of shape matching methods in predicting left/right parts on our benchmark. In this experiment, we use the \textbf{left/right accuracy} as evaluation metric (see Sec.~\ref{sec:eval_metrics} for more details) and compare all baselines presented in Sec.~\ref{sec:baselines}. We present the results on our benchmark in Table~\ref{tab:chirality_benchmark}.
Similar to the dense correspondence prediction, the supervised methods perform best in this task. While unsupervised methods perform well in full-to-full shape matching, in the partial-to-full case, they perform poorly due to the XYZ coordinates as input (shapes are by default randomly rotated around the $x$-axis in our benchmark). Similarly, the partial-to-partial methods perform unconvincingly.

\input{tab/tab_chirality}

%% file: tab/tab_benchmark_list_methods.tex
\begin{table}[tbh!]
    \centering
    \footnotesize
    \begin{tabular}{@{}l@{}cccccc@{}}
    \toprule
    \multirow{2}{*}{Methods} & \multirow{2}{*}{Type} & \multirow{2}{*}{F2F} & \multirow{2}{*}{P2F} & \multirow{2}{*}{P2P} & Input         
    \\ 
            &      &     &     &     & features & 
            \\ 
    \midrule
    GeomFMaps~\cite{donati2020deepGeoMaps} & \textit{Sup} & \cmark & \cmark & \xmark & wks, xyz\\
    DPFM~\cite{attaiki2021dpfm} & \textit{Sup/Unsup} & \cmark & \cmark  & \cmark & wks, xyz  \\
    ULRSSM~\cite{cao2023unsupervised} & \textit{Unsup} & \cmark & \cmark & \xmark & wks, xyz\\
    Smooth Shells~\cite{eisenberger2020} & \textit{Axio} & \cmark & \xmark & \xmark & shot \\
    PFM~\cite{rodola2017partial} & \textit{Axio} & \xmark & \cmark  & \xmark & shot\\
    SM-COMB~\cite{roetzer2022scalable} & \textit{Axio} & \xmark & \xmark  & \cmark  & xyz\\
    GC-PPSM ~\cite{ehm2024partial} & \textit{Axio} & \xmark & \xmark  & \cmark & xyz\\
    \bottomrule
    \end{tabular}
    \caption{\textbf{Overview of baselines} with their methodology type (\textbf{Sup}ervised, \textbf{Unsup}ervised or \textbf{Axio}matic), their possible problem statements (F2F:~Full-to-Full, P2F:~Partial-to-Full and P2P:~Partial-to-Partial) and their corresponding input features. All default values are used for all baselines to produce the results shown in this paper.}
    \label{tab:benchmarks}
\end{table}

%% file: vis/qualitative_results_all/qualitative_all_double_column.tex
\begin{figure*}[!ht]
    \centering
    \begin{tabular}{cccccccccccc} \toprule
    \setlength{\tabcolsep}{0pt}  %

        \multirow{4}{*}[-5.5em]{\rotatebox[origin=c]{90}{\normalsize{\textit{Full-to-Full (F2F)}}}} & 
        \small{Source} &
        \small{GT} &
        \small{GeomFMaps} &
        \small{ULRSSM} &
        \small{Smooth Shells}\\

        &
        \adjustbox{valign=m}{\includegraphics[height=0.06\textheight]{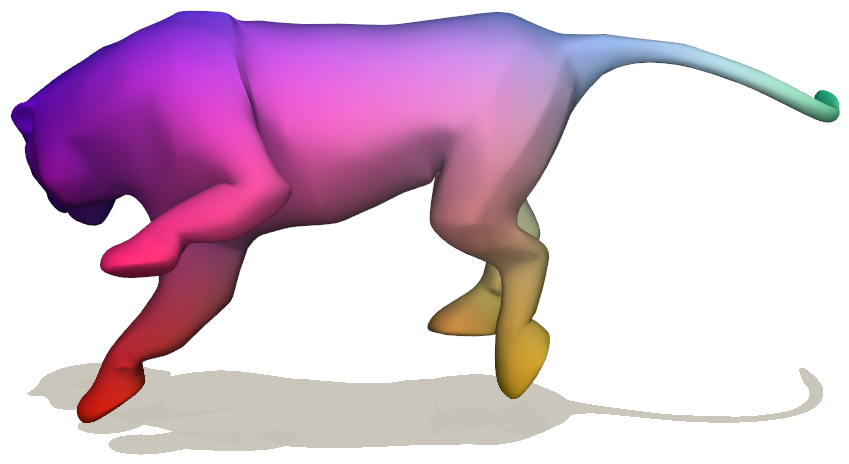}} &
        \adjustbox{valign=m}{\includegraphics[width=0.11\textwidth]{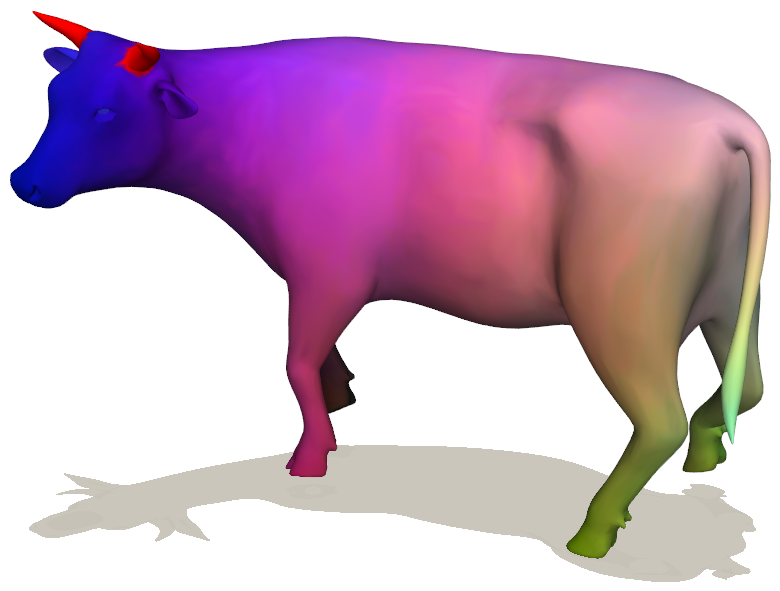}} &
        \adjustbox{valign=m}{\includegraphics[width=0.11\textwidth]{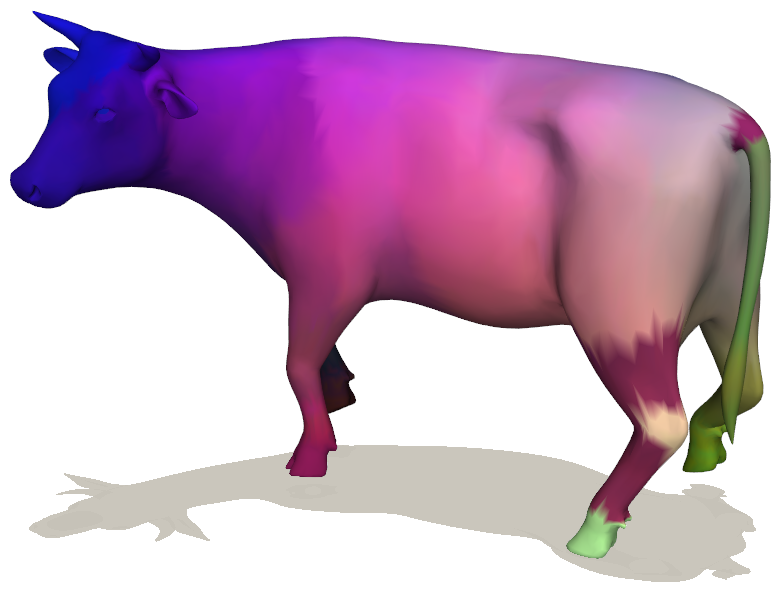}} &
        \adjustbox{valign=m}{\includegraphics[width=0.11\textwidth]{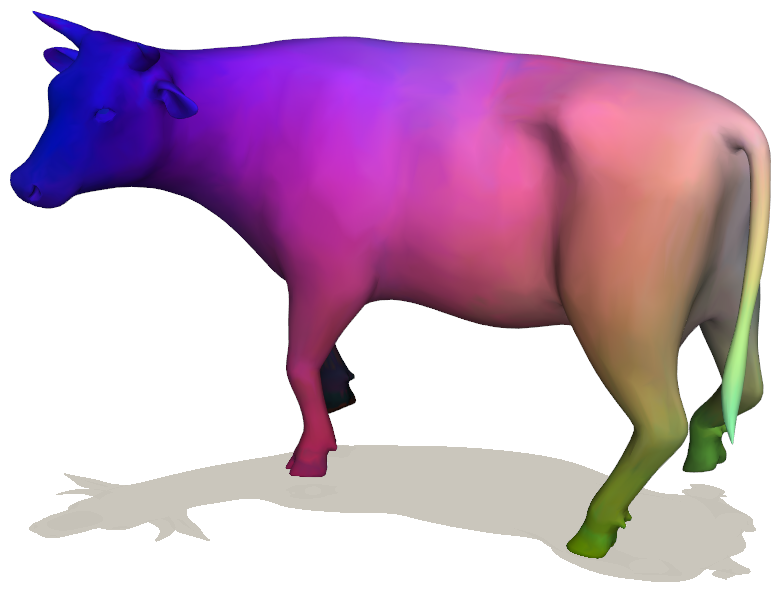}}  &
        \adjustbox{valign=m}{\includegraphics[width=0.11\textwidth]{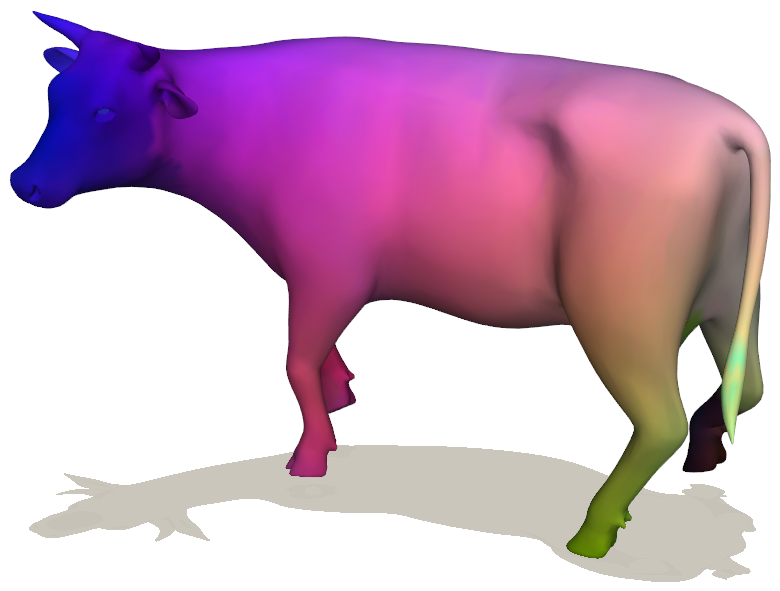}} \\

        &
        \adjustbox{valign=m}{\includegraphics[width=0.125\textwidth]{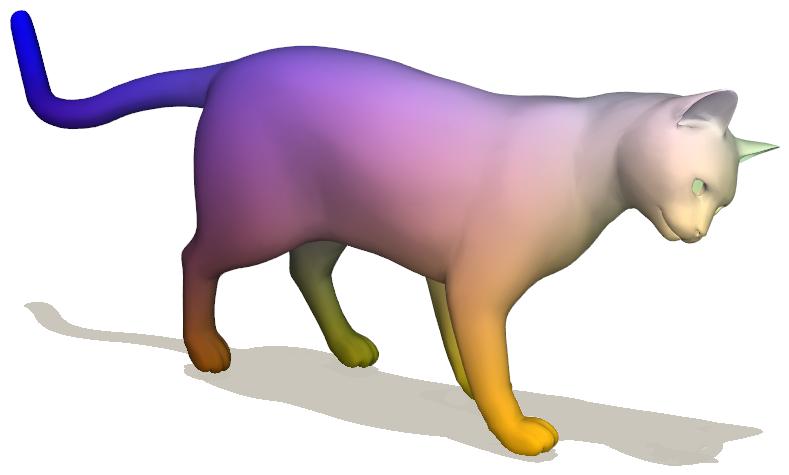}} &
        \adjustbox{valign=m}{\includegraphics[width=0.11\textwidth]{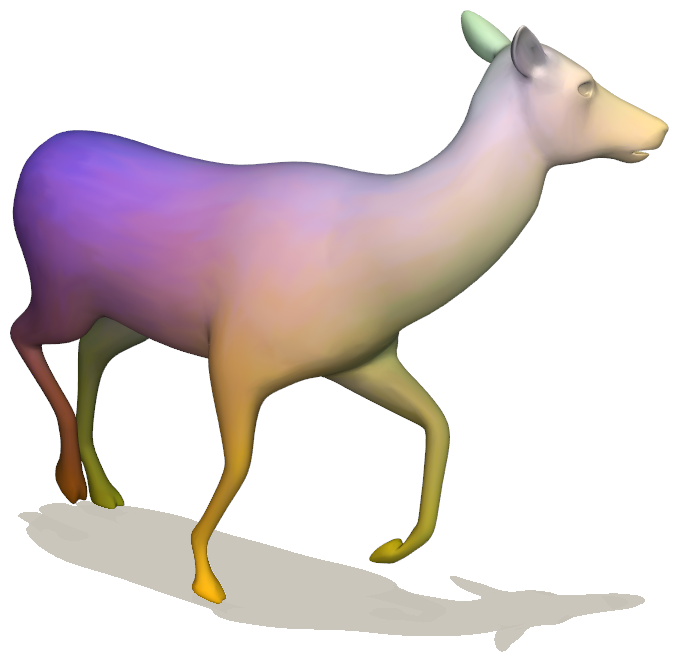}} &
        \adjustbox{valign=m}{\includegraphics[width=0.11\textwidth]{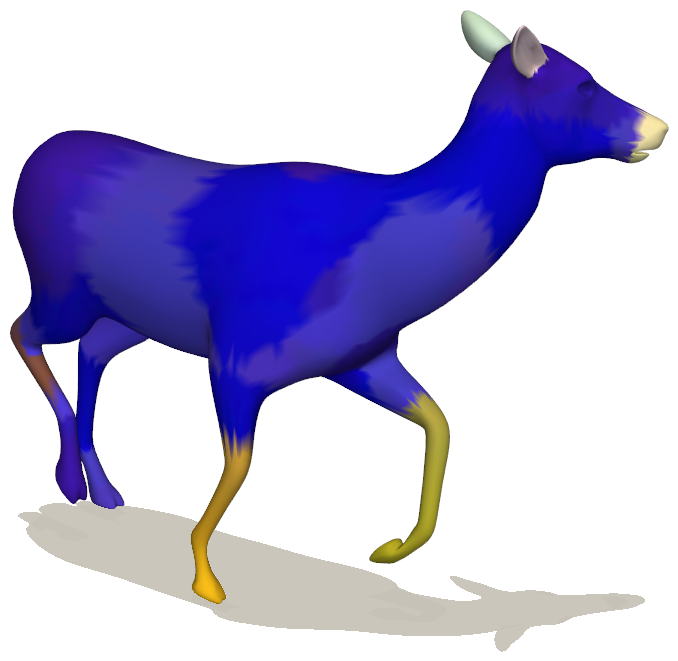}} &     
        \adjustbox{valign=m}{
            \begin{overpic}[width=0.11\textwidth]{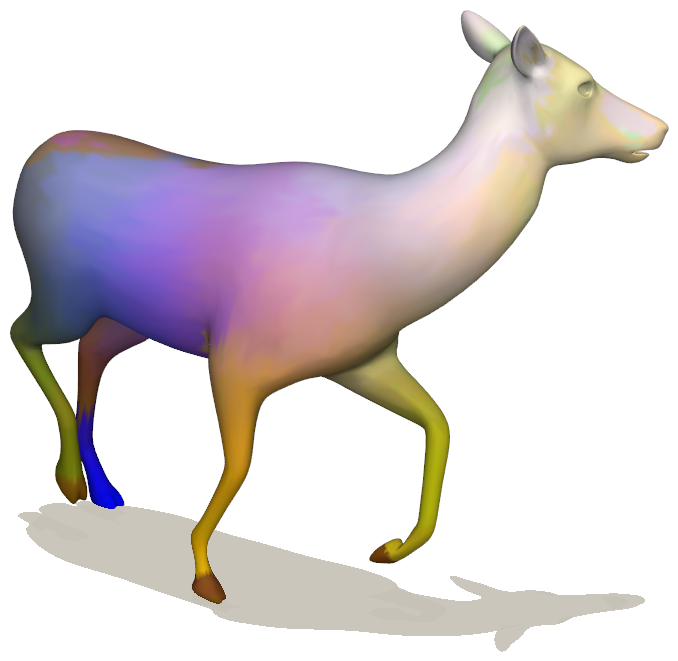}
                \put(-3, 15){\includegraphics[width=0.6cm]{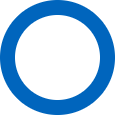}}
            \end{overpic}   
        } 
        &
        \adjustbox{valign=m}{
            \begin{overpic}[width=0.11\textwidth]{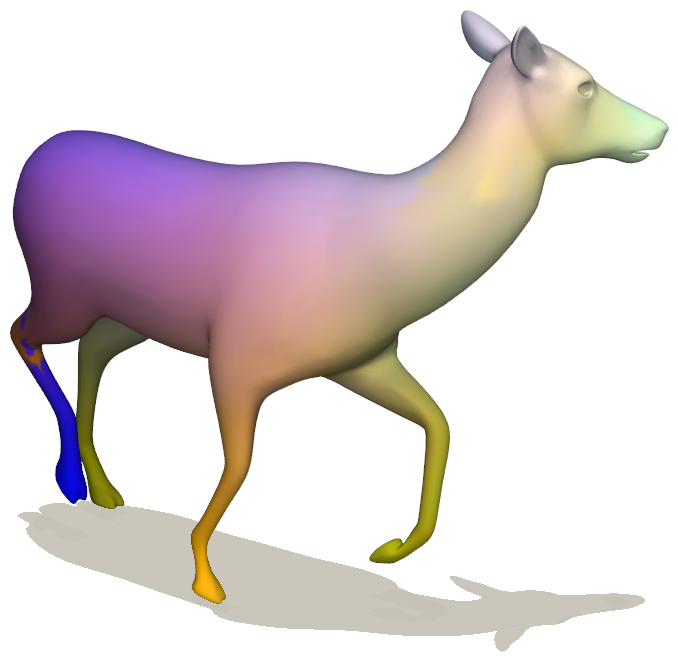}
                \put(-3, 15){\includegraphics[width=0.6cm]{vis/blue_circle.png}}
            \end{overpic}   
        } \\

        &
        \adjustbox{valign=m}{\includegraphics[height=0.1\textheight]{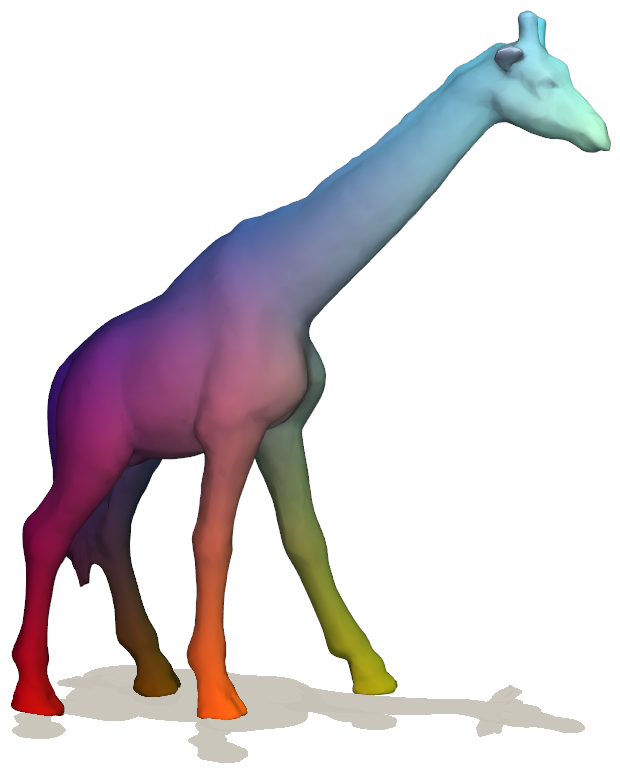}} &
        \adjustbox{valign=m}{\includegraphics[height=0.08\textheight]{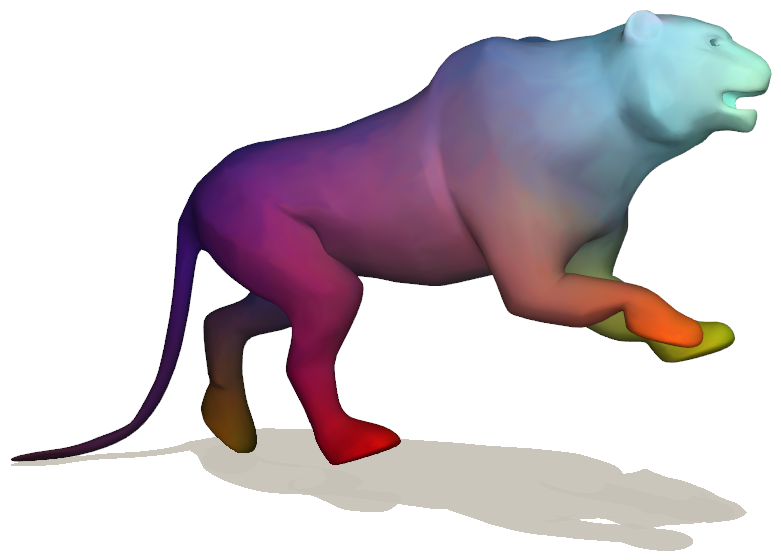}} &
        \adjustbox{valign=m}{
            \begin{overpic}[height=0.08\textheight]{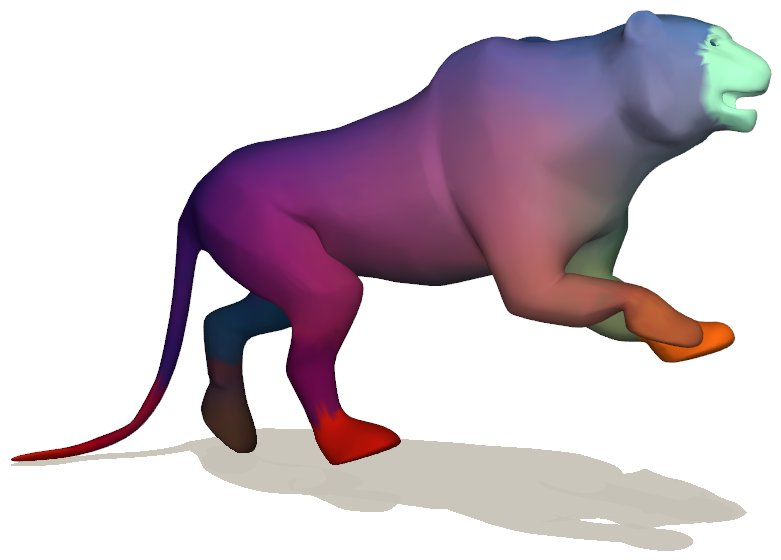}
                \put(80, 50){\includegraphics[width=0.6cm]{vis/blue_circle.png}}
            \end{overpic}   
        } &
        \adjustbox{valign=m}{\includegraphics[height=0.08\textheight]{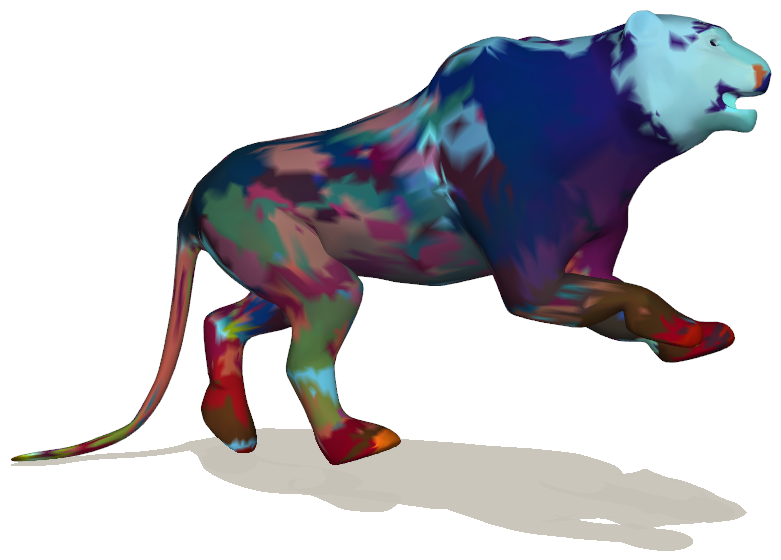}}  &
        \adjustbox{valign=m}{\includegraphics[height=0.08\textheight]{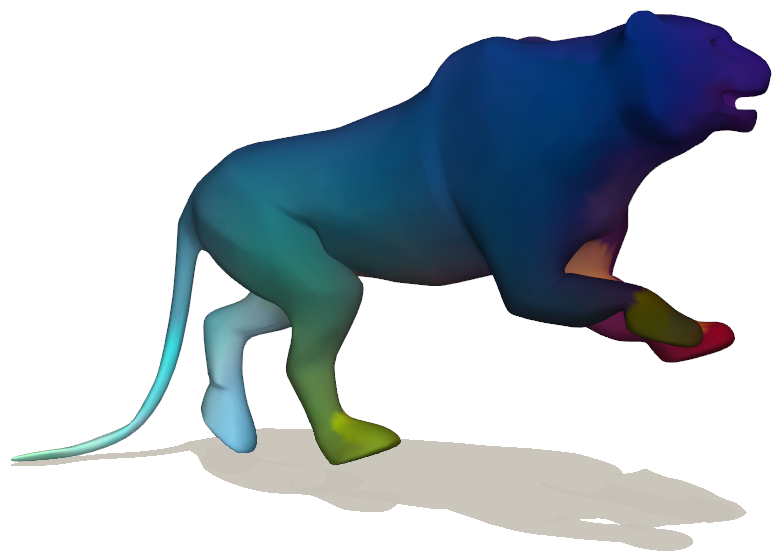}} \\

        \midrule
        \multirow{4}{*}[-5.5em]{\rotatebox[origin=c]{90}{\normalsize{\textit{Partial-to-Full (P2F)}}}} & 
        \small{Source} &
        \small{GT} &
        \small{DPFM~\tiny{(\textit{Unsup})}} &
        \small{ULRSSM} &
        \small{PFM} \\

        &
        \adjustbox{valign=m}{\includegraphics[width=0.15\textwidth]{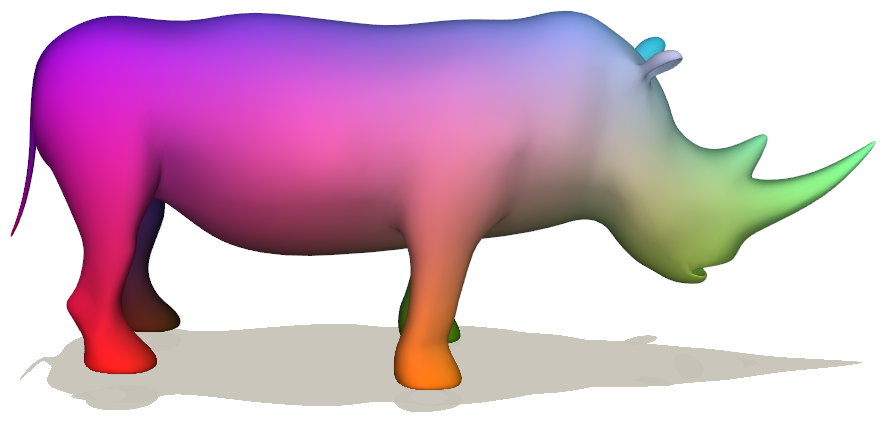}} &
        \adjustbox{valign=m}{\includegraphics[width=0.13\textwidth]{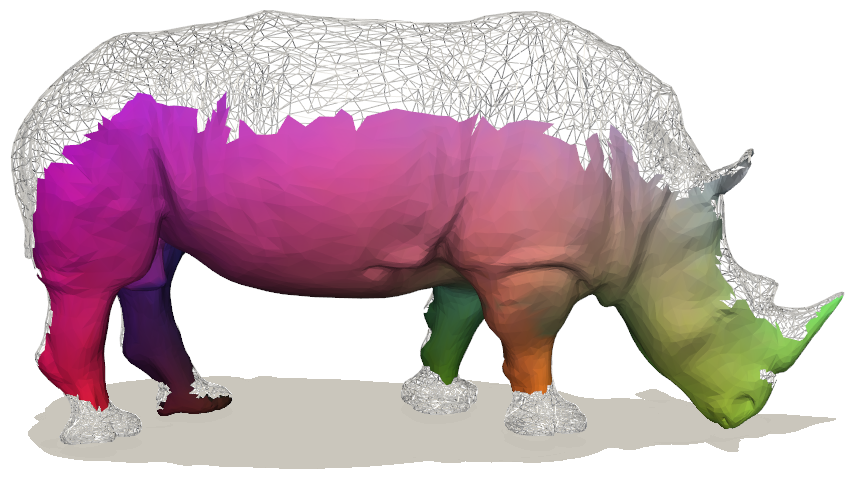}} &
        \adjustbox{valign=m}{\includegraphics[width=0.13\textwidth]{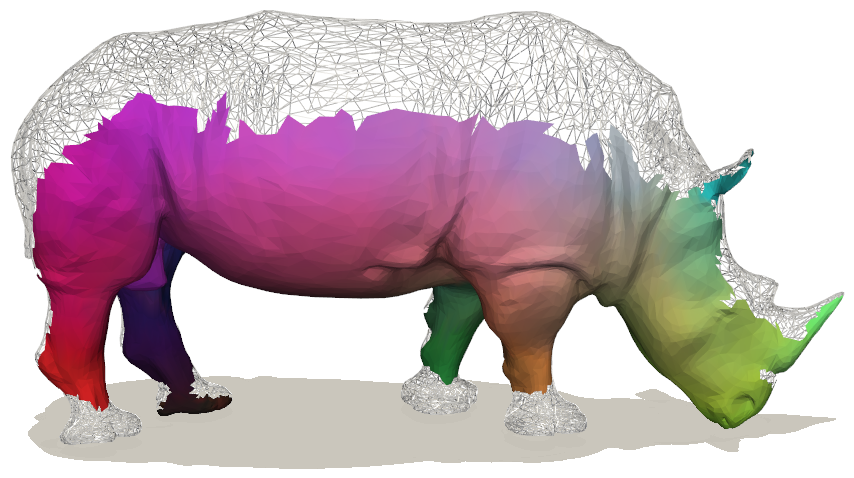}} &
        \adjustbox{valign=m}{\includegraphics[width=0.13\textwidth]{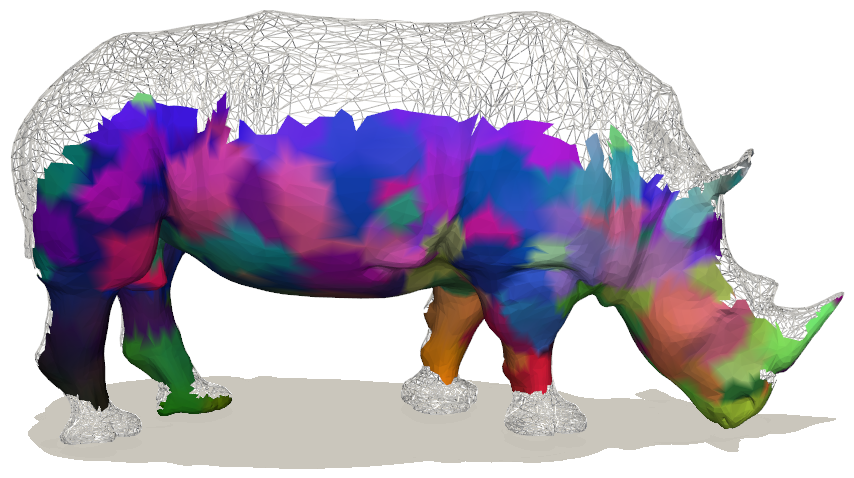}} &
        \adjustbox{valign=m}{\includegraphics[width=0.13\textwidth]{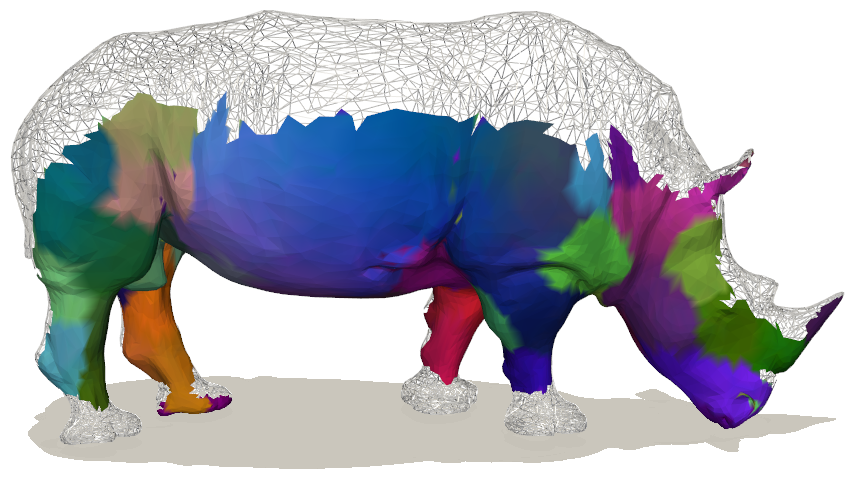}}\\

        &
        \adjustbox{valign=m}{\includegraphics[height=0.12\textheight]{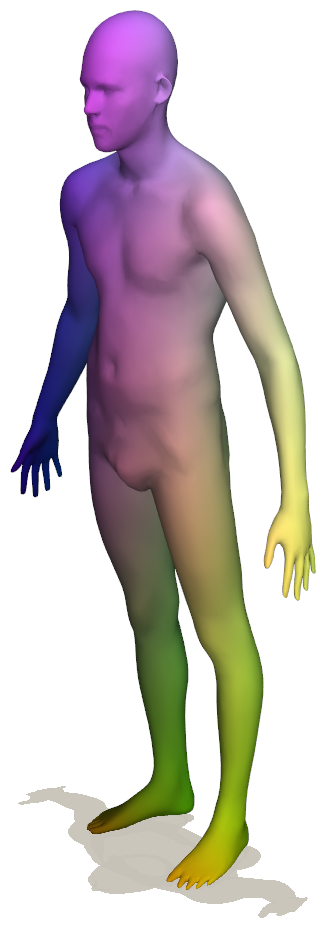}} &
        \adjustbox{valign=m}{\includegraphics[height=0.12\textheight]{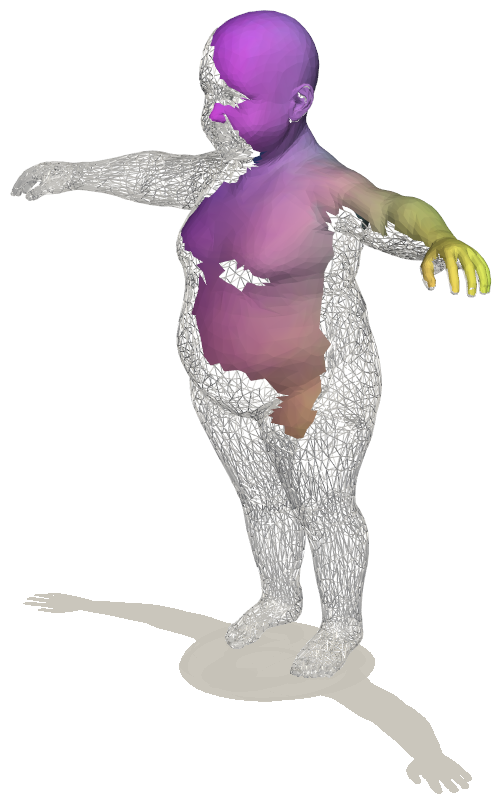}} &
        \adjustbox{valign=m}{\includegraphics[height=0.12\textheight]{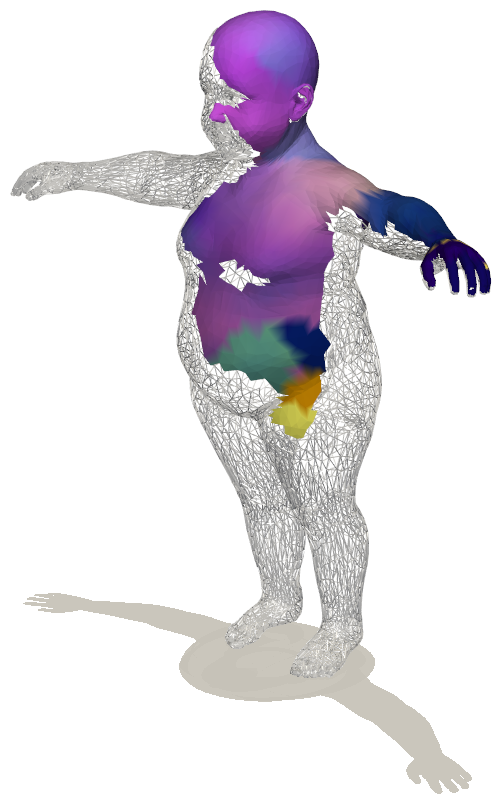}} &
        \adjustbox{valign=m}{\includegraphics[height=0.12\textheight]{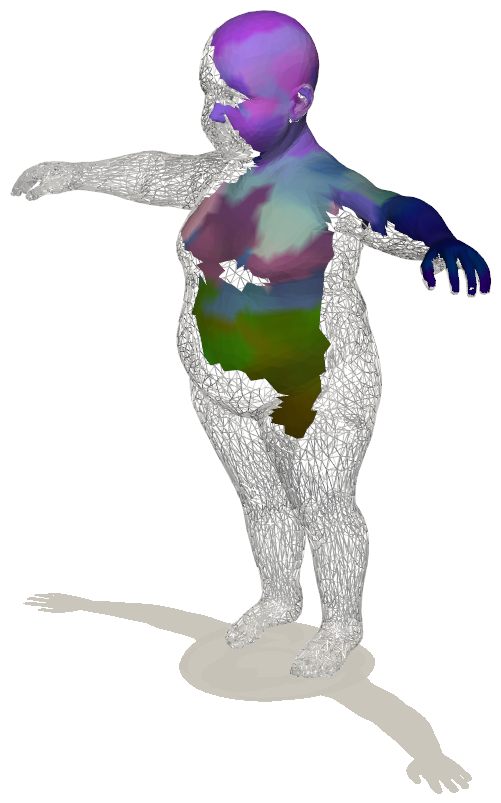}} &
        \adjustbox{valign=m}{\includegraphics[height=0.12\textheight]{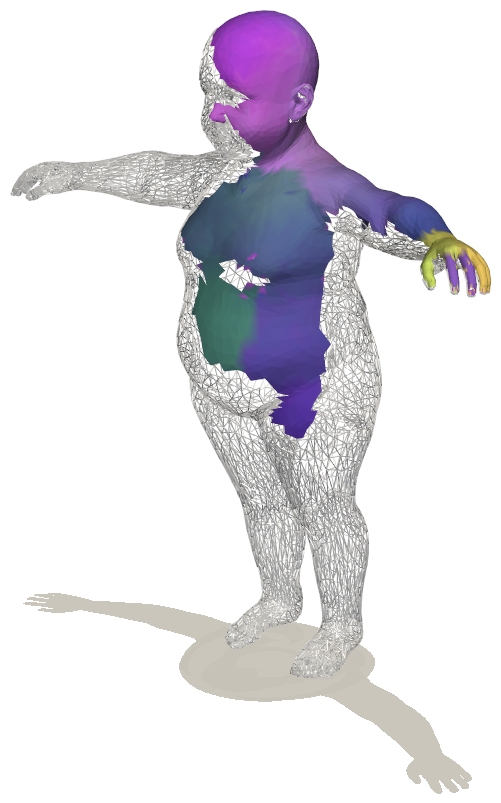}}\\

        &
        \adjustbox{valign=m}{\includegraphics[height=0.08\textheight]{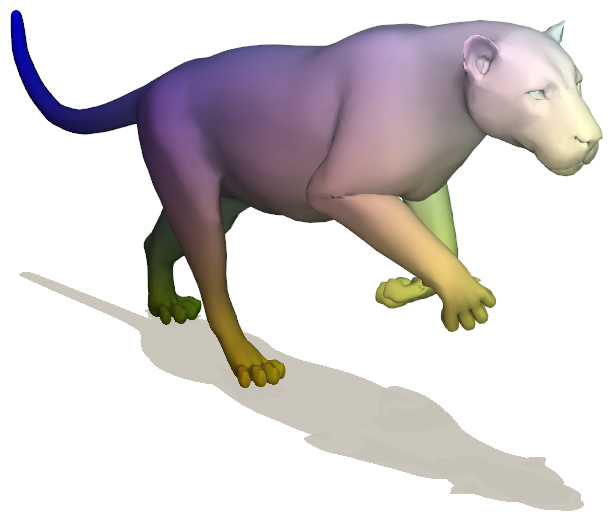}} &
        \adjustbox{valign=m}{\includegraphics[width=0.1\textwidth]{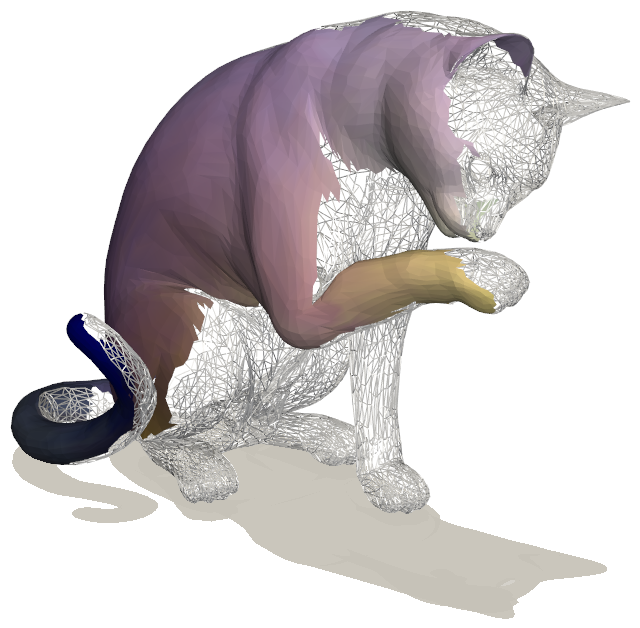}} &
        \adjustbox{valign=m}{\includegraphics[width=0.1\textwidth]{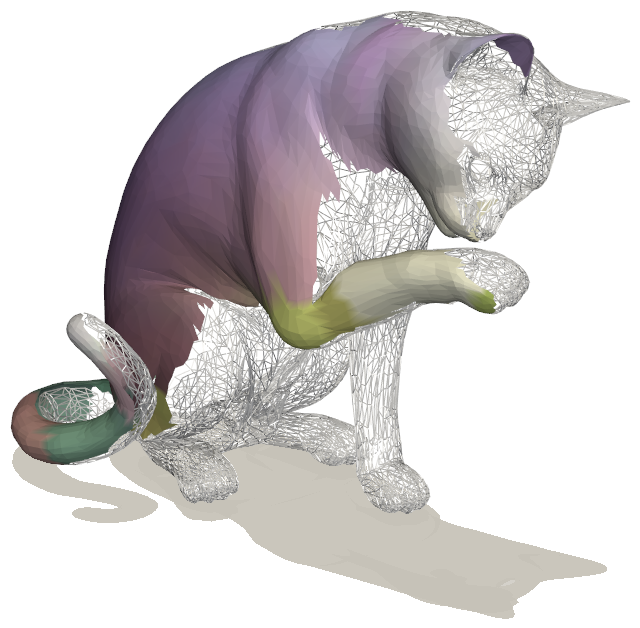}} &
        \adjustbox{valign=m}{\includegraphics[width=0.1\textwidth]{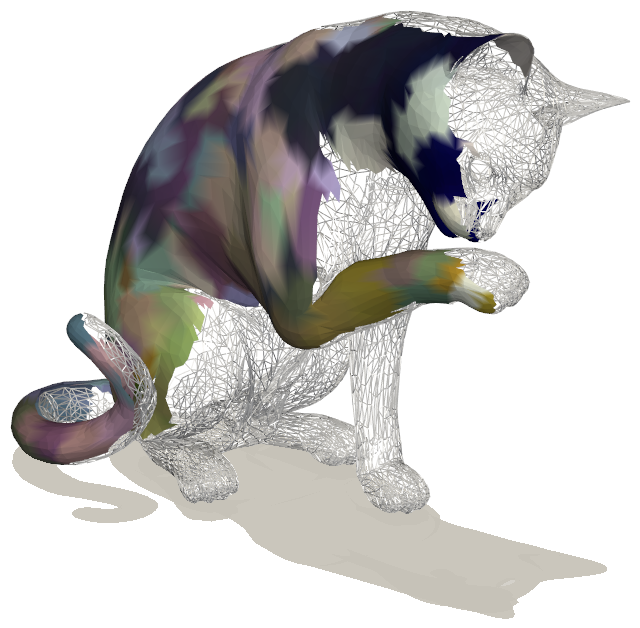}} &
        \adjustbox{valign=m}{\includegraphics[width=0.1\textwidth]{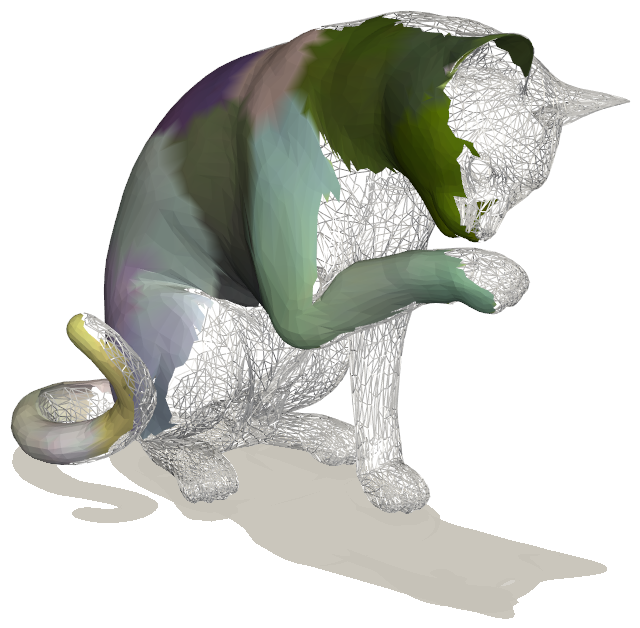}}\\

        \midrule

        \multirow{4}{*}[-5.5em]{\rotatebox[origin=c]{90}{\normalsize{\textit{Partial-to-Partial (P2P)}}}} & 
        \small{Source} &
        \small{GT} &
        \small{DPFM~\tiny{\textit{(Sup)}}} &
        \small{SM-COMB} &
        \small{GC-PPSM} \\

        &
        \adjustbox{valign=m}{\includegraphics[height=0.08\textheight]{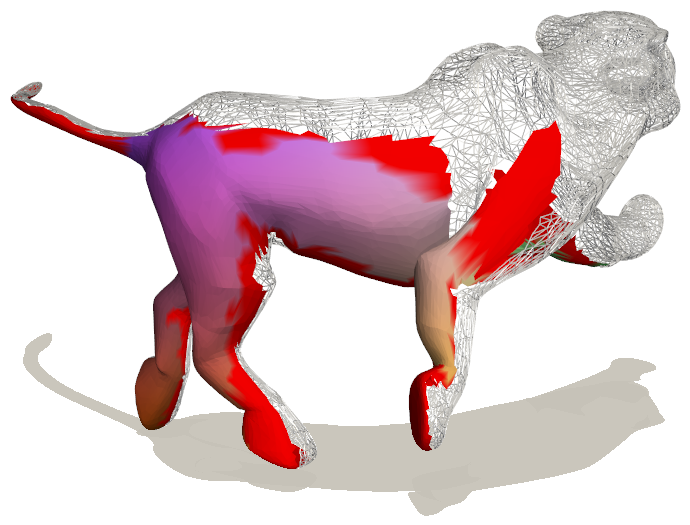}} &
        \adjustbox{valign=m}{\includegraphics[height=0.08\textheight]{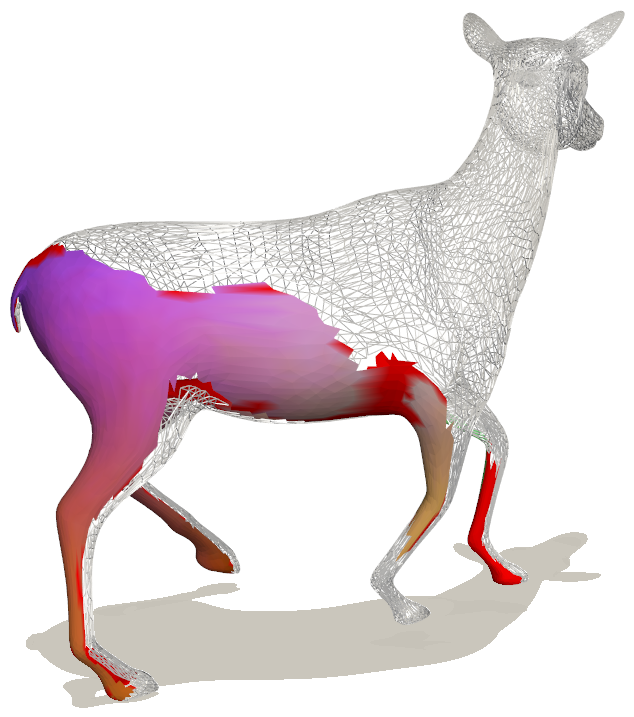}} &
        \adjustbox{valign=m}{
            \begin{overpic}[height=0.08\textheight]{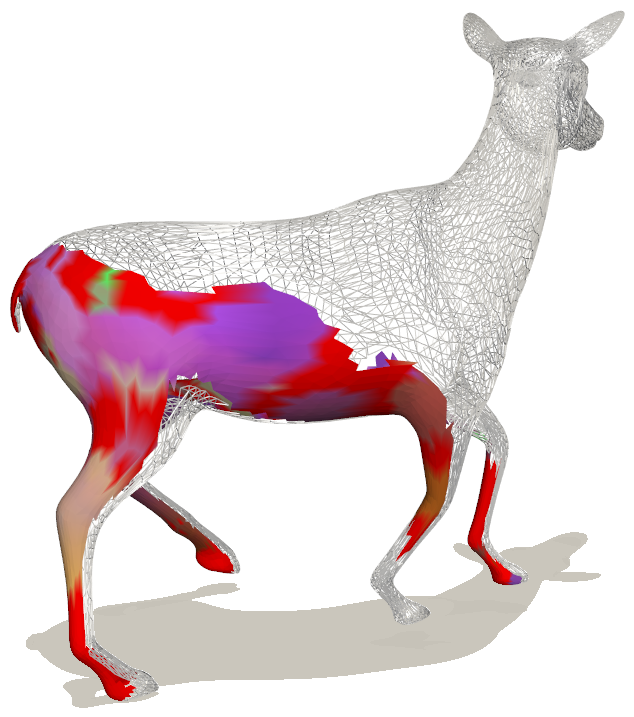}
                \put(0, 30){\includegraphics[width=0.6cm]{vis/blue_circle.png}}
            \end{overpic}    
        } &
        \adjustbox{valign=m}{\includegraphics[height=0.08\textheight]{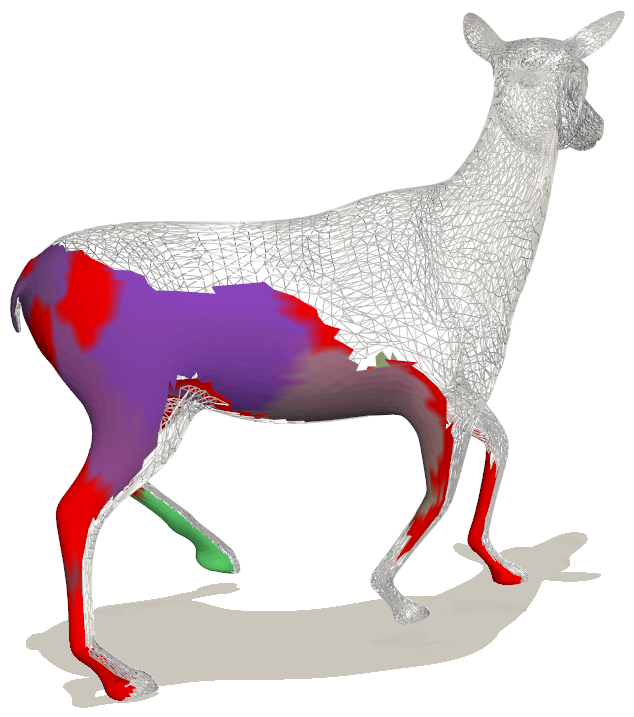}} &
        \adjustbox{valign=m}{\includegraphics[height=0.08\textheight]{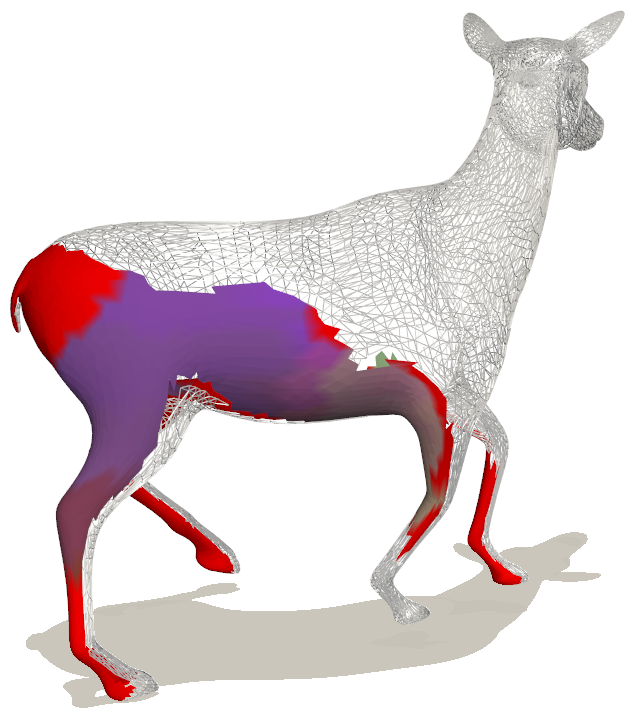}}  \\

        &
        \adjustbox{valign=m}{\includegraphics[height=0.1\textheight]{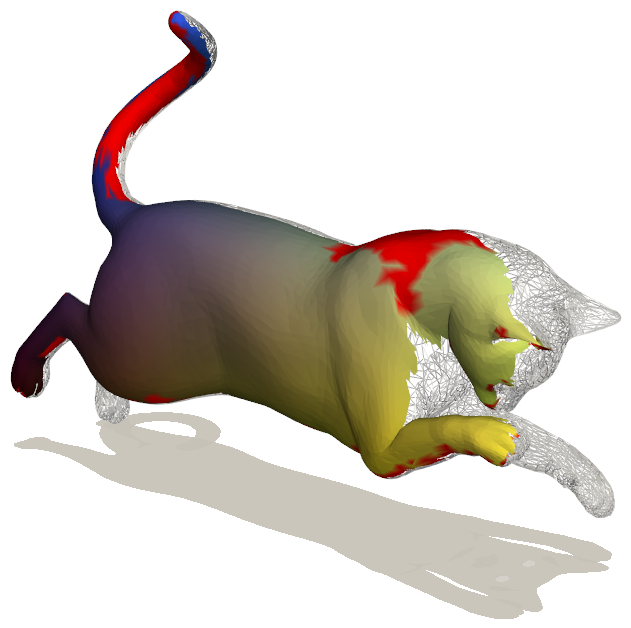}} &
        \adjustbox{valign=m}{\includegraphics[height=0.08\textheight]{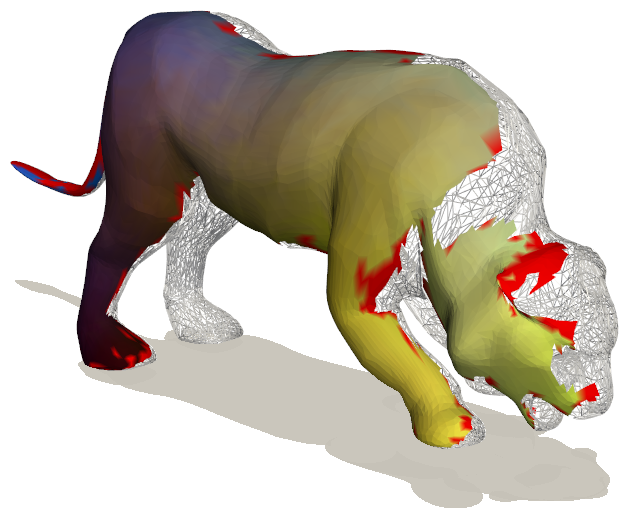}} &
        \adjustbox{valign=m}{\includegraphics[height=0.08\textheight]{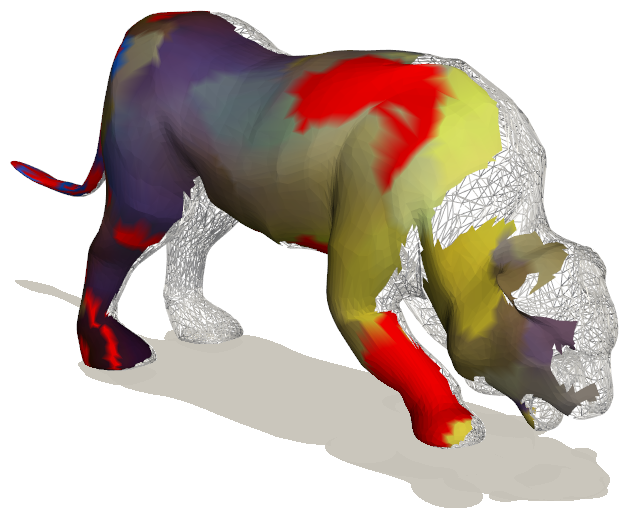}} &
        \adjustbox{valign=m}{\includegraphics[height=0.08\textheight]{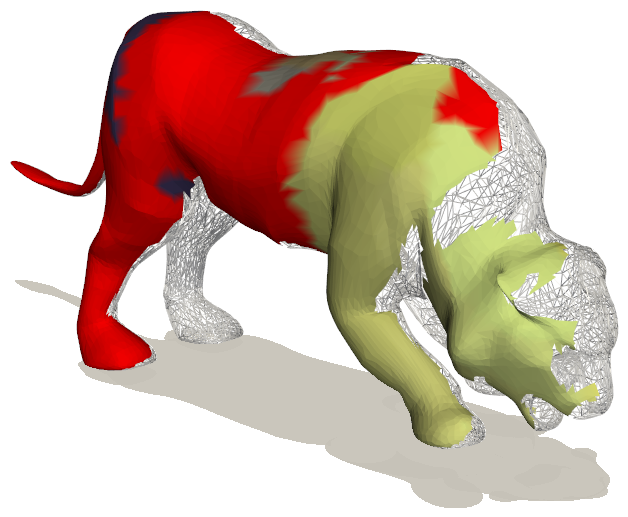}} &
        \adjustbox{valign=m}{\includegraphics[height=0.08\textheight]{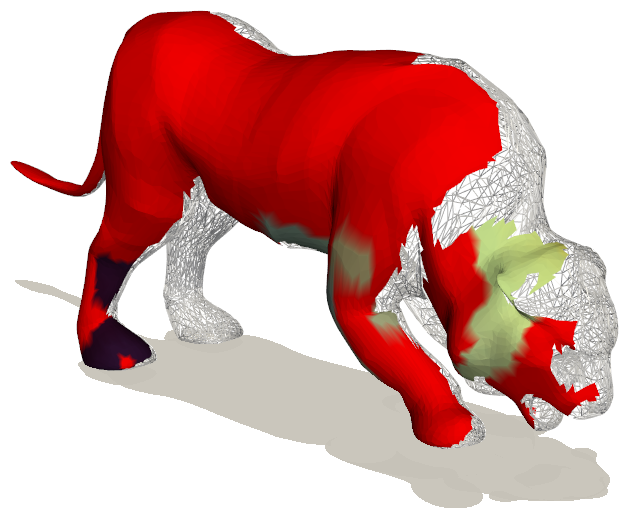}} \\

        &
        \adjustbox{valign=m}{\includegraphics[height=0.11\textheight]{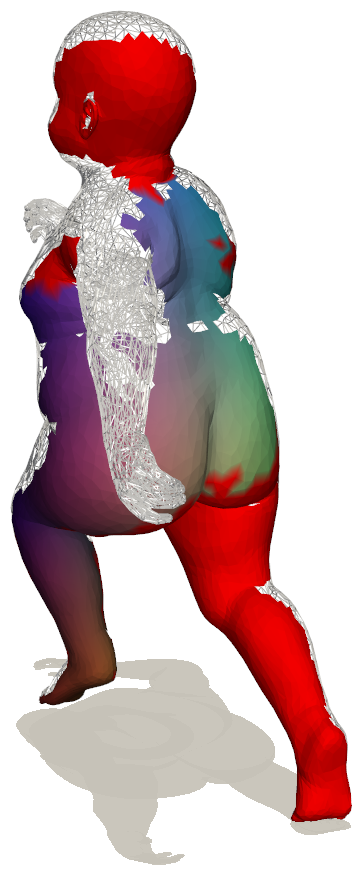}} &
        \adjustbox{valign=m}{\includegraphics[height=0.11\textheight]{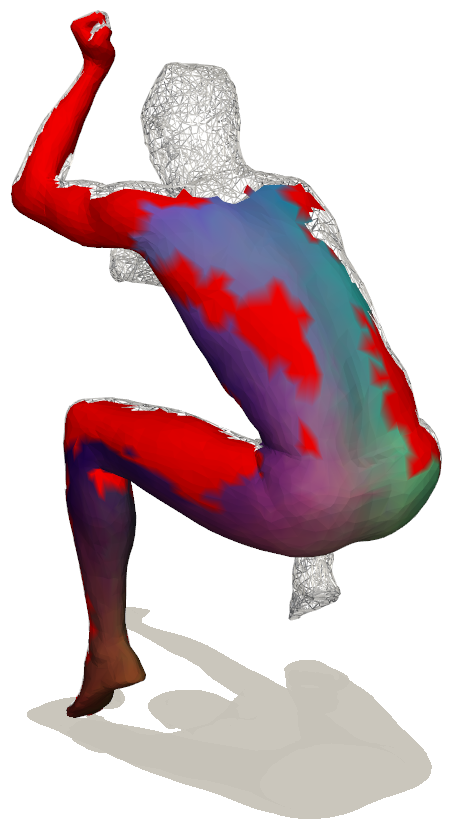}} &
        \adjustbox{valign=m}{\includegraphics[height=0.11\textheight]{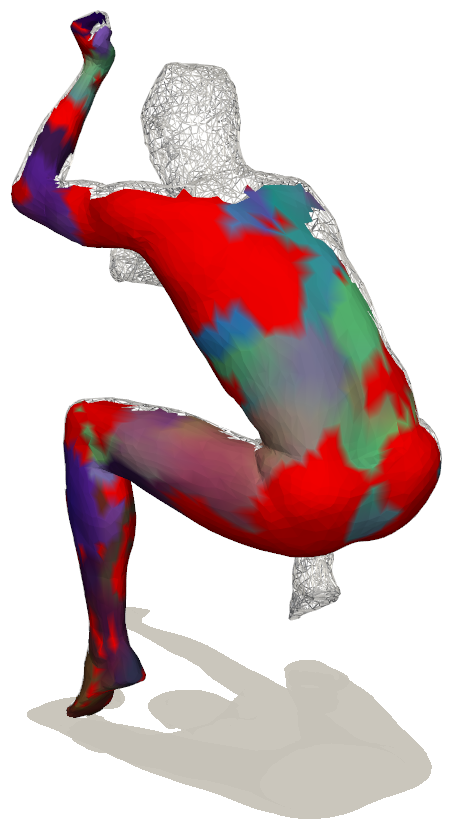}} &
        \adjustbox{valign=m}{\includegraphics[height=0.11\textheight]{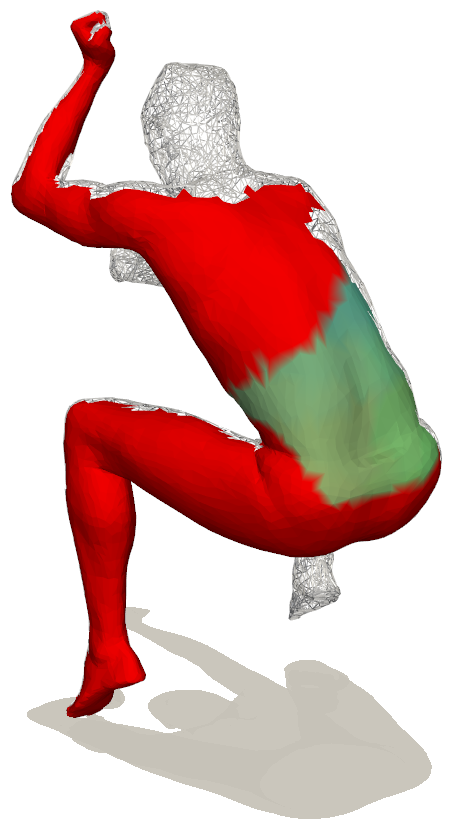}} &
        \adjustbox{valign=m}{\includegraphics[height=0.11\textheight]{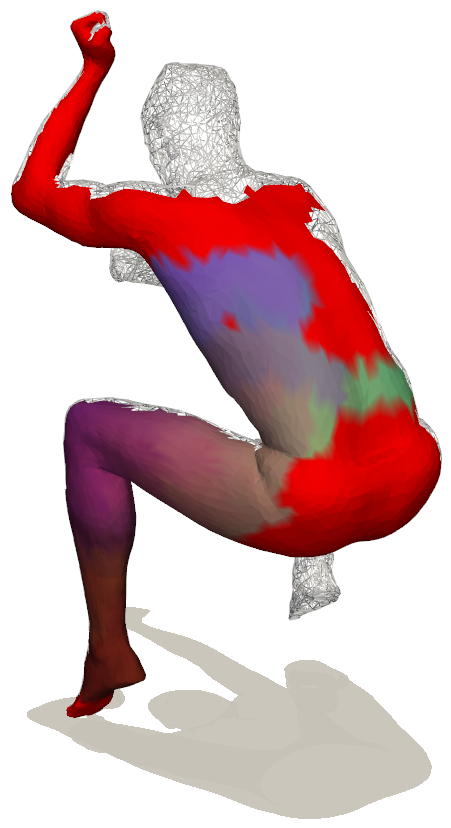}} \\

        \bottomrule

    \end{tabular}
    \caption{\textbf{Cross-dataset and cross-category shape matching} results visualised via colour transfer (red denotes missing parts).
    All baselines show sub-optimal performance on all three settings (some examples are highlighted with blue circles).
    The green square shows the semantic missing parts between the source and target mesh.
    As GC-PPSM and SM-COMB are solved on very low resolution when upsampling, we do nearest neighbour search in the features space of the matched patches (see~\cite{ehm2024partial} for more information), ensuring geometric consistency in low resolution.
    Both methods show patch-wise inaccuracies due to this low resolution solution.
    }
    \label{fig:all_qualitative_results}
\end{figure*}

%% file: fig/fig_benchmark_results.tex
\begin{figure*}[ht]%
    \centering
    \begin{tabular}{@{}c@{}@{}c@{}@{}c@{}}
    {\input{vis/tikz_vis/full_to_full}}&  {\input{vis/tikz_vis/partial_full}} & {\input{vis/tikz_vis/partial_partial}}
    \end{tabular}
\caption{We show \textbf{geodesic errors (AUC x100)} for the different methods on different settings: F2F~(left), P2F~(middle), P2P~(right). The partial-to-partial setting is the most difficult and more realistic case, in which errors are computed for \emph{all} vertices. 
Unmatched vertices are set to an infinite geodesic error; matchings that predict wrong overlaps lead to curves that do not reach 100\%.
}
\label{fig:geo_errors}
\end{figure*}

%% file: vis/tikz_vis/full_to_full.tex
\newcommand{\pckLineWidth}{3pt}
\newcommand{\plotWidth}{1*\linewidth}
\newcommand{\plotHeight}{0.65*\linewidth}
\newcommand{\pckTitle}{\textsc{Full-to-Full (F2F)}}

\pgfplotsset{%
    label style = {font=\Huge},
    tick label style = {font=\huge},
    title style =  {font=\Huge},
    legend style={  fill= gray!10,
                    fill opacity=0.6, 
                    font=\Huge,
                    draw=gray!20, %
                    text opacity=1}
}
\begin{tikzpicture}[scale=0.29, transform shape]
	\begin{axis}[
		width=\plotWidth,
		height=\plotHeight,
		grid=major,
		title=\pckTitle,
		legend style={
			at={(0.99,0.01)},
			anchor=south east,
			legend columns=1},
		legend cell align={left},
	ylabel={{\huge
        $\%$ Samples $<$ Geo Dist Thresh}},
        xmin=0,
        xmax=1.0,
        xlabel=\huge
        Geodesic Distance Threshold,
        ylabel near ticks,
        xtick={0, 0.2, 0.4, 0.6, 0.8, 1}, %
        xticklabels={$0$, $0.2$, $0.4$, $0.6$, $0.8$, $1$}, %
        ymin=0,
        ymax=100,
        ytick={0, 20, 40, 60, 80, 100},
        yticklabels={$0$, $20$, $40$, $60$, $80$, $100$},
	]

\addplot [color=cPLOT3, solid, smooth, line width=\pckLineWidth]  
table[row sep=crcr]{%
0.0  2.500351204323409\\
0.01  6.285197287398417\\
0.02  16.44713135936009\\
0.03  27.952883098706486\\
0.04  39.166716040218766\\
0.05  49.38842005751181\\
0.06  58.442635198759284\\
0.07  66.17676924068694\\
0.08  72.53537475271304\\
0.09  77.67339153077761\\
0.1  81.75664018381121\\
0.11  84.95999624388374\\
0.12  87.39163436891099\\
0.13  89.16774077431889\\
0.14  90.44299441464021\\
0.15  91.32694868094363\\
0.16  91.95993625036006\\
0.17  92.38581574237514\\
0.18  92.69345955076402\\
0.19  92.91852842221132\\
0.2  93.08695744521492\\
0.21  93.2172663596936\\
0.22  93.32313070800183\\
0.23  93.39929639893532\\
0.24  93.45932718567398\\
0.25  93.51279967417182\\
0.26  93.5605336517089\\
0.27  93.61039162356263\\
0.28  93.66043591070728\\
0.29  93.70518884358943\\
0.3  93.75907122572731\\
0.31  93.82357357944838\\
0.32  93.8938144441302\\
0.33  93.97970579325039\\
0.34  94.08247730772914\\
0.35  94.16568571665991\\
0.36  94.25399916456223\\
0.37  94.3280035981209\\
0.38  94.41255347714637\\
0.39  94.50127681868877\\
0.4  94.59771361327579\\
0.41  94.69608808688851\\
0.42  94.79397814074481\\
0.43  94.88467642437108\\
0.44  94.97168566523688\\
0.45  95.05675722707699\\
0.46  95.13609027795633\\
0.47  95.21013197457317\\
0.48  95.280372839255\\
0.49  95.35191791097334\\
0.5  95.41928952017477\\
0.51  95.47708452342225\\
0.52  95.53868035860476\\
0.53  95.59859935616888\\
0.54  95.64953795671002\\
0.55  95.69730919730529\\
0.56  95.74664548634439\\
0.57  95.8082785845851\\
0.58  95.86134117944286\\
0.59  95.91242883221675\\
0.6  95.97454635021387\\
0.61  96.02369632396204\\
0.62  96.07478397673592\\
0.63  96.12713857348815\\
0.64  96.18377842193185\\
0.65  96.23385997213468\\
0.66  96.28614004277054\\
0.67  96.3447920963563\\
0.68  96.4086982411464\\
0.69  96.47074123302714\\
0.7  96.54198820027999\\
0.71  96.62091135751929\\
0.72  96.69774778350015\\
0.73  96.77491957700467\\
0.74  96.8545879954077\\
0.75  96.9351879902654\\
0.76  97.01921618647627\\
0.77  97.09314609391856\\
0.78  97.16394590447317\\
0.79  97.22781478620507\\
0.8  97.28132453776108\\
0.81  97.32745620379613\\
0.82  97.39192129445901\\
0.83  97.44684704222612\\
0.84  97.49469280893777\\
0.85  97.54965581976306\\
0.86  97.60756261218509\\
0.87  97.66647550717818\\
0.88  97.72367430149465\\
0.89  97.77494826955947\\
0.9  97.82555150257693\\
0.91  97.87962020000573\\
0.92  97.93342805602724\\
0.93  97.98440391962656\\
0.94  98.03914335210273\\
0.95  98.09980761083058\\
0.96  98.15175231394275\\
0.97  98.19956081759621\\
0.98  98.23402914641886\\
0.99  98.27621092828612\\
1.0  98.31123820298157\\
};
\addlegendentry{\textcolor{black}{GeomFmaps \huge{(\textit{Sup})}~\Huge{(\textbf{88.72})}  }}

\addplot [color=cPLOT4, solid, smooth, line width=\pckLineWidth]  
table[row sep=crcr]{%
0.0  3.325578890924694\\
0.01  8.761886449773309\\
0.02  21.339659110091098\\
0.03  34.32233216066489\\
0.04  46.34454988647809\\
0.05  56.76535042366234\\
0.06  65.57400201145988\\
0.07  72.86336419087182\\
0.08  78.56647524633676\\
0.09  82.83316993482318\\
0.1  85.99028253968609\\
0.11  88.28210967039706\\
0.12  89.94631511207051\\
0.13  91.12278438513904\\
0.14  91.95002427688242\\
0.15  92.53408545089977\\
0.16  92.96831186794866\\
0.17  93.29555604494372\\
0.18  93.55919218161418\\
0.19  93.77986401219546\\
0.2  93.96524772667398\\
0.21  94.11467259000239\\
0.22  94.23547942464347\\
0.23  94.33992777674062\\
0.24  94.42131029582023\\
0.25  94.49110400380381\\
0.26  94.5494206898659\\
0.27  94.60032202734885\\
0.28  94.64988189473708\\
0.29  94.69649798052856\\
0.3  94.74423195806565\\
0.31  94.78999099351884\\
0.32  94.82822289121833\\
0.33  94.86190869581905\\
0.34  94.89592986794345\\
0.35  94.93229861273362\\
0.36  94.9716484021787\\
0.37  95.0095449323545\\
0.38  95.04673346442478\\
0.39  95.0828786308658\\
0.4  95.12438967768573\\
0.41  95.16862092775327\\
0.42  95.21717469257045\\
0.43  95.26997644602092\\
0.44  95.32806955373388\\
0.45  95.38593908309774\\
0.46  95.44436755833439\\
0.47  95.50447287118944\\
0.48  95.56811817457222\\
0.49  95.63280684358422\\
0.5  95.68955848120247\\
0.51  95.74716716915901\\
0.52  95.80816679541053\\
0.53  95.86726600569455\\
0.54  95.92979341733172\\
0.55  95.99295430095806\\
0.56  96.0550718189552\\
0.57  96.11759923059238\\
0.58  96.16965572287911\\
0.59  96.22145137375854\\
0.6  96.27116029337951\\
0.61  96.32545256915745\\
0.62  96.38671303681627\\
0.63  96.45490443329781\\
0.64  96.5400505212543\\
0.65  96.63242564249896\\
0.66  96.71421805521864\\
0.67  96.80409655156481\\
0.68  96.89304347145631\\
0.69  96.9851204882355\\
0.7  97.05849144980499\\
0.71  97.12713000298477\\
0.72  97.19513508417539\\
0.73  97.26910225467587\\
0.74  97.33304566252413\\
0.75  97.38618278349827\\
0.76  97.45247376401231\\
0.77  97.51910011205001\\
0.78  97.58822308498623\\
0.79  97.65790500379524\\
0.8  97.72766144872062\\
0.81  97.80043620135913\\
0.82  97.87533494831428\\
0.83  97.95403452720446\\
0.84  98.04450923248162\\
0.85  98.11970608390224\\
0.86  98.19698966658133\\
0.87  98.27416146008586\\
0.88  98.34678716049162\\
0.89  98.42079159405029\\
0.9  98.49233666576863\\
0.91  98.54416957970625\\
0.92  98.59376671015265\\
0.93  98.63606028119449\\
0.94  98.68085047713483\\
0.95  98.72850992855554\\
0.96  98.76983466008453\\
0.97  98.80460109337265\\
0.98  98.83105786468516\\
0.99  98.85699295318307\\
1.0  98.8792017358623\\
1.01  98.90517408741839\\
1.02  98.93420200974577\\
1.03  98.9631554059568\\
1.04  98.99754920866306\\
1.05  99.03499858214063\\
1.06  99.06812544086856\\
1.07  99.10210934993478\\
1.08  99.13150990284403\\
1.09  99.16538202273568\\
1.1  99.20570065169363\\
1.11  99.2473234876881\\
1.12  99.29252357726848\\
1.13  99.34230702300584\\
1.14  99.3915315228704\\
1.15  99.44481769607728\\
1.16  99.4955327182693\\
1.17  99.54274501299176\\
1.18  99.58339900947338\\
1.19  99.62572984357341\\
1.2  99.66478152855301\\
1.21  99.69861638538647\\
1.22  99.72179400757855\\
1.23  99.74351837050138\\
1.24  99.76662146657709\\
1.25  99.79300371177322\\
1.26  99.81934869391118\\
1.27  99.84353241867431\\
1.28  99.86015174262555\\
1.29  99.87628664682036\\
1.3  99.8956634370774\\
1.31  99.91179834127223\\
1.32  99.92558567280129\\
1.33  99.93583301380261\\
1.34  99.946490248444\\
1.35  99.95479991041961\\
1.36  99.96392935967533\\
1.37  99.96970513369426\\
1.38  99.97525732936407\\
1.39  99.97816384790264\\
1.4  99.98174110148855\\
1.41  99.9821509951286\\
1.42  99.9821509951286\\
1.43  99.9821509951286\\
1.44  99.98222552124497\\
1.45  99.98226278430316\\
1.46  99.98226278430316\\
1.47  99.98226278430316\\
1.48  99.98226278430316\\
1.49  99.98226278430316\\
};
\addlegendentry{\textcolor{black}{DPFM \huge{(\textit{Sup})}~\Huge{(\textbf{90.02})}}}

\addplot [color=cPLOT5, solid, smooth, line width=\pckLineWidth]  
table[row sep=crcr]{%
0.0  1.8033084378841588\\
0.01  5.4132044627729\\
0.02  14.415437637850026\\
0.03  25.094433905209492\\
0.04  35.980277408562976\\
0.05  46.440427735192316\\
0.06  56.00156952001082\\
0.07  64.21621069725516\\
0.08  70.85086094432788\\
0.09  75.99521244228417\\
0.1  79.98426008422196\\
0.11  83.06494615674407\\
0.12  85.39545234185279\\
0.13  87.08052509611073\\
0.14  88.29515174076239\\
0.15  89.20504109556371\\
0.16  89.89496661788932\\
0.17  90.4299523442749\\
0.18  90.8423798722846\\
0.19  91.16820805306855\\
0.2  91.43016735212062\\
0.21  91.66548356456924\\
0.22  91.85641947471758\\
0.23  92.02693522897964\\
0.24  92.19845708581272\\
0.25  92.36662526740902\\
0.26  92.5312161954194\\
0.27  92.6990862725502\\
0.28  92.87221044088533\\
0.29  93.056625315851\\
0.3  93.21272026659483\\
0.31  93.35212136727104\\
0.32  93.4704688400718\\
0.33  93.59410766713506\\
0.34  93.71066651314287\\
0.35  93.83534870583536\\
0.36  93.96528498973217\\
0.37  94.08918465820274\\
0.38  94.2164380019101\\
0.39  94.34536818323586\\
0.4  94.44441339189595\\
0.41  94.54312323303236\\
0.42  94.65807976753814\\
0.43  94.75056667795737\\
0.44  94.84607189608973\\
0.45  94.94157711422207\\
0.46  95.03585265143427\\
0.47  95.1289357707845\\
0.48  95.22101278756368\\
0.49  95.30526456212367\\
0.5  95.38061046577705\\
0.51  95.45364605982286\\
0.52  95.52008609256963\\
0.53  95.58507286604713\\
0.54  95.6537859453433\\
0.55  95.72074766090468\\
0.56  95.7881937962225\\
0.57  95.85277067605993\\
0.58  95.91686313614095\\
0.59  95.98613516130992\\
0.6  96.0564505521081\\
0.61  96.12821920217557\\
0.62  96.20036048282489\\
0.63  96.27883648336596\\
0.64  96.35284091692462\\
0.65  96.42371525359562\\
0.66  96.49019254940056\\
0.67  96.557564158602\\
0.68  96.62214103843945\\
0.69  96.68936359540815\\
0.7  96.74909627768132\\
0.71  96.81028221922378\\
0.72  96.86774185494758\\
0.73  96.92892779649004\\
0.74  96.98247481110425\\
0.75  97.03434498810005\\
0.76  97.08546990393211\\
0.77  97.1317878852581\\
0.78  97.17818039270047\\
0.79  97.22859731042699\\
0.8  97.27800812558247\\
0.81  97.32775430826163\\
0.82  97.37343881759844\\
0.83  97.41852711800428\\
0.84  97.46696909364691\\
0.85  97.50807024682678\\
0.86  97.55576696130566\\
0.87  97.59992368525683\\
0.88  97.64702419080474\\
0.89  97.68346746171127\\
0.9  97.72378609066922\\
0.91  97.76913523248236\\
0.92  97.81657110555395\\
0.93  97.86542297483662\\
0.94  97.91755399323974\\
0.95  97.97523720731265\\
0.96  98.03180252963996\\
0.97  98.07886577212969\\
0.98  98.12868648092523\\
0.99  98.18461833126338\\
1.0  98.23991670961234\\
1.01  98.29059446874618\\
1.02  98.33780676346863\\
1.03  98.3881118920206\\
1.04  98.43916228173629\\
1.05  98.49039898674292\\
1.06  98.54592094344102\\
1.07  98.59719491150582\\
1.08  98.64928866685074\\
1.09  98.69866221894804\\
1.1  98.73860821732411\\
1.11  98.78444177889368\\
1.12  98.82349346387328\\
1.13  98.86437103870402\\
1.14  98.90804334289875\\
1.15  98.95067228146426\\
1.16  98.99199701299327\\
1.17  99.03384342733685\\
1.18  99.07963972584822\\
1.19  99.12454171096313\\
1.2  99.17212663626746\\
1.21  99.21516546847303\\
1.22  99.26278765683556\\
1.23  99.31070794966357\\
1.24  99.3633233878231\\
1.25  99.4012571810571\\
1.26  99.43594908822885\\
1.27  99.46110165250485\\
1.28  99.48383211799869\\
1.29  99.51017710013664\\
1.3  99.5312307280121\\
1.31  99.55530266360067\\
1.32  99.58250469607691\\
1.33  99.60706105142191\\
1.34  99.63314519215255\\
1.35  99.65393797862069\\
1.36  99.67394824086692\\
1.37  99.68687852205768\\
1.38  99.6960079713134\\
1.39  99.7054727880928\\
1.4  99.71739696671253\\
1.41  99.72790514912117\\
1.42  99.74448721001423\\
1.43  99.7560014949939\\
1.44  99.76587620541335\\
1.45  99.7788437496623\\
1.46  99.79523949526443\\
1.47  99.8117842930993\\
1.48  99.82717393613038\\
1.49  99.84345789255794\\
};
\addlegendentry{\textcolor{black}{DPFM \huge{(\textit{Unsup})}  \Huge{(\textbf{88.75})}}}

\addplot [color=cPLOT2, solid, smooth, line width=\pckLineWidth]  
table[row sep=crcr]{%
0.0  3.458868850058298\\
0.01  10.665879670877764\\
0.02  23.360024861912425\\
0.03  34.56252983373596\\
0.04  43.84091953303426\\
0.05  51.594467628277144\\
0.06  58.044852052616925\\
0.07  63.35826604556601\\
0.08  67.67388712945149\\
0.09  71.16454136814299\\
0.1  74.05000627882531\\
0.11  76.45470321278361\\
0.12  78.40479083686493\\
0.13  79.92322319491225\\
0.14  81.12234840735826\\
0.15  82.06614714510943\\
0.16  82.86130354375409\\
0.17  83.53106975160073\\
0.18  84.13659444713359\\
0.19  84.67582816215243\\
0.2  85.1728428322458\\
0.21  85.6277129835301\\
0.22  86.03108558840046\\
0.23  86.40401427473233\\
0.24  86.76099437216033\\
0.25  87.08779139245713\\
0.26  87.39163436891099\\
0.27  87.70050785822002\\
0.28  88.00364283656833\\
0.29  88.3266390249301\\
0.3  88.64434385902939\\
0.31  88.96212321924503\\
0.32  89.27163018054324\\
0.33  89.579646619514\\
0.34  89.88870642411398\\
0.35  90.19705823060839\\
0.36  90.5121919136928\\
0.37  90.81350100218995\\
0.38  91.1120153613231\\
0.39  91.41462865685679\\
0.4  91.70472156483977\\
0.41  91.99153532370232\\
0.42  92.28576443114402\\
0.43  92.5663552592894\\
0.44  92.84105852424129\\
0.45  93.10994875211608\\
0.46  93.3586424024537\\
0.47  93.59216998810935\\
0.48  93.8056873115188\\
0.49  94.01551559216776\\
0.5  94.20648876537427\\
0.51  94.38781080651046\\
0.52  94.55176826253165\\
0.53  94.70242280678023\\
0.54  94.85352450772704\\
0.55  94.99810517349121\\
0.56  95.1395184793095\\
0.57  95.27687011178546\\
0.58  95.41425900731959\\
0.59  95.547511703395\\
0.6  95.68281386767067\\
0.61  95.8129737299166\\
0.62  95.94484769283912\\
0.63  96.07713154940168\\
0.64  96.2112412958154\\
0.65  96.35258007551731\\
0.66  96.47860373830453\\
0.67  96.60421750745168\\
0.68  96.74406576482613\\
0.69  96.8729959461519\\
0.7  96.99111984060355\\
0.71  97.11263467335017\\
0.72  97.22133101408059\\
0.73  97.31795412395854\\
0.74  97.40578315210445\\
0.75  97.47714190853186\\
0.76  97.55490991096738\\
0.77  97.62913792287516\\
0.78  97.69475816834182\\
0.79  97.76648955535111\\
0.8  97.838742625175\\
0.81  97.91718136265787\\
0.82  97.99275084466036\\
0.83  98.07785966955865\\
0.84  98.17135268254893\\
0.85  98.26797579242688\\
0.86  98.3613197531844\\
0.87  98.44311216590408\\
0.88  98.51290587388765\\
0.89  98.59879722300785\\
0.9  98.6895327696923\\
0.91  98.77125065629562\\
0.92  98.85736558376493\\
0.93  98.9322270676619\\
0.94  99.00384666549661\\
0.95  99.07464647605121\\
0.96  99.13177074425134\\
0.97  99.17838683004283\\
0.98  99.22436944384513\\
0.99  99.2671101715852\\
1.0  99.30411238836454\\
};
\addlegendentry{\textcolor{black}{ULRSSM \huge{(\textit{Unsup})}~\Huge{(\textbf{87.45})}  }}

\addplot [color=cPLOT1, solid, smooth, line width=\pckLineWidth]
table[row sep=crcr]{%
0.0  4.880305467645791\\
0.01  11.79733516965684\\
0.02  20.741288921730064\\
0.03  28.039258867583115\\
0.04  34.298073909785394\\
0.05  39.44786581423695\\
0.06  43.69917085969229\\
0.07  47.245048950616386\\
0.08  50.24297377090598\\
0.09  52.820794873199404\\
0.1  55.03287905939098\\
0.11  56.89755975410853\\
0.12  58.519024468041906\\
0.13  59.921643241245135\\
0.14  61.148790273447496\\
0.15  62.194987895095544\\
0.16  63.12827845043808\\
0.17  63.94847562418418\\
0.18  64.6913146891348\\
0.19  65.37665685530345\\
0.2  66.01296083689847\\
0.21  66.61371586098346\\
0.22  67.17698424853268\\
0.23  67.70347399765168\\
0.24  68.19303605610773\\
0.25  68.64973209724316\\
0.26  69.07710211158572\\
0.27  69.49064753134103\\
0.28  69.90560894730743\\
0.29  70.3070066100939\\
0.3  70.69346178654752\\
0.31  71.05003199033546\\
0.32  71.38834329561195\\
0.33  71.69893088559756\\
0.34  72.00683553539375\\
0.35  72.3148519743645\\
0.36  72.62733998031766\\
0.37  72.9509696406686\\
0.38  73.26196712429429\\
0.39  73.57724985961143\\
0.4  73.88455830047663\\
0.41  74.19849956569907\\
0.42  74.49805729046145\\
0.43  74.80275731725358\\
0.44  75.10864976190769\\
0.45  75.4190137735442\\
0.46  75.73183714702103\\
0.47  76.04574114918526\\
0.48  76.36113567367696\\
0.49  76.69419288774914\\
0.5  77.03157261657095\\
0.51  77.3635864650139\\
0.52  77.69228390127823\\
0.53  78.02429774972119\\
0.54  78.34759204254846\\
0.55  78.66917223469913\\
0.56  79.00759532915018\\
0.57  79.33156035702481\\
0.58  79.66033231940553\\
0.59  79.98895522955348\\
0.6  80.32752737623727\\
0.61  80.67776286013348\\
0.62  81.01395017109333\\
0.63  81.35319305282448\\
0.64  81.69709381682897\\
0.65  82.0260520945006\\
0.66  82.36074888313298\\
0.67  82.70170586554073\\
0.68  83.04184306066836\\
0.69  83.37754595187178\\
0.7  83.72167029422538\\
0.71  84.0754830317075\\
0.72  84.4282896666186\\
0.73  84.8008457223686\\
0.74  85.16397422439739\\
0.75  85.52184863522186\\
0.76  85.86832055024122\\
0.77  86.23014484523348\\
0.78  86.60940825145708\\
0.79  86.98118178298516\\
0.8  87.34758943413438\\
0.81  87.70680531505357\\
0.82  88.06982202790779\\
0.83  88.42523707689195\\
0.84  88.79168199109935\\
0.85  89.13949537621343\\
0.86  89.49334537675374\\
0.87  89.82882468960804\\
0.88  90.15629244495221\\
0.89  90.49542353750881\\
0.9  90.84029314102614\\
0.91  91.19824207796698\\
0.92  91.56960571585502\\
0.93  91.95960088283638\\
0.94  92.33819355401262\\
0.95  92.72192852721861\\
0.96  93.09001301598623\\
0.97  93.44606153695955\\
0.98  93.77788907011157\\
0.99  94.08109857457625\\
1.0  94.36489402572566\\
    };
\addlegendentry{\textcolor{black}{Smooth Shells \huge{(\textit{Axio})}  \Huge{(\textbf{73.23})}}}
\end{axis}
\end{tikzpicture}

%% file: vis/tikz_vis/partial_full.tex
\newcommand{\pckLineWidth}{3pt}
\newcommand{\plotWidth}{1*\linewidth}
\newcommand{\plotHeight}{0.65*\linewidth}
\newcommand{\pckTitle}{\textsc{Partial-to-Full (P2F)}}

\pgfplotsset{%
    label style = {font=\Huge},
    tick label style = {font=\huge},
    title style =  {font=\Huge},
    legend style={  fill= gray!10,
                    fill opacity=0.6, 
                    font=\Huge,
                    draw=gray!20, %
                    text opacity=1}
}
\begin{tikzpicture}[scale=0.29, transform shape]
	\begin{axis}[
		width=\plotWidth,
		height=\plotHeight,
		grid=major,
		title=\pckTitle,
		legend style={
			at={(0.99,0.01)},
			anchor=south east,
			legend columns=1},
		legend cell align={left},
	ylabel={{\huge
        $\%$ Samples $<$ Geo Dist Thresh}},
        xmin=0,
        xmax=1.0,
        xlabel=\huge
        Geodesic Distance Threshold,
        ylabel near ticks,
        xtick={0, 0.2, 0.4, 0.6, 0.8, 1}, %
        xticklabels={$0$, $0.2$, $0.4$, $0.6$, $0.8$, $1$}, %
        ymin=0,
        ymax=100,
        ytick={0, 20, 40, 60, 80, 100},
        yticklabels={$0$, $20$, $40$, $60$, $80$, $100$},
	]
\addplot [color=cPLOT3, solid, smooth, line width=\pckLineWidth]
table[row sep=crcr]{%
0.0  2.524035362355019\\
0.01  6.419619477621617\\
0.02  16.59827358701741\\
0.03  27.866931893048946\\
0.04  38.82620509403566\\
0.05  48.658702152093106\\
0.06  57.23142372082021\\
0.07  64.58596245898191\\
0.08  70.65703321396153\\
0.09  75.6826820201999\\
0.1  79.75797781201688\\
0.11  82.93557472795673\\
0.12  85.34615518284514\\
0.13  87.03426405471248\\
0.14  88.29508590023816\\
0.15  89.18344231008541\\
0.16  89.85060649420973\\
0.17  90.33570470034414\\
0.18  90.69561386167844\\
0.19  90.9783542536061\\
0.2  91.20814154807651\\
0.21  91.40276380041055\\
0.22  91.57343196984401\\
0.23  91.71517586444904\\
0.24  91.84817521085012\\
0.25  91.99488929738303\\
0.26  92.12545960261635\\
0.27  92.25431089410016\\
0.28  92.37408130860295\\
0.29  92.49243166826922\\
0.3  92.58936909579502\\
0.31  92.70693468831477\\
0.32  92.8060021980954\\
0.33  92.93391924297623\\
0.34  93.06396637011188\\
0.35  93.17876659268678\\
0.36  93.31266281582673\\
0.37  93.42477240818504\\
0.38  93.53067860309953\\
0.39  93.63748167475289\\
0.4  93.748283321867\\
0.41  93.8455570781699\\
0.42  93.94813735517776\\
0.43  94.0345364810219\\
0.44  94.14133955267525\\
0.45  94.26708914543715\\
0.46  94.3852526557828\\
0.47  94.52004575566163\\
0.48  94.66653562300982\\
0.49  94.81067118891849\\
0.5  94.95387250822417\\
0.51  95.06897168971203\\
0.52  95.17756851484313\\
0.53  95.28702484684895\\
0.54  95.40694474080823\\
0.55  95.51823219615592\\
0.56  95.61083471944387\\
0.57  95.70523099620958\\
0.58  95.79125642341252\\
0.59  95.88471845357523\\
0.6  95.97048229172934\\
0.61  96.05527451341635\\
0.62  96.13109796771468\\
0.63  96.20535188772\\
0.64  96.2787836707147\\
0.65  96.34582520694497\\
0.66  96.41309096235995\\
0.67  96.47781556701482\\
0.68  96.54261491139792\\
0.69  96.6013977076578\\
0.7  96.66571124380735\\
0.71  96.7321174923476\\
0.72  96.79893480939315\\
0.73  96.87520670206091\\
0.74  96.94949799193036\\
0.75  97.01463366509053\\
0.76  97.08679487270516\\
0.77  97.16743903947491\\
0.78  97.2458783842616\\
0.79  97.31628320826262\\
0.8  97.38336211435701\\
0.81  97.43952902012853\\
0.82  97.49334162446051\\
0.83  97.54864902335729\\
0.84  97.61139302521383\\
0.85  97.6782477121235\\
0.86  97.7478303991139\\
0.87  97.83123993582848\\
0.88  97.91909648637326\\
0.89  97.99002448847196\\
0.9  98.0523947916873\\
0.91  98.11226131400663\\
0.92  98.1783312337698\\
0.93  98.24653123578777\\
0.94  98.30826625131309\\
0.95  98.3738503628427\\
0.96  98.43887392641052\\
0.97  98.49788094185512\\
0.98  98.55303886129539\\
0.99  98.61312960279946\\
1.0  98.67628467316129\\
};
\addlegendentry{\textcolor{black}{GeomFmaps \huge{(\textit{Sup})}~\Huge{(\textbf{85.35})} }}

\addplot [color=cPLOT4, solid, smooth, line width=\pckLineWidth]  
table[row sep=crcr]{%
0.0  1.585427476306092\\
0.01  4.011387865268244\\
0.02  11.181799134818029\\
0.03  20.67858859488039\\
0.04  30.997054437446852\\
0.05  40.97585624668778\\
0.06  49.86677183598516\\
0.07  57.49010956506735\\
0.08  63.61996080799616\\
0.09  68.44649305512762\\
0.1  72.17577243988711\\
0.11  75.12133499303664\\
0.12  77.50871960462909\\
0.13  79.4280185853905\\
0.14  80.97560975609757\\
0.15  82.22137319907813\\
0.16  83.18810929392771\\
0.17  83.92486966810043\\
0.18  84.51804927346899\\
0.19  84.99426909377735\\
0.2  85.37522030096501\\
0.21  85.6896190487928\\
0.22  85.94449031908206\\
0.23  86.1671945673474\\
0.24  86.34060069756838\\
0.25  86.50796780832891\\
0.26  86.6633801254637\\
0.27  86.80720738485809\\
0.28  86.97198634442131\\
0.29  87.12789164273654\\
0.3  87.29797014998952\\
0.31  87.50181786810288\\
0.32  87.70184498206781\\
0.33  87.90544620959095\\
0.34  88.10707551239231\\
0.35  88.31301840052255\\
0.36  88.53190204463944\\
0.37  88.75694795351188\\
0.38  89.02365077213177\\
0.39  89.31709781979073\\
0.4  89.64382109712962\\
0.41  89.97719962040449\\
0.42  90.33448773093087\\
0.43  90.703730635083\\
0.44  91.07445248277646\\
0.45  91.42804323444953\\
0.46  91.76006605947819\\
0.47  92.06472842899223\\
0.48  92.332417209973\\
0.49  92.57102010130764\\
0.5  92.76525468640234\\
0.51  92.93373100481888\\
0.52  93.06079690407819\\
0.53  93.18589087861571\\
0.54  93.29299104006704\\
0.55  93.38542501140019\\
0.56  93.47884494509422\\
0.57  93.58619159713578\\
0.58  93.70130270276931\\
0.59  93.82183660138774\\
0.6  93.94274023589149\\
0.61  94.07128507869213\\
0.62  94.21856320634959\\
0.63  94.37779612763283\\
0.64  94.5408496530645\\
0.65  94.72904521869877\\
0.66  94.91625482197217\\
0.67  95.09520699047313\\
0.68  95.27662406487633\\
0.69  95.4273530607969\\
0.7  95.58030047202948\\
0.71  95.69097474703904\\
0.72  95.79450079493216\\
0.73  95.88742774744574\\
0.74  95.9743156804989\\
0.75  96.05479485820628\\
0.76  96.15203539604875\\
0.77  96.22733827136149\\
0.78  96.31015910967598\\
0.79  96.39224047621981\\
0.8  96.4771564845512\\
0.81  96.5589913605048\\
0.82  96.6350337075882\\
0.83  96.71612911177115\\
0.84  96.79993591244654\\
0.85  96.88361946782682\\
0.86  96.94955570071112\\
0.87  97.00760423470834\\
0.88  97.0684874104931\\
0.89  97.12209911386634\\
0.9  97.18199632729021\\
0.91  97.24990448489629\\
0.92  97.33827136149078\\
0.93  97.40901416088441\\
0.94  97.48394730031181\\
0.95  97.54741862729391\\
0.96  97.59807244358447\\
0.97  97.65106792048213\\
0.98  97.7015984914776\\
0.99  97.77049261144455\\
1.0  97.82077669184979\\
};
\addlegendentry{\textcolor{black}{DPFM \huge{(\textit{Sup})}~\Huge{(\textbf{85.44})}}}

\addplot [color=cPLOT5, solid, smooth, line width=\pckLineWidth]
table[row sep=crcr]{%
0.0  0.19441117964494906\\
0.01  0.7530927912535011\\
0.02  2.126247012431292\\
0.03  4.15027783762658\\
0.04  6.574404866292208\\
0.05  9.303176217700436\\
0.06  12.147191309619847\\
0.07  15.115970277136142\\
0.08  17.983535176899142\\
0.09  20.711805468566016\\
0.1  23.323078310626975\\
0.11  25.841154040796283\\
0.12  28.207408669335642\\
0.13  30.38451324551426\\
0.14  32.377979426487016\\
0.15  34.229896230527565\\
0.16  35.98360532526293\\
0.17  37.61756114181494\\
0.18  39.132765799666295\\
0.19  40.57982633269365\\
0.2  41.95398267335415\\
0.21  43.27076767362973\\
0.22  44.50838523477154\\
0.23  45.73272471276751\\
0.24  46.94553981671235\\
0.25  48.045115419111426\\
0.26  49.091829218797756\\
0.27  50.068144124824\\
0.28  51.06124453218557\\
0.29  52.049835401874965\\
0.3  53.00986586630724\\
0.31  54.03829098543419\\
0.32  54.98379071736723\\
0.33  55.91125229861157\\
0.34  56.86276474744084\\
0.35  57.75189525847166\\
0.36  58.60294522915967\\
0.37  59.43395281019356\\
0.38  60.23765263532371\\
0.39  61.023063779894485\\
0.4  61.79394419196601\\
0.41  62.54954228192627\\
0.42  63.284847452361745\\
0.43  64.02491269034007\\
0.44  64.75395461400862\\
0.45  65.46520891685915\\
0.46  66.16819573397737\\
0.47  66.83836313803695\\
0.48  67.50276835507097\\
0.49  68.18045165525085\\
0.5  68.91400311659159\\
0.51  69.66158425069021\\
0.52  70.41643075103845\\
0.53  71.128436643501\\
0.54  71.79659980859518\\
0.55  72.37732804882326\\
0.56  72.9968884190062\\
0.57  73.57937036832901\\
0.58  74.16986927351348\\
0.59  74.76688195533553\\
0.6  75.34861231504632\\
0.61  75.927336316309\\
0.62  76.54313873843178\\
0.63  77.16094539951999\\
0.64  77.76672662681572\\
0.65  78.35922977096558\\
0.66  78.91991562153956\\
0.67  79.49914068254358\\
0.68  80.0793678630303\\
0.69  80.6270261603291\\
0.7  81.26061620326992\\
0.71  81.85787941496264\\
0.72  82.38850168105543\\
0.73  82.89732784839936\\
0.74  83.39913917936435\\
0.75  83.90495898826018\\
0.76  84.40677031922516\\
0.77  84.89805939562174\\
0.78  85.44797246175662\\
0.79  85.95128697194566\\
0.8  86.45410042239337\\
0.81  86.91457432469673\\
0.82  87.32970232040766\\
0.83  87.74758614469603\\
0.84  88.15920672221749\\
0.85  88.54076371525778\\
0.86  88.9005246095492\\
0.87  89.25001377914289\\
0.88  89.62255169683881\\
0.89  89.98907689763851\\
0.9  90.37113495042014\\
0.91  90.75845413048597\\
0.92  91.14251642223302\\
0.93  91.51480381005828\\
0.94  91.88558801865948\\
0.95  92.25186268958848\\
0.96  92.5647744980634\\
0.97  92.85789444675488\\
0.98  93.15502287337719\\
0.99  93.46091984547317\\
1.0  93.82093126963528\\
};

\addlegendentry{\textcolor{black}{DPFM \huge{(\textit{Unsup})}~\Huge{ (\textbf{62.72})} }}

\addplot [color=cPLOT2, solid, smooth, line width=\pckLineWidth]  
table[row sep=crcr]{%
0.0  0.2327422498584506\\
0.01  1.0850448698998383\\
0.02  3.014375403979417\\
0.03  5.607610095351669\\
0.04  8.494716325027433\\
0.05  11.441949723665552\\
0.06  14.293230181834579\\
0.07  17.003712852683424\\
0.08  19.433852598245288\\
0.09  21.654549371921615\\
0.1  23.739709485561963\\
0.11  25.75271699644749\\
0.12  27.672276865570684\\
0.13  29.442520931770694\\
0.14  31.091007480821936\\
0.15  32.649553806300325\\
0.16  34.15273303035921\\
0.17  35.54843493989788\\
0.18  36.91307114547268\\
0.19  38.208811636611436\\
0.2  39.44191966008107\\
0.21  40.63794926269059\\
0.22  41.80341422107758\\
0.23  42.944076722267596\\
0.24  44.03964384673585\\
0.25  45.11967811922216\\
0.26  46.129313498048376\\
0.27  47.17352199902795\\
0.28  48.16236339858801\\
0.29  49.181017852758586\\
0.3  50.20693767317878\\
0.31  51.31077228337935\\
0.32  52.37627582336643\\
0.33  53.35760132680619\\
0.34  54.329657225030935\\
0.35  55.24308913351739\\
0.36  56.0918843353693\\
0.37  56.92765198394605\\
0.38  57.70279140381908\\
0.39  58.4598926730034\\
0.4  59.196951552533605\\
0.41  59.91847758008188\\
0.42  60.625973934872256\\
0.43  61.367542352074636\\
0.44  62.09583268613117\\
0.45  62.81385129548996\\
0.46  63.50882115674652\\
0.47  64.19878042058956\\
0.48  64.89249763249272\\
0.49  65.59598550935229\\
0.5  66.32302319405542\\
0.51  67.05607359565481\\
0.52  67.74352756079108\\
0.53  68.40943595704915\\
0.54  69.05956097145463\\
0.55  69.69640790271424\\
0.56  70.35129298466256\\
0.57  71.02221197833418\\
0.58  71.72469773571103\\
0.59  72.44071210610441\\
0.6  73.19555860645265\\
0.61  73.92735635869865\\
0.62  74.6318463550409\\
0.63  75.34109641892603\\
0.64  76.05911502828482\\
0.65  76.78114211557444\\
0.66  77.50567450157082\\
0.67  78.25976941230704\\
0.68  79.03641201140412\\
0.69  79.75418009089223\\
0.7  80.43962981706308\\
0.71  81.08624741327908\\
0.72  81.67123466130867\\
0.73  82.21814136899542\\
0.74  82.74550674676942\\
0.75  83.26009510113892\\
0.76  83.79973644257605\\
0.77  84.31357320733352\\
0.78  84.81338029933309\\
0.79  85.31469057055673\\
0.8  85.81073971449617\\
0.81  86.32983760653784\\
0.82  86.7827956127209\\
0.83  87.22548189420624\\
0.84  87.69071586405246\\
0.85  88.14592863907164\\
0.86  88.58185061404872\\
0.87  88.99572596040626\\
0.88  89.42413203926304\\
0.89  89.87132785842056\\
0.9  90.31651943861266\\
0.91  90.78050075910551\\
0.92  91.23596406399534\\
0.93  91.68741889095436\\
0.94  92.09052145287284\\
0.95  92.41871558345902\\
0.96  92.70281645680615\\
0.97  92.95134208851722\\
0.98  93.1890949357892\\
0.99  93.42835096228524\\
1.0  93.72873627722633\\
};
\addlegendentry{\textcolor{black}{ULRSSM \huge{(\textit{Unsup})}~\Huge{ (\textbf{60.93})}  }}
\addplot [color=cPLOT6, solid, smooth, line width=\pckLineWidth]
table[row sep=crcr]{%
0.0  0.5593576693103909\\
0.01  1.951629276153461\\
0.02  4.6126190992235365\\
0.03  7.444918331086352\\
0.04  10.224196137640192\\
0.05  12.933309332872916\\
0.06  15.522542677349321\\
0.07  17.922098913734263\\
0.08  20.11324298874305\\
0.09  22.052014875387403\\
0.1  23.90458102665643\\
0.11  25.64106331835754\\
0.12  27.334075388385003\\
0.13  28.91994254583659\\
0.14  30.39278712693672\\
0.15  31.856447858387487\\
0.16  33.235127367749506\\
0.17  34.566418212920816\\
0.18  35.83599358844843\\
0.19  37.111691530408955\\
0.2  38.38996095027129\\
0.21  39.676679511811045\\
0.22  40.89947847979125\\
0.23  42.0703580844204\\
0.24  43.200216493949064\\
0.25  44.29872736334126\\
0.26  45.38462574588166\\
0.27  46.46501381863244\\
0.28  47.55078974984418\\
0.29  48.61648366315598\\
0.3  49.64409521325511\\
0.31  50.67403333859875\\
0.32  51.67127695919065\\
0.33  52.641581287477756\\
0.34  53.60490589003136\\
0.35  54.52133163370889\\
0.36  55.40996092578102\\
0.37  56.286834890301975\\
0.38  57.158198545033315\\
0.39  58.0317663236805\\
0.4  58.92615082819957\\
0.41  59.81600463355827\\
0.42  60.668510783676254\\
0.43  61.54587455351185\\
0.44  62.41393202236941\\
0.45  63.27329544689224\\
0.46  64.11061763225662\\
0.47  64.9631237823746\\
0.48  65.76517998508542\\
0.49  66.56441980723712\\
0.5  67.33378150519623\\
0.51  68.08306118525539\\
0.52  68.78850328965494\\
0.53  69.48476154440513\\
0.54  70.1872648169169\\
0.55  70.86723704495556\\
0.56  71.53324982152719\\
0.57  72.18493579264582\\
0.58  72.84592806474247\\
0.59  73.52161449627809\\
0.6  74.23293426445323\\
0.61  74.91731474032359\\
0.62  75.62116497745059\\
0.63  76.32256618800443\\
0.64  76.97217048653586\\
0.65  77.60475405038383\\
0.66  78.21958217157635\\
0.67  78.81898142535796\\
0.68  79.40846212151826\\
0.69  79.98888141935784\\
0.7  80.56391285873647\\
0.71  81.11335197042555\\
0.72  81.66266863078597\\
0.73  82.1879848307294\\
0.74  82.68917811892717\\
0.75  83.18253452209083\\
0.76  83.6664621729478\\
0.77  84.13459360240788\\
0.78  84.60064335928077\\
0.79  85.03889666454826\\
0.8  85.4816806689761\\
0.81  85.92568918669052\\
0.82  86.34679930594586\\
0.83  86.76668491191464\\
0.84  87.18020304879317\\
0.85  87.5926191237138\\
0.86  87.98250415416132\\
0.87  88.36296043230217\\
0.88  88.74133503785583\\
0.89  89.11101559907476\\
0.9  89.4973495409912\\
0.91  89.88882643871128\\
0.92  90.29169130999655\\
0.93  90.68586213694712\\
0.94  91.0819921851562\\
0.95  91.45093803840318\\
0.96  91.79808755514902\\
0.97  92.11658346098885\\
0.98  92.41009929578242\\
0.99  92.68132898876019\\
1.0  92.92978273460756\\
};
\addlegendentry{\textcolor{black}{PFM \huge{(\textit{Axio})}~\Huge{(\textbf{60.70})}  }}
\end{axis}
\end{tikzpicture}

%% file: vis/tikz_vis/partial_partial.tex
\newcommand{\pckLineWidth}{3pt}
\newcommand{\plotWidth}{1*\linewidth}
\newcommand{\plotHeight}{0.65*\linewidth}
\newcommand{\pckTitle}{
\textsc{Partial-to-Partial (P2P)}
}

\pgfplotsset{%
    label style = {font=\Huge},
    tick label style = {font=\huge},
    title style =  {font=\Huge},
    legend style={  fill= gray!10,
                    fill opacity=0.6, 
                    font=\Huge,
                    draw=gray!20, %
                    text opacity=1}
}
\begin{tikzpicture}[scale=0.29, transform shape]
	\begin{axis}[
		width=\plotWidth,
		height=\plotHeight,
		grid=major,
		title=\pckTitle,
		legend style={
			at={(0.99,0.98)},
			anchor=north east,
			legend columns=1},
		legend cell align={left},
	ylabel={{\huge
        $\%$ Samples $<$ Geo Dist Thresh}},
        xmin=0,
        xmax=1,
        xlabel=\huge
        Geodesic Distance Threshold,
        ylabel near ticks,
        xtick={0, 0.25, 0.5, 0.75, 1, 1.25, 1.5},
        xticklabels={$0$, $0.25$, $0.5$, $0.75$, $1$, $1.25$,$1.5$},
        ymin=0,
        ymax=100,
        ytick={0, 20, 40, 60, 80, 100},
        yticklabels={$0$, $20$, $40$, $60$, $80$, $100$},
	]

\addplot [color=cPLOT4, solid, smooth, line width=\pckLineWidth]
table[row sep=crcr]{%
0.0  0.6094774203643541\\
0.01  1.1816177643922035\\
0.02  3.388378408944786\\
0.03  6.355668961115217\\
0.04  9.787736397142552\\
0.05  13.32112519385069\\
0.06  16.806177194637716\\
0.07  20.022588156555447\\
0.08  22.84254443654324\\
0.09  25.26387707580976\\
0.1  27.308430587896282\\
0.11  29.044062397845916\\
0.12  30.49958867245779\\
0.13  31.72350113328097\\
0.14  32.80038917317742\\
0.15  33.69369034395039\\
0.16  34.446164118914716\\
0.17  35.08290844979867\\
0.18  35.64358429735805\\
0.19  36.122304487730666\\
0.2  36.565391783674244\\
0.21  36.94046672894677\\
0.22  37.3093446491012\\
0.23  37.62880129393885\\
0.24  37.912160267463605\\
0.25  38.18684340582303\\
0.26  38.44727338641085\\
0.27  38.68245048964244\\
0.28  38.89655770747253\\
0.29  39.09889057757823\\
0.3  39.27782967786314\\
0.31  39.44437472791187\\
0.32  39.60704663726178\\
0.33  39.7732818360546\\
0.34  39.919531628841575\\
0.35  40.06082380153407\\
0.36  40.202735676738385\\
0.37  40.326986030356125\\
0.38  40.451391309601824\\
0.39  40.57564166321957\\
0.4  40.696793504278276\\
0.41  40.806016071984644\\
0.42  40.904858622618214\\
0.43  41.00788416520649\\
0.44  41.100219839466064\\
0.45  41.188527447398876\\
0.46  41.275905501563976\\
0.47  41.360030117542074\\
0.48  41.433619790819414\\
0.49  41.499463182699145\\
0.5  41.56267283890369\\
0.51  41.62138965189761\\
0.52  41.68583871312576\\
0.53  41.75586509696021\\
0.54  41.81830012502498\\
0.55  41.87732678927481\\
0.56  41.94162092487501\\
0.57  42.00080251475279\\
0.58  42.063082617189615\\
0.59  42.11296866939026\\
0.6  42.163784275358616\\
0.61  42.21506465821083\\
0.62  42.26665489231895\\
0.63  42.31700572140344\\
0.64  42.36363833541708\\
0.65  42.40934139566301\\
0.66  42.45457967902509\\
0.67  42.49331108601316\\
0.68  42.53002845983785\\
0.69  42.560703734172414\\
0.7  42.58905512408768\\
0.71  42.620195175306094\\
0.72  42.660011061689836\\
0.73  42.69905231993382\\
0.74  42.723840420406184\\
0.75  42.7520368846935\\
0.76  42.78503604344734\\
0.77  42.817880276573234\\
0.78  42.84762599714007\\
0.79  42.872104246356535\\
0.8  42.8968923468289\\
0.81  42.917497455346556\\
0.82  42.939187043259885\\
0.83  42.960721705545254\\
0.84  42.98504502913376\\
0.85  43.006114914535274\\
0.86  43.02702987430884\\
0.87  43.054296784828445\\
0.88  43.07970458781262\\
0.89  43.10247865512161\\
0.9  43.12246406112746\\
0.91  43.13826647517859\\
0.92  43.15825188118444\\
0.93  43.1794766922139\\
0.94  43.20503942082603\\
0.95  43.22998244692635\\
0.96  43.26050279563295\\
0.97  43.28947388806004\\
0.98  43.31503661667217\\
0.99  43.336261427701636\\
1.0  43.35531727993976\\
1.01  43.37654209096923\\
1.02  43.3980767532546\\
1.03  43.42085082056359\\
1.04  43.43696308587063\\
1.05  43.451216243642236\\
1.06  43.4620610375989\\
1.07  43.47739867476618\\
1.08  43.493201088817315\\
1.09  43.50884857724049\\
1.1  43.51798918928968\\
1.11  43.52589039631525\\
1.12  43.53348175208491\\
1.13  43.53797459529553\\
1.14  43.54448147166953\\
1.15  43.5533122324628\\
1.16  43.56012896009271\\
1.17  43.563227472651754\\
1.18  43.568185092746226\\
1.19  43.57035405153756\\
1.2  43.576396151027694\\
1.21  43.58243825051784\\
1.22  43.58584661433279\\
1.23  43.592043639450885\\
1.24  43.59886036708078\\
1.25  43.60149410275597\\
1.26  43.60443768968707\\
1.27  43.606916499734304\\
1.28  43.61032486354925\\
1.29  43.615282483643725\\
1.3  43.61822607057482\\
1.31  43.62287383941339\\
1.32  43.62581742634448\\
1.33  43.629070864531485\\
1.34  43.63185952583462\\
1.35  43.633408782114145\\
1.36  43.63464818713776\\
1.37  43.63619744341729\\
1.38  43.63774669969681\\
1.39  43.63960580723224\\
1.4  43.64409865044286\\
1.41  43.64797179114166\\
1.42  43.65230970872433\\
1.43  43.65649270067904\\
1.44  43.65804195695856\\
1.45  43.658971510726275\\
1.46  43.659436287610134\\
1.47  43.65959121323809\\
1.48  43.65959121323809\\
1.49  43.66005599012194\\
    };
\addlegendentry{\textcolor{black}{DPFM \huge{(\textit{Sup})}~\Huge{(\textbf{38.11})}  }}
\addplot [color=cPLOT7, solid, smooth, line width=\pckLineWidth]
table[row sep=crcr]{%
0.0  0.054296667525597\\
0.01  0.1284451432748384\\
0.02  0.38336339589501417\\
0.03  0.7731687692930653\\
0.04  1.2893157831078204\\
0.05  1.9258883355507759\\
0.06  2.634768798742105\\
0.07  3.409120788392871\\
0.08  4.294169615952965\\
0.09  5.233778048895924\\
0.1  6.176410267197451\\
0.11  7.1387628247939885\\
0.12  8.115051088825668\\
0.13  9.092785511072313\\
0.14  10.070388464390327\\
0.15  11.051015203066907\\
0.16  12.018232111022881\\
0.17  12.986500770407922\\
0.18  13.921902197634612\\
0.19  14.832718935206854\\
0.2  15.71329781919341\\
0.21  16.59729489532444\\
0.22  17.450791180012516\\
0.23  18.267739102540506\\
0.24  19.070882787561988\\
0.25  19.839976020067418\\
0.26  20.608148970072413\\
0.27  21.35278898185203\\
0.28  22.057725377184354\\
0.29  22.73741973821907\\
0.3  23.444196698552265\\
0.31  24.07340699099175\\
0.32  24.716815927723644\\
0.33  25.331564638013454\\
0.34  25.921991596506082\\
0.35  26.535820024295457\\
0.36  27.146361728869\\
0.37  27.75585168201347\\
0.38  28.322877171209353\\
0.39  28.904627180412177\\
0.4  29.477042895682033\\
0.41  30.053271209882254\\
0.42  30.63186596479788\\
0.43  31.158530492903306\\
0.44  31.681382422078364\\
0.45  32.203839944467525\\
0.46  32.710126788634774\\
0.47  33.21785979101699\\
0.48  33.7007451658875\\
0.49  34.17258715075279\\
0.5  34.61708359846234\\
0.51  35.03186806830074\\
0.52  35.47255191707992\\
0.53  35.924410624792934\\
0.54  36.39822464358773\\
0.55  36.85560504630336\\
0.56  37.29037279329403\\
0.57  37.711073364920935\\
0.58  38.10179902081942\\
0.59  38.49646874457691\\
0.6  38.885485304403154\\
0.61  39.250180112432226\\
0.62  39.60961616331596\\
0.63  39.96011232705263\\
0.64  40.297461597925945\\
0.65  40.64296194237454\\
0.66  40.971897201815324\\
0.67  41.301752743756545\\
0.68  41.624640432480184\\
0.69  41.939114109771296\\
0.7  42.253324849205136\\
0.71  42.556097791847876\\
0.72  42.8411224291251\\
0.73  43.12312328104376\\
0.74  43.38277441509474\\
0.75  43.634931820213616\\
0.76  43.878675213899946\\
0.77  44.11729131936958\\
0.78  44.348413695907105\\
0.79  44.582954264589105\\
0.8  44.799352121119696\\
0.81  45.01745907372251\\
0.82  45.20243585630972\\
0.83  45.38833292139736\\
0.84  45.58619365899064\\
0.85  45.77432569586504\\
0.86  45.95851366488044\\
0.87  46.14112400675224\\
0.88  46.3034881336145\\
0.89  46.4864928822722\\
0.9  46.65661367592383\\
0.91  46.82292187064509\\
0.92  46.979106957861575\\
0.93  47.11820108435572\\
0.94  47.2657092222824\\
0.95  47.40454041091928\\
0.96  47.52351979133252\\
0.97  47.64591736389022\\
0.98  47.77922685752449\\
0.99  47.90543702901257\\
1.0  48.02178703085313\\
1.01  48.14392166555356\\
1.02  48.25212059381886\\
1.03  48.36110833565595\\
1.04  48.457475060344244\\
1.05  48.54279839502732\\
1.06  48.631145515068965\\
1.07  48.72356817189825\\
1.08  48.8170425801566\\
1.09  48.90078828769609\\
1.1  48.97901230023296\\
1.11  49.05145167990997\\
1.12  49.11981552279934\\
1.13  49.180291229970706\\
1.14  49.238137558569406\\
1.15  49.28835868930737\\
1.16  49.33424134540043\\
1.17  49.377363153992185\\
1.18  49.42061643151258\\
1.19  49.4604515168885\\
1.2  49.498446037263555\\
1.21  49.534205585851836\\
1.22  49.56746722479609\\
1.23  49.59823095409631\\
1.24  49.620054796249455\\
1.25  49.6422730451885\\
1.26  49.66646332805705\\
1.27  49.68210813056442\\
1.28  49.69748999521453\\
1.29  49.70787604057657\\
1.3  49.72128587129718\\
1.31  49.733906888445986\\
1.32  49.74836847059566\\
1.33  49.76046361202993\\
1.34  49.77676575918048\\
1.35  49.785574177398914\\
1.36  49.79753784990456\\
1.37  49.80910711562429\\
1.38  49.81949316098633\\
1.39  49.82646101420391\\
1.4  49.83211417813514\\
1.41  49.836978528494576\\
1.42  49.840265251710406\\
1.43  49.84355197492625\\
1.44  49.84815338742842\\
1.45  49.85341214457375\\
1.46  49.85525270957462\\
1.47  49.85893383957635\\
1.48  49.86314084529263\\
1.49  49.86655903743709\\
    };
\addlegendentry{\textcolor{black}{SM-COMB \huge{(\textit{Axio})}~\Huge{(\textbf{30.03})}  }}
\addplot [color=cPLOT8, solid, smooth, line width=\pckLineWidth] 
table[row sep=crcr]{%
0.0  0.0934718386832998\\
0.01  0.17937232077222331\\
0.02  0.5085198410841082\\
0.03  1.0045125285302483\\
0.04  1.7050695050535365\\
0.05  2.545352426000314\\
0.06  3.495351187051053\\
0.07  4.575577281525064\\
0.08  5.751615452175182\\
0.09  6.988775119696265\\
0.1  8.270261638551696\\
0.11  9.46529703120225\\
0.12  10.696674935505808\\
0.13  11.906027075171181\\
0.14  13.098171586212972\\
0.15  14.258516399558383\\
0.16  15.468006200252745\\
0.17  16.6496884730914\\
0.18  17.813612473190517\\
0.19  18.908568297766312\\
0.2  20.008892902472667\\
0.21  21.085952793279937\\
0.22  22.12804678272409\\
0.23  23.14302154945748\\
0.24  24.127022584668417\\
0.25  25.10331460225599\\
0.26  25.979664712797796\\
0.27  26.828069634454906\\
0.28  27.6292568231689\\
0.29  28.41640258692606\\
0.3  29.201758757306358\\
0.31  29.970044960092068\\
0.32  30.688222548325903\\
0.33  31.386989931472343\\
0.34  32.02821500450151\\
0.35  32.6876113333572\\
0.36  33.342602509285236\\
0.37  33.91898923766075\\
0.38  34.5253860703558\\
0.39  35.1092064942967\\
0.4  35.69784505425221\\
0.41  36.25372028930842\\
0.42  36.79954626924845\\
0.43  37.29664024492651\\
0.44  37.80034194999601\\
0.45  38.331300538805266\\
0.46  38.84023336297634\\
0.47  39.33402347395866\\
0.48  39.785276326983485\\
0.49  40.22000985652968\\
0.5  40.65254080961204\\
0.51  41.05987979438949\\
0.52  41.44712026893458\\
0.53  41.828991963349125\\
0.54  42.22724532021332\\
0.55  42.60994298080179\\
0.56  42.97130318189703\\
0.57  43.34408924839831\\
0.58  43.67502636208705\\
0.59  44.009405001500504\\
0.6  44.31556312997129\\
0.61  44.60933176583308\\
0.62  44.90681724947758\\
0.63  45.189297680962305\\
0.64  45.46819892569333\\
0.65  45.75976498509131\\
0.66  46.029718262938076\\
0.67  46.287970353320794\\
0.68  46.525435628326235\\
0.69  46.768682666549196\\
0.7  46.99912722907621\\
0.71  47.20493046741426\\
0.72  47.384853432302435\\
0.73  47.558581650886126\\
0.74  47.7117983761505\\
0.75  47.87492669550206\\
0.76  48.03750437073767\\
0.77  48.186178282045425\\
0.78  48.35247280506373\\
0.79  48.49811817373372\\
0.8  48.621462455707565\\
0.81  48.736547075942084\\
0.82  48.87310681669884\\
0.83  49.00760164202075\\
0.84  49.1478782305602\\
0.85  49.263651155939655\\
0.86  49.375982555594405\\
0.87  49.48335815820556\\
0.88  49.57958321746863\\
0.89  49.6770472259926\\
0.9  49.769142454386014\\
0.91  49.85284036001112\\
0.92  49.941081079592855\\
0.93  50.04198661384154\\
0.94  50.13256757091607\\
0.95  50.20910710303378\\
0.96  50.28027785502091\\
0.97  50.34291362321075\\
0.98  50.421931053850244\\
0.99  50.4990212300839\\
1.0  50.56716343943328\\
1.01  50.636544598043564\\
1.02  50.70799067208868\\
1.03  50.78191464465559\\
1.04  50.84812959959913\\
1.05  50.90814980823819\\
1.06  50.955367541181296\\
1.07  51.00162164692149\\
1.08  51.03700053137157\\
1.09  51.0766469077203\\
1.1  51.11354006348927\\
1.11  51.14410081192475\\
1.12  51.17328495007034\\
1.13  51.19737563014336\\
1.14  51.22146631021638\\
1.15  51.24294143073861\\
1.16  51.262489296854994\\
1.17  51.276943704898805\\
1.18  51.28919553647879\\
1.19  51.29952011365294\\
1.2  51.313010894493836\\
1.21  51.32429909887091\\
1.22  51.33600028633494\\
1.23  51.343709303958306\\
1.24  51.3570624237702\\
1.25  51.364909102422565\\
1.26  51.3741323913648\\
1.27  51.38087778178524\\
1.28  51.3851452736839\\
1.29  51.39037639278546\\
1.3  51.3935425964522\\
1.31  51.39629581703198\\
1.32  51.39877371555377\\
1.33  51.401113953046575\\
1.34  51.40331652951039\\
1.35  51.40606975009017\\
1.36  51.41171385227871\\
1.37  51.41474239491646\\
1.38  51.41845924269914\\
1.39  51.42135012430791\\
1.4  51.424516327974644\\
1.41  51.42740720958341\\
1.42  51.43029809119217\\
1.43  51.43332663382992\\
1.44  51.43910839704745\\
1.45  51.44489016026497\\
1.46  51.4481940249607\\
1.47  51.45287449994631\\
1.48  51.456729008757996\\
1.49  51.46003287345372\\
};
\addlegendentry{\textcolor{black}{GC-PPSM \huge{(\textit{Axio})}~\Huge{(\textbf{34.29})}  }}
\end{axis}
\end{tikzpicture}

%% file: tab/tab_benchmark_overlap_region.tex
\begin{table}[tbh!]
    \centering
    \footnotesize
    \begin{tabular}{@{}lccc@{}}
    \toprule
                          & \textbf{DPFM}       & \textbf{SM-COMB}      &\textbf{GC-PPSM}       \\ \midrule
    mIoU ($\uparrow$)     & 46.52 &  46.91       & 49.24                     \\
    F1 Score($\uparrow$)  & 60.28 &  60.61      &  62.42                 \\
    \bottomrule
    \end{tabular}
    \caption{\textbf{Overlapping Region Prediction:} We show the mean IoU (x100) and F1 Score (x100) of the partial-to-partial shape matching methods on \bm{}.} 
    \label{tab:mIoU}
\end{table}

%% file: tab/tab_benchmark_scale_analysis.tex
\begin{table}[tbh!]
    \centering
    \footnotesize
    \begin{tabular}{@{}llccc}
    \toprule
                                                                            & Methods           & Default ($\uparrow$)  & Normalised ($\uparrow$)\\
    \midrule
    \parbox[t]{5mm}{\multirow{5}{*}{F2F}}                                   & GeomFmaps \footnotesize{(\textit{Sup})}           & 88.72             & 88.96 (-0.24)     \\     
                                                                            & DPFM \footnotesize{(\textit{Sup})}                & 90.20             & 91.47 (\underline{-1.27})      \\
                                                                            & DPFM \footnotesize{(\textit{Unsup})}              & 88.75             & 89.40 (-0.65)      \\
                                                                            & ULRSSM \footnotesize{(\textit{Unsup})}            & 87.45             & \textbf{94.06} (\underline{-6.61})    \\
                                                                            & Smooth Shells \footnotesize{(\textit{Axio})}      & 73.23             & 73.93 (-0.7)      \\
                                                                            
    \midrule
    \parbox[t]{5mm}{\multirow{5}{*}{P2F}}                                   & GeomFmaps \footnotesize{(\textit{Sup})}           & 85.35            & 89.92 (\underline{-4.57})   \\
                                                                            & DPFM \footnotesize{(\textit{Sup})}                & 85.44            & \textbf{90.02} (\underline{-4.58})      \\
                                                                            & DPFM \footnotesize{(\textit{Unsup})}              & 62.72             & 61.85 (+0.87)      \\
                                                                            & ULRSSM \footnotesize{(\textit{Unsup})}            & 60.93             & 61.00 (-0.07)      \\
                                                                            & PFM \footnotesize{(\textit{Axio})}                & 60.70             & 62.83 (-2.03)     \\
    \midrule
    \parbox[t]{5mm}{\multirow{3}{*}{P2P}}                                   & DPFM \footnotesize{(\textit{Sup})}                & 38.11            &  \textbf{39.58} (\underline{-1.47})   \\
                                                                            & SM-COMB \footnotesize{(\textit{Axio})}            & 30.03            &       25.09 (+4.94)     \\
                                                                            & GC-PPSM \footnotesize{(\textit{Axio})}            & 34.29            &     32.81 (+1.48)       \\
    \bottomrule
    \end{tabular}
    \caption{\textbf{Scale Analysis:} Correspondence prediction accuracy measured by Area-under-the-curve \textbf{(AUC x100)} for baselines on all three shape matching settings. 
    Unsupervised ULRSSM in the full-to-full setting, supervised GeomFmaps in partial-to-full and DPFM in all settings exhibit a notable performance drop (\underline{underlined}) in the \textit{default} compared to the \textit{normalised} scale setting. The best performing method in each setting is in \textbf{bold}.}
    \label{tab:p2p_normalized}
\end{table}

%% file: fig/fig_benchmark_rotation.tex
\begin{figure}[ht]%
    \centering
    \hspace{-15pt}
    \begin{tabular}{@{}l@{}@{}r@{}}%
      {\input{vis/tikz_vis/ablation_rotation_unsupervised}}&{\input{vis/tikz_vis/ablation_rotation_supervised}} 
    \end{tabular}
           
\caption{\textbf{Rotation evaluation} for the test set of \bm{} presented in Sec.~\ref{sec:challenges} with (solid) and without (dashed) rotation. Unsupervised methods struggle the most with rotation, whereas supervised methods can better accommodate rotational changes in the data due to their supervision signals.}
\label{fig:ablation_rot}
\end{figure}

%% file: vis/tikz_vis/ablation_rotation_unsupervised.tex
\newcommand{\pckLineWidth}{2pt}
\newcommand{\plotWidth}{1*\linewidth}
\newcommand{\plotHeight}{0.7*\linewidth}
\newcommand{\pckTitle}{\textsc{Unsupervised}}

\pgfplotsset{%
    label style = {font=\large},
    tick label style = {font=\large},
    title style =  {font=\LARGE},
    legend style={  fill= gray!10,
                    fill opacity=0.6, 
                    font=\large,
                    draw=gray!20, %
                    text opacity=1}
}
\begin{tikzpicture}[scale=0.45, transform shape]
\begin{axis}[
    width=\plotWidth,
    height=\plotHeight,
    grid=major,
    title=\pckTitle,
    legend style={
        at={(0.99,0.01)},
        anchor=south east,
        legend columns=1},
    legend cell align={left},
	ylabel={{\Large
        $\%$ Samples $<$ Geo Dist Thresh}},
        xmin=0,
        xmax=1,
        xlabel=\Large
        Geodesic Distance Threshold,
        ylabel near ticks,
        xtick={0, 0.2, 0.4, 0.6, 0.8, 1}, %
        xticklabels={$0$, $0.2$, $0.4$, $0.6$, $0.8$, $1$}, %
        ymin=0,
        ymax=100,
        ytick={0, 20, 40, 60, 80, 100},
        yticklabels={$0$, $20$, $40$, $60$, $80$, $100$},
	]

\addplot [color=cPLOT2, solid, smooth, line width=\pckLineWidth]
table[row sep=crcr]{%
.0  0.2327422498584506\\
0.01  1.0850448698998383\\
0.02  3.014375403979417\\
0.03  5.607610095351669\\
0.04  8.494716325027433\\
0.05  11.441949723665552\\
0.06  14.293230181834579\\
0.07  17.003712852683424\\
0.08  19.433852598245288\\
0.09  21.654549371921615\\
0.1  23.739709485561963\\
0.11  25.75271699644749\\
0.12  27.672276865570684\\
0.13  29.442520931770694\\
0.14  31.091007480821936\\
0.15  32.649553806300325\\
0.16  34.15273303035921\\
0.17  35.54843493989788\\
0.18  36.91307114547268\\
0.19  38.208811636611436\\
0.2  39.44191966008107\\
0.21  40.63794926269059\\
0.22  41.80341422107758\\
0.23  42.944076722267596\\
0.24  44.03964384673585\\
0.25  45.11967811922216\\
0.26  46.129313498048376\\
0.27  47.17352199902795\\
0.28  48.16236339858801\\
0.29  49.181017852758586\\
0.3  50.20693767317878\\
0.31  51.31077228337935\\
0.32  52.37627582336643\\
0.33  53.35760132680619\\
0.34  54.329657225030935\\
0.35  55.24308913351739\\
0.36  56.0918843353693\\
0.37  56.92765198394605\\
0.38  57.70279140381908\\
0.39  58.4598926730034\\
0.4  59.196951552533605\\
0.41  59.91847758008188\\
0.42  60.625973934872256\\
0.43  61.367542352074636\\
0.44  62.09583268613117\\
0.45  62.81385129548996\\
0.46  63.50882115674652\\
0.47  64.19878042058956\\
0.48  64.89249763249272\\
0.49  65.59598550935229\\
0.5  66.32302319405542\\
0.51  67.05607359565481\\
0.52  67.74352756079108\\
0.53  68.40943595704915\\
0.54  69.05956097145463\\
0.55  69.69640790271424\\
0.56  70.35129298466256\\
0.57  71.02221197833418\\
0.58  71.72469773571103\\
0.59  72.44071210610441\\
0.6  73.19555860645265\\
0.61  73.92735635869865\\
0.62  74.6318463550409\\
0.63  75.34109641892603\\
0.64  76.05911502828482\\
0.65  76.78114211557444\\
0.66  77.50567450157082\\
0.67  78.25976941230704\\
0.68  79.03641201140412\\
0.69  79.75418009089223\\
0.7  80.43962981706308\\
0.71  81.08624741327908\\
0.72  81.67123466130867\\
0.73  82.21814136899542\\
0.74  82.74550674676942\\
0.75  83.26009510113892\\
0.76  83.79973644257605\\
0.77  84.31357320733352\\
0.78  84.81338029933309\\
0.79  85.31469057055673\\
0.8  85.81073971449617\\
0.81  86.32983760653784\\
0.82  86.7827956127209\\
0.83  87.22548189420624\\
0.84  87.69071586405246\\
0.85  88.14592863907164\\
0.86  88.58185061404872\\
0.87  88.99572596040626\\
0.88  89.42413203926304\\
0.89  89.87132785842056\\
0.9  90.31651943861266\\
0.91  90.78050075910551\\
0.92  91.23596406399534\\
0.93  91.68741889095436\\
0.94  92.09052145287284\\
0.95  92.41871558345902\\
0.96  92.70281645680615\\
0.97  92.95134208851722\\
0.98  93.1890949357892\\
0.99  93.42835096228524\\
1.0  93.72873627722633\\
};
\addlegendentry{\textcolor{black}{ULRSSM~(\textbf{60.93})}}

\addplot [color=cPLOT2, dashed, smooth, line width=\pckLineWidth]
table[row sep=crcr]{%
0.0  0.664405217034027\\
0.01  2.5285979847377202\\
0.02  7.129829589581965\\
0.03  13.086678324656647\\
0.04  19.373725429282935\\
0.05  25.48640374391839\\
0.06  31.368344047660802\\
0.07  36.80183588289232\\
0.08  41.64432775319801\\
0.09  45.85272852082154\\
0.1  49.66905004083637\\
0.11  53.13087179384398\\
0.12  56.36020182686382\\
0.13  59.15160564594116\\
0.14  61.64412732930147\\
0.15  63.92720604077624\\
0.16  65.93520295424824\\
0.17  67.78010492190984\\
0.18  69.46667201130391\\
0.19  71.0167003211793\\
0.2  72.4141559398127\\
0.21  73.67081377112594\\
0.22  74.82074587753098\\
0.23  75.89977803053458\\
0.24  76.83976610531273\\
0.25  77.70133833056916\\
0.26  78.50654133492337\\
0.27  79.25888253656484\\
0.28  79.9728926679928\\
0.29  80.65633815519824\\
0.3  81.2959409150353\\
0.31  81.91750552418365\\
0.32  82.51677297484179\\
0.33  83.13508069567135\\
0.34  83.7055372112017\\
0.35  84.27023153970647\\
0.36  84.82340149416015\\
0.37  85.34700892387399\\
0.38  85.82877786518488\\
0.39  86.30077614153936\\
0.4  86.73744970612847\\
0.41  87.1568367096409\\
0.42  87.58474172875633\\
0.43  88.02868065959505\\
0.44  88.45708673845184\\
0.45  88.88724652640335\\
0.46  89.31014094810524\\
0.47  89.75458093868532\\
0.48  90.19100397340375\\
0.49  90.60688355872671\\
0.5  90.99821121672336\\
0.51  91.38051979937568\\
0.52  91.73827645470169\\
0.53  92.10730695420814\\
0.54  92.44376857052666\\
0.55  92.77096058163015\\
0.56  93.11819498238775\\
0.57  93.41782870771682\\
0.58  93.70092746158124\\
0.59  93.9677417738517\\
0.6  94.24132039263041\\
0.61  94.52191384778807\\
0.62  94.77595113665403\\
0.63  95.01545769302075\\
0.64  95.25571583899949\\
0.65  95.47668318493614\\
0.66  95.67961238018408\\
0.67  95.87076667151024\\
0.68  96.05841354464692\\
0.69  96.23704134243926\\
0.7  96.43120199221353\\
0.71  96.58878528086903\\
0.72  96.7298335980599\\
0.73  96.8736377438282\\
0.74  97.01343341166567\\
0.75  97.13068139114228\\
0.76  97.24993360958428\\
0.77  97.36091834229396\\
0.78  97.47490943345176\\
0.79  97.59416165189376\\
0.8  97.70564744434479\\
0.81  97.81437740821838\\
0.82  97.9431497617461\\
0.83  98.04837230743021\\
0.84  98.16587081677748\\
0.85  98.27259654168566\\
0.86  98.3637894146119\\
0.87  98.45047274986597\\
0.88  98.51586104611253\\
0.89  98.58049775274706\\
0.9  98.64964399705377\\
0.91  98.7217965998086\\
0.92  98.77666264148675\\
0.93  98.82576649613934\\
0.94  98.87712511962802\\
0.95  98.92071731712572\\
0.96  98.95629255876179\\
0.97  98.99612680819934\\
0.98  99.02944728099932\\
0.99  99.06351934341131\\
1.0  99.08757021099626\\
};
\addlegendentry{\textcolor{black}{ULRSSM~(\textbf{81.78}) }}

\addplot [color=cPLOT5, solid, smooth, line width=\pckLineWidth]
table[row sep=crcr]{%
0.0  0.19441117964494906\\
0.01  0.7530927912535011\\
0.02  2.126247012431292\\
0.03  4.15027783762658\\
0.04  6.574404866292208\\
0.05  9.303176217700436\\
0.06  12.147191309619847\\
0.07  15.115970277136142\\
0.08  17.983535176899142\\
0.09  20.711805468566016\\
0.1  23.323078310626975\\
0.11  25.841154040796283\\
0.12  28.207408669335642\\
0.13  30.38451324551426\\
0.14  32.377979426487016\\
0.15  34.229896230527565\\
0.16  35.98360532526293\\
0.17  37.61756114181494\\
0.18  39.132765799666295\\
0.19  40.57982633269365\\
0.2  41.95398267335415\\
0.21  43.27076767362973\\
0.22  44.50838523477154\\
0.23  45.73272471276751\\
0.24  46.94553981671235\\
0.25  48.045115419111426\\
0.26  49.091829218797756\\
0.27  50.068144124824\\
0.28  51.06124453218557\\
0.29  52.049835401874965\\
0.3  53.00986586630724\\
0.31  54.03829098543419\\
0.32  54.98379071736723\\
0.33  55.91125229861157\\
0.34  56.86276474744084\\
0.35  57.75189525847166\\
0.36  58.60294522915967\\
0.37  59.43395281019356\\
0.38  60.23765263532371\\
0.39  61.023063779894485\\
0.4  61.79394419196601\\
0.41  62.54954228192627\\
0.42  63.284847452361745\\
0.43  64.02491269034007\\
0.44  64.75395461400862\\
0.45  65.46520891685915\\
0.46  66.16819573397737\\
0.47  66.83836313803695\\
0.48  67.50276835507097\\
0.49  68.18045165525085\\
0.5  68.91400311659159\\
0.51  69.66158425069021\\
0.52  70.41643075103845\\
0.53  71.128436643501\\
0.54  71.79659980859518\\
0.55  72.37732804882326\\
0.56  72.9968884190062\\
0.57  73.57937036832901\\
0.58  74.16986927351348\\
0.59  74.76688195533553\\
0.6  75.34861231504632\\
0.61  75.927336316309\\
0.62  76.54313873843178\\
0.63  77.16094539951999\\
0.64  77.76672662681572\\
0.65  78.35922977096558\\
0.66  78.91991562153956\\
0.67  79.49914068254358\\
0.68  80.0793678630303\\
0.69  80.6270261603291\\
0.7  81.26061620326992\\
0.71  81.85787941496264\\
0.72  82.38850168105543\\
0.73  82.89732784839936\\
0.74  83.39913917936435\\
0.75  83.90495898826018\\
0.76  84.40677031922516\\
0.77  84.89805939562174\\
0.78  85.44797246175662\\
0.79  85.95128697194566\\
0.8  86.45410042239337\\
0.81  86.91457432469673\\
0.82  87.32970232040766\\
0.83  87.74758614469603\\
0.84  88.15920672221749\\
0.85  88.54076371525778\\
0.86  88.9005246095492\\
0.87  89.25001377914289\\
0.88  89.62255169683881\\
0.89  89.98907689763851\\
0.9  90.37113495042014\\
0.91  90.75845413048597\\
0.92  91.14251642223302\\
0.93  91.51480381005828\\
0.94  91.88558801865948\\
0.95  92.25186268958848\\
0.96  92.5647744980634\\
0.97  92.85789444675488\\
0.98  93.15502287337719\\
0.99  93.46091984547317\\
1.0  93.82093126963528\\
};
\addlegendentry{\textcolor{black}{DPFM~\normalsize{(\textit{Unsup})}~\large{(\textbf{62.72})} }}

\addplot [color=cPLOT5, dashed, smooth, line width=\pckLineWidth]
table[row sep=crcr]{%
0.0  0.5150894141108444\\
0.01  1.9761796198960802\\
0.02  5.5046423185036355\\
0.03  10.7139099194797\\
0.04  16.998952785140574\\
0.05  23.990990945850474\\
0.06  30.867534836178518\\
0.07  37.44193970247073\\
0.08  43.273774032077846\\
0.09  48.51761475520726\\
0.1  53.18323253681536\\
0.11  57.42294953827345\\
0.12  61.19693151014396\\
0.13  64.40196014570817\\
0.14  67.33466281184705\\
0.15  69.84021204848254\\
0.16  72.1052526092686\\
0.17  74.07992904994063\\
0.18  75.84290774989103\\
0.19  77.4397851455829\\
0.2  78.87507077468847\\
0.21  80.17807663207684\\
0.22  81.3417878813691\\
0.23  82.40353347329602\\
0.24  83.33951307014335\\
0.25  84.24593014225086\\
0.26  85.06491228949228\\
0.27  85.84932131458034\\
0.28  86.5816201265677\\
0.29  87.24903170204983\\
0.3  87.8896365813696\\
0.31  88.52322662431041\\
0.32  89.1357721581144\\
0.33  89.67892091774102\\
0.34  90.25739438913301\\
0.35  90.83461521117164\\
0.36  91.37125019416065\\
0.37  91.84625482896325\\
0.38  92.21553585834039\\
0.39  92.57629887211452\\
0.4  92.92428486248416\\
0.41  93.2304323644508\\
0.42  93.49599402736789\\
0.43  93.75303767468195\\
0.44  94.00056118691032\\
0.45  94.22253065232967\\
0.46  94.41243229430245\\
0.47  94.59782439860305\\
0.48  94.75440556777585\\
0.49  94.91474468500878\\
0.5  95.04101173982974\\
0.51  95.16752932452135\\
0.52  95.29379637934231\\
0.53  95.42732880041287\\
0.54  95.53179975648497\\
0.55  95.63451700346232\\
0.56  95.75652505048177\\
0.57  95.88504687413881\\
0.58  96.03386161732064\\
0.59  96.16338556046037\\
0.6  96.29215791398809\\
0.61  96.4314525220842\\
0.62  96.58151991461942\\
0.63  96.73008412793057\\
0.64  96.87564198279361\\
0.65  97.00792175451079\\
0.66  97.132435100237\\
0.67  97.24968307971359\\
0.68  97.34187807212254\\
0.69  97.4498564463841\\
0.7  97.55031892452537\\
0.71  97.64426762602905\\
0.72  97.74848805223047\\
0.73  97.85320953817323\\
0.74  97.95091618773706\\
0.75  98.04561647885276\\
0.76  98.14733160634742\\
0.77  98.24178136759247\\
0.78  98.3362311288375\\
0.79  98.42792506150508\\
0.8  98.51460839675914\\
0.81  98.587512589126\\
0.82  98.64538498925228\\
0.83  98.68471817894847\\
0.84  98.7217965998086\\
0.85  98.75236124403114\\
0.86  98.77991952980554\\
0.87  98.81699795066565\\
0.88  98.84931630398293\\
0.89  98.88589366510169\\
0.9  98.92723109376331\\
0.91  98.97608441854521\\
0.92  99.02067873552564\\
0.93  99.04873808104139\\
0.94  99.08030484474664\\
0.95  99.1148779669\\
0.96  99.14869949944132\\
0.97  99.174504076121\\
0.98  99.20506872034352\\
0.99  99.23538283469539\\
1.0  99.26444429969385\\
};
\addlegendentry{\textcolor{black}{DPFM~\normalsize{(\textit{Unsup})}~\large{(\textbf{84.74})} }}
\end{axis}
\end{tikzpicture}

%% file: vis/tikz_vis/ablation_rotation_supervised.tex
\newcommand{\pckLineWidth}{2pt}
\newcommand{\plotWidth}{1*\linewidth}
\newcommand{\plotHeight}{0.7*\linewidth}
\newcommand{\pckTitle}{\textsc{Supervised}}

\pgfplotsset{%
    label style = {font=\large},
    tick label style = {font=\large},
    title style =  {font=\LARGE},
    legend style={  fill= gray!10,
                    fill opacity=0.6, 
                    font=\large,
                    draw=gray!20, %
                    text opacity=1}
}
\begin{tikzpicture}[scale=0.45, transform shape]
\begin{axis}[
    width=\plotWidth,
    height=\plotHeight,
    grid=major,
    title=\pckTitle,
    legend style={
        at={(0.99,0.01)},
        anchor=south east,
        legend columns=1},
    legend cell align={left},
	ylabel={{\Large
        $\%$ Samples $<$ Geo Dist Thresh}},
        xmin=0,
        xmax=1,
        xlabel=\Large
        Geodesic Distance Threshold,
        ylabel near ticks,
        xtick={0, 0.2, 0.4, 0.6, 0.8, 1}, %
        xticklabels={$0$, $0.2$, $0.4$, $0.6$, $0.8$, $1$}, %
        ymin=0,
        ymax=100,
        ytick={0, 20, 40, 60, 80, 100},
        yticklabels={$0$, $20$, $40$, $60$, $80$, $100$},
	]

\addplot [color=cPLOT4, solid, smooth, line width=\pckLineWidth]
table[row sep=crcr]{%
0.0  1.8133352039563677\\
0.01  4.400306648561708\\
0.02  12.094580036777785\\
0.03  21.98424668173186\\
0.04  32.47869243449897\\
0.05  42.29144640915536\\
0.06  50.76311398608057\\
0.07  57.918497622471534\\
0.08  63.64886735445467\\
0.09  68.11957289667647\\
0.1  71.79384398001774\\
0.11  74.69497988245138\\
0.12  77.17297083331246\\
0.13  79.08726957515144\\
0.14  80.63328940709602\\
0.15  81.8721596175912\\
0.16  82.85198194180691\\
0.17  83.61985599543034\\
0.18  84.20734854216668\\
0.19  84.70114291727002\\
0.2  85.10675077789524\\
0.21  85.43970497602429\\
0.22  85.69374226489023\\
0.23  85.88840397440588\\
0.24  86.04222931500122\\
0.25  86.19931154391539\\
0.26  86.32783336757241\\
0.27  86.4460834665317\\
0.28  86.58312330579176\\
0.29  86.69736492682021\\
0.3  86.83340264659756\\
0.31  86.98672692745156\\
0.32  87.17387274084689\\
0.33  87.38131147376703\\
0.34  87.63835512108109\\
0.35  87.87736061770644\\
0.36  88.14217069101149\\
0.37  88.44305706569394\\
0.38  88.76173106119442\\
0.39  89.11648135807232\\
0.4  89.53486624210205\\
0.41  89.9625207313468\\
0.42  90.3991942959359\\
0.43  90.8679356839716\\
0.44  91.2908301056735\\
0.45  91.6743913376792\\
0.46  91.99657275136914\\
0.47  92.27165454937192\\
0.48  92.5339593239702\\
0.49  92.73262951141665\\
0.5  92.90699830140747\\
0.51  93.05831834329608\\
0.52  93.18984652540122\\
0.53  93.30684397500714\\
0.54  93.43586685840552\\
0.55  93.56764557038136\\
0.56  93.69466421481432\\
0.57  93.83095246446234\\
0.58  93.97901561803214\\
0.59  94.12081552483502\\
0.6  94.26436914073265\\
0.61  94.42195242938816\\
0.62  94.57627882972487\\
0.63  94.77695325613674\\
0.64  94.9721160253937\\
0.65  95.17078621284016\\
0.66  95.36670057170916\\
0.67  95.56862764747441\\
0.68  95.75702611022312\\
0.69  95.89832495728466\\
0.7  96.03336055757929\\
0.71  96.16012867214158\\
0.72  96.2535763139039\\
0.73  96.33925752967527\\
0.74  96.41466702074888\\
0.75  96.48631856376235\\
0.76  96.55671745742245\\
0.77  96.6168446263848\\
0.78  96.6777233849592\\
0.79  96.74486539030049\\
0.8  96.80449149952149\\
0.81  96.87614304253496\\
0.82  96.93827445046273\\
0.83  96.9928899622702\\
0.84  97.05426978058594\\
0.85  97.12692344308212\\
0.86  97.20533929260385\\
0.87  97.2752371265226\\
0.88  97.32509257078722\\
0.89  97.3869734488443\\
0.9  97.44234055026381\\
0.91  97.50747831663969\\
0.92  97.56635283624867\\
0.93  97.63149060262455\\
0.94  97.6873587637854\\
0.95  97.73170255089514\\
0.96  97.77454315878083\\
0.97  97.81086999002892\\
0.98  97.84945159011309\\
0.99  97.8865300109732\\
1.0  97.92811796950551\\
};
\addlegendentry{\textcolor{black}{DPFM~\normalsize{(\textit{Sup})}~\large{(\textbf{85.44})}  }}

\addplot [color=cPLOT4, dashed, smooth, line width=\pckLineWidth]
table[row sep=crcr]{%
0.0  2.1781066956613238\\
0.01  5.150142551496415\\
0.02  13.926454451164213\\
0.03  25.018163415624045\\
0.04  36.45986260941892\\
0.05  47.1612460353648\\
0.06  56.24570967596466\\
0.07  63.84152482500488\\
0.08  69.9003392174449\\
0.09  74.70324736818371\\
0.1  78.2825676305386\\
0.11  81.08624741327908\\
0.12  83.20723329842617\\
0.13  84.81012341101429\\
0.14  86.04924415138017\\
0.15  87.03507919249212\\
0.16  87.76086422784188\\
0.17  88.26292608867755\\
0.18  88.64999473887272\\
0.19  88.94486839665893\\
0.2  89.20241310371435\\
0.21  89.37603030409316\\
0.22  89.54438637718776\\
0.23  89.69821171778311\\
0.24  89.79692048682965\\
0.25  89.89061865846266\\
0.26  89.96552708979492\\
0.27  90.03692810293771\\
0.28  90.1030679887963\\
0.29  90.20478311629095\\
0.3  90.31952579706079\\
0.31  90.46909212985464\\
0.32  90.65523582376727\\
0.33  90.89975297754751\\
0.34  91.19587928468711\\
0.35  91.54887587247028\\
0.36  91.93268763434665\\
0.37  92.29169693902604\\
0.38  92.61888895012952\\
0.39  92.93104916899242\\
0.4  93.23519243199367\\
0.41  93.5180406559874\\
0.42  93.82569133717813\\
0.43  94.06820425199297\\
0.44  94.2956853745672\\
0.45  94.48483542692794\\
0.46  94.6386607675233\\
0.47  94.78321650290364\\
0.48  94.8901927576825\\
0.49  95.00217960987489\\
0.5  95.10715162568833\\
0.51  95.21337629085515\\
0.52  95.33613592748664\\
0.53  95.45112913812714\\
0.54  95.56211387083681\\
0.55  95.67284807367582\\
0.56  95.77731902974791\\
0.57  95.89832495728466\\
0.58  96.03611638615672\\
0.59  96.19119437610546\\
0.6  96.37358012195793\\
0.61  96.55471321845704\\
0.62  96.73584631495613\\
0.63  96.91848259067929\\
0.64  97.09485561963554\\
0.65  97.24993360958428\\
0.66  97.38722397871499\\
0.67  97.49219599452843\\
0.68  97.58088356874791\\
0.69  97.65929941826964\\
0.7  97.7427258652049\\
0.71  97.81813535627852\\
0.72  97.92260631235062\\
0.73  98.00553169954453\\
0.74  98.09847828156552\\
0.75  98.18666479604363\\
0.76  98.27009124297889\\
0.77  98.33998907689764\\
0.78  98.40612896275623\\
0.79  98.47302043822685\\
0.8  98.5597037734809\\
0.81  98.64087545158009\\
0.82  98.72355030890333\\
0.83  98.81148629351077\\
0.84  98.8986706885062\\
0.85  98.97332858996778\\
0.86  99.02694198229256\\
0.87  99.07078470966093\\
0.88  99.12289492276165\\
0.89  99.16723870987137\\
0.9  99.208576138533\\
0.91  99.2586821126683\\
0.92  99.30202378029533\\
0.93  99.34636756740505\\
0.94  99.38244386878247\\
0.95  99.41150533378095\\
0.96  99.43204878317641\\
0.97  99.45184064295987\\
0.98  99.46912720403654\\
0.99  99.49668548981096\\
1.0  99.51672787946507\\
};
\addlegendentry{\textcolor{black}{DPFM~\normalsize{(\textit{Sup})}~\large{(\textbf{89.48})} }}

\addplot [color=cPLOT3, solid, smooth, line width=\pckLineWidth]
table[row sep=crcr]{%
0.0  1.2116244962348561\\
0.01  3.220769297132082\\
0.02  8.943664575604826\\
0.03  16.720812432985372\\
0.04  25.51152959735762\\
0.05  34.52766240648763\\
0.06  43.15458657365755\\
0.07  50.97906062436066\\
0.08  57.812765747667584\\
0.09  63.86866981352986\\
0.1  68.89436645756048\\
0.11  72.96959538569615\\
0.12  76.26690001109208\\
0.13  78.80809475098289\\
0.14  80.76301162203133\\
0.15  82.25156829638028\\
0.16  83.37248425541355\\
0.17  84.21733075339849\\
0.18  84.88433429053845\\
0.19  85.4476885344902\\
0.2  85.91441846707501\\
0.21  86.30781744906888\\
0.22  86.63910080232688\\
0.23  86.901613280913\\
0.24  87.13047979393387\\
0.25  87.32705603963569\\
0.26  87.4940534145109\\
0.27  87.6508214298919\\
0.28  87.81165654001158\\
0.29  87.96226229063706\\
0.3  88.11114260713097\\
0.31  88.28602768089328\\
0.32  88.44143999802807\\
0.33  88.58378831388112\\
0.34  88.72367172383194\\
0.35  88.86281566201211\\
0.36  89.00245258137271\\
0.37  89.14960746373507\\
0.38  89.29996672377032\\
0.39  89.46018560741443\\
0.4  89.61362599982746\\
0.41  89.8024377919373\\
0.42  90.01515917129863\\
0.43  90.25462477969903\\
0.44  90.52724337248425\\
0.45  90.85174823451115\\
0.46  91.19178200372201\\
0.47  91.55116528426527\\
0.48  91.89575912939524\\
0.49  92.2325885209332\\
0.5  92.56966440306141\\
0.51  92.91610692761803\\
0.52  93.23531224195516\\
0.53  93.53060796904077\\
0.54  93.8100050530571\\
0.55  94.05131934088416\\
0.56  94.2854854015948\\
0.57  94.49352345974192\\
0.58  94.68775804483663\\
0.59  94.85574138207275\\
0.6  95.00930501978087\\
0.61  95.14733975030502\\
0.62  95.271817498367\\
0.63  95.38828430224676\\
0.64  95.49809586019053\\
0.65  95.59558288862323\\
0.66  95.7134053907492\\
0.67  95.83541823290895\\
0.68  95.96593500043136\\
0.69  96.09669825854398\\
0.7  96.21304181712863\\
0.71  96.34318884876569\\
0.72  96.4578069732188\\
0.73  96.55960758698036\\
0.74  96.6564783889375\\
0.75  96.74915885086087\\
0.76  96.8255709338296\\
0.77  96.91344482924364\\
0.78  96.99910030934569\\
0.79  97.07686809056064\\
0.8  97.14798062583961\\
0.81  97.22032561406968\\
0.82  97.29562848938241\\
0.83  97.35577219339652\\
0.84  97.4167786144764\\
0.85  97.48542624385314\\
0.86  97.54704889140856\\
0.87  97.59905840594534\\
0.88  97.64946573164569\\
0.89  97.69777788732915\\
0.9  97.74572030712727\\
0.91  97.79588114223739\\
0.92  97.8412354108382\\
0.93  97.88683617002921\\
0.94  97.93366938217135\\
0.95  97.97372410308236\\
0.96  98.00773980453297\\
0.97  98.03929060008134\\
0.98  98.07133437681016\\
0.99  98.10017377586611\\
1.0  98.13406623202158\\
};
\addlegendentry{\textcolor{black}{GeoFMaps~(\textbf{85.35}) }}

\addplot [color=cPLOT3, dashed, smooth, line width=\pckLineWidth]
table[row sep=crcr]{%
0.0  1.3087417887822133\\
0.01  3.4782287186186664\\
0.02  9.674139439726888\\
0.03  18.153415743354\\
0.04  27.73894181589618\\
0.05  37.57970889461294\\
0.06  46.92441366050851\\
0.07  55.32259455995268\\
0.08  62.47686069584294\\
0.09  68.65909118919386\\
0.1  73.64571907467432\\
0.11  77.5937588582556\\
0.12  80.79825977643303\\
0.13  83.27857134053907\\
0.14  85.16779846929343\\
0.15  86.70343484637473\\
0.16  87.87389541404256\\
0.17  88.76557512416964\\
0.18  89.46252726802155\\
0.19  90.05459766573412\\
0.2  90.54770209147266\\
0.21  90.95810892419182\\
0.22  91.28335325798938\\
0.23  91.53230875411332\\
0.24  91.74182575580177\\
0.25  91.89600561998546\\
0.26  92.02553642514697\\
0.27  92.11057567877347\\
0.28  92.18329040288887\\
0.29  92.24417357867364\\
0.3  92.30776815095084\\
0.31  92.37234868558892\\
0.32  92.43347835196391\\
0.33  92.48055805469626\\
0.34  92.5208592661975\\
0.35  92.5535192694019\\
0.36  92.59258802795203\\
0.37  92.63634010771639\\
0.38  92.69130750933583\\
0.39  92.75872268576147\\
0.4  92.84795227942173\\
0.41  92.94408360960821\\
0.42  93.06708241412885\\
0.43  93.21004695645743\\
0.44  93.3787697654642\\
0.45  93.58064555885579\\
0.46  93.8042125241869\\
0.47  94.07411972047967\\
0.48  94.38284918473238\\
0.49  94.7145022738757\\
0.5  95.06254698726876\\
0.51  95.43943109971777\\
0.52  95.81730117452766\\
0.53  96.15758143432875\\
0.54  96.48430471166763\\
0.55  96.75482813443597\\
0.56  96.98850121396616\\
0.57  97.18680289379952\\
0.58  97.3555257028063\\
0.59  97.50280383046376\\
0.6  97.62654210675508\\
0.61  97.73167034348464\\
0.62  97.81214952119203\\
0.63  97.88806862298031\\
0.64  97.96176930945661\\
0.65  98.02671957998004\\
0.66  98.09339528463501\\
0.67  98.15107408274689\\
0.68  98.21072480558055\\
0.69  98.26963605664353\\
0.7  98.32435696767276\\
0.71  98.37747568986553\\
0.72  98.4303479214681\\
0.73  98.49024513489198\\
0.74  98.54940287654519\\
0.75  98.6175575247415\\
0.76  98.70247353307288\\
0.77  98.77691369131983\\
0.78  98.86404811496321\\
0.79  98.94169265088306\\
0.8  99.00491748727492\\
0.81  99.06666338012546\\
0.82  99.12163078174491\\
0.83  99.16932671095282\\
0.84  99.2157901872096\\
0.85  99.2611444558104\\
0.86  99.30465004498453\\
0.87  99.34569072825646\\
0.88  99.38278756208481\\
0.89  99.42050062238874\\
0.9  99.45439307854423\\
0.91  99.48002809992728\\
0.92  99.507511800737\\
0.93  99.5370906715636\\
0.94  99.56593007061956\\
0.95  99.59033263905151\\
0.96  99.61399573571279\\
0.97  99.64714872009762\\
0.98  99.675371892678\\
0.99  99.69854200815884\\
1.0  99.71345468886726\\
};
\addlegendentry{\textcolor{black}{GeoFMaps~(\textbf{88.20})  }}

\end{axis}
\end{tikzpicture}

%% file: tab/tab_chirality.tex
\begin{table}[tbh!]
    \centering
    \footnotesize
    \begin{tabular}{@{}llccc}
    \toprule
                                                                            & Methods                                           & Left/Right Accuracy ($\uparrow$)  \\
    \midrule
    \parbox[t]{5mm}{\multirow{4}{*}{F2F}}                                   & GeomFmaps \footnotesize{(\textit{Sup})}           & 88.09                           \\     
                                                                            & DPFM \footnotesize{(\textit{Sup})}                & \textbf{89.05}                    \\
                                                                            & DPFM \footnotesize{(\textit{Unsup})}              & 86.92                           \\
                                                                            & ULRSSM \footnotesize{(\textit{Unsup})}            & 82.13                             \\
                                                                            & Smooth Shells \footnotesize{(\textit{Axio})}      & 72.88                             \\
                                                                            
    \midrule
    \parbox[t]{5mm}{\multirow{5}{*}{P2F}}                                   & GeomFmaps \footnotesize{(\textit{Sup})}           & 78.96                           \\
                                                                            & DPFM \footnotesize{(\textit{Sup})}                & \textbf{80.84}                             \\
                                                                            & DPFM \footnotesize{(\textit{Unsup})}              & 51.24                             \\
                                                                            & ULRSSM \footnotesize{(\textit{Unsup})}            & 53.39                            \\
                                                                            & PFM \footnotesize{(\textit{Axio})}                & 52.85                             \\
    \midrule
    \parbox[t]{5mm}{\multirow{3}{*}{P2P}}                                   & DPFM \footnotesize{(\textit{Sup})}                & \textbf{29.17}                            \\
                                                                            & SM-COMB \footnotesize{(\textit{Axio})}            & 14.23                           \\
                                                                            & GC-PPSM \footnotesize{(\textit{Axio})}            & 16.82                           \\
    \bottomrule
    \end{tabular}
    \caption{\textbf{Chirality benchmark:} While methods in the full-to-full setting and supervised methods in the partial-to-full setting perform well, unsupervised methods in the partial-to-full setting and all methods in the partial-to-partial setting perform poorly.
    }
    \label{tab:chirality_benchmark}
\end{table}

%% file: sec/05_limitations.tex
\section{Discussion and Future Work}
\label{sec:lim}
The cross-category and cross-dataset correspondences of \bm{} are generated pairwise for the selected template shapes. Although they are of good quality, for cross-category shapes they may give rise to correspondence ambiguity (e.g. hooves to paws). Ambiguities may potentially accumulate if many pairwise ground truth correspondences are chained (i.e.~for shapes that are distant in our shape network of shapes), which deteriorates the composed ground truth quality (see left example in Figure~\ref{fig:additional_custom_annotations} in the supplementary material). This could potentially be improved by incorporating cycle-consistency or synchronisation techniques~\cite{huang2013consistent,pachauri2013solving,bernard2021sparse} in the process of generating ground truth correspondences.

%% file: sec/06_conclusion.tex
\section{Conclusion}
While several shape matching datasets exist, they are typically static and limited in size, which constrains their adaptability to diverse problem settings and custom use cases.
To address this we connect multiple existing mesh-based datasets through manual cross-category annotations, enabling dense correspondence and annotation propagation across a diverse set of shapes.
Building on this foundation, we introduced a scalable framework for generating unlimited training instances for partial shape correspondence, simulating realistic occlusions and semantic variability.
As a concrete instantiation, we release a challenging non-rigid shape surface matching benchmark and provided a comprehensive evaluation of current state-of-the-art methods.
Our findings highlight persistent challenges, particularly in partial shape matching, and underscore the need for more robust and generalizable solutions.
We hope our benchmark and framework will foster future advancements in shape matching research.

\subsubsection*{Acknowledgments}
We thank Paul Roetzer and Daniel Scholz for their insightful feedback. This work is supported by the ERC starting grant no. 101160648 (Harmony), the Deutsche Forschungsgemeinschaft (DFG, German Research Foundation) – 458610525 and the ERC Advanced Grant grant no. 884679 (SIMULACRON). Open Access funding enabled and organized by Projekt DEAL.

%% file: supplementary/main_supplementary.tex
\clearpage

\twocolumn[
        \centering
        \Large
        \textbf{Beyond Complete Shapes: A Benchmark for Quantitative Evaluation \\ of 3D Shape Surface Matching Algorithms}\\
        \vspace{0.5em}Supplementary Material \\
        \vspace{1.0em}
]

The supplementary material provides more information about our framework and the \bm{} benchmark. In Sec.~\ref{sec:additional_results}, we show additional results on the partial-to-partial (P2P) setting and qualitative results of custom annotation propagation.
In Sec.~\ref{sec:procedure_data}, we provide details on the procedure followed to create our framework.
Additionally, we present details about the \bm{} benchmark in Sec.~\ref{sec:benchmark_details}.
We discuss the licenses of our framework and the assets used to create the \bm{} benchmark and our long-term preservation plan in Sec.~\ref{sec:further_distribution_information}.
In Sec.~\ref{sec:compute_info}, we provide information about the computation resources used for the baseline methods. 

\input{supplementary/sec/additional_results}
\input{supplementary/sec/procedural_generation}
\input{supplementary/sec/details_benchmark}
\input{supplementary/sec/further_distribution}

%% file: supplementary/sec/additional_results.tex
\section{The \bm{} Benchmark: Additional Results}
\label{sec:additional_results}

\subsection{Alternative Energies on the Partial-to-Partial Setting}
\label{sec:p2p_alternative_energies}

In the main paper, we use XYZ coordinates for the axiomatic methods SM-COMB~\cite{roetzer2022scalable} and GC-PPSM~\cite{ehm2024partial} for the partial-to-partial setting.
While these methods work best with learned features (e.g. unsupervised features from ULRSSM~\cite{cao2023unsupervised}), the latter features are not reliable for our challenging benchmark dataset (see Fig.~\ref{fig:geo_errors} in the main paper for the performance of ULRSSM in the partial-to-full setting).
In our case, we found that XYZ coordinates as feature descriptors are most suited for our benchmark. 
We compare the XYZ features to the original energy formulation used for SM-COMB (a combination of bending and membrane energy) and the rotation-invariant SHOT features.
We show the results in Fig.~\ref{fig:energy_ablation_p2p} for the geodesic distance comparison and Table~\ref{tab:p2p_mIoU} for the overlapping region prediction.
Both show that XYZ as input features work best in the given setting.
This is the case because in \bm{} the shapes are rotated only around the z-axis, which leaves the z-coordinates as reliable information. 
While this is not always a realistic setting, as it would not work for completely random rotations, we choose this setting to be able to evaluate the axiomatic methods in a (reasonably) meaningful manner. 
As these methods will benefit from better features, we hope to encourage further research in this direction.

\begin{figure}[!ht]%
    \centering
      \input{vis/tikz_vis/ablation_energies_partial_partial}

\caption{We compare \textbf{different energy formulations} as input for SM-COMB and GC-PPSM. For both methods, XYZ coordinates work best as input features.}
\label{fig:energy_ablation_p2p}
\end{figure}

\begin{table}[tbh!]
    \centering
    \footnotesize
    \begin{tabular}{c|c|c|c}
    \toprule
    \multicolumn{2}{c}{\multirow{2}{*}{\textbf{Method}}} & \multicolumn{2}{|c}{\textbf{Metrics}}               \\
    \multicolumn{2}{c}{}                        & \multicolumn{1}{|c}{mIoU ($\uparrow$) } & F1 Score ($\uparrow$)        \\
    \midrule
                                 & XYZ          & \textbf{46.91}  & \textbf{60.61} \\
    \textbf{SM-COMB}             & Original     & 23.81          & 33.78         \\
                                 & Shot         & 24.47          & 34.29         \\
    \midrule
                                 & XYZ          & \textbf{49.24}           & \textbf{62.42} \\
    \textbf{GC-PPSM}             & Original     & 43.69                    & 53.72         \\
                                 & Shot         & 39.89                    & 51.20          \\
    \bottomrule
    \end{tabular}
    \caption{\textbf{Overlapping Region Prediction:} We show the mean IoU (x100) and F1 Score (x100) of the partial-to-partial shape matching methods. Out of the three options, XYZ features are the input features that suit both methods the best.} 
    \label{tab:p2p_mIoU}
\end{table}

\subsection{Custom Annotation Propagation}
\label{sec:supp_custom_annotation}
\begin{figure}[!ht]
    \centering
    \begin{tabular}{@{}c@{}c@{}c@{}c@{}}
    \setlength{\tabcolsep}{0pt}  %
        
        \adjustbox{valign=m}{\includegraphics[height=0.095\textheight]{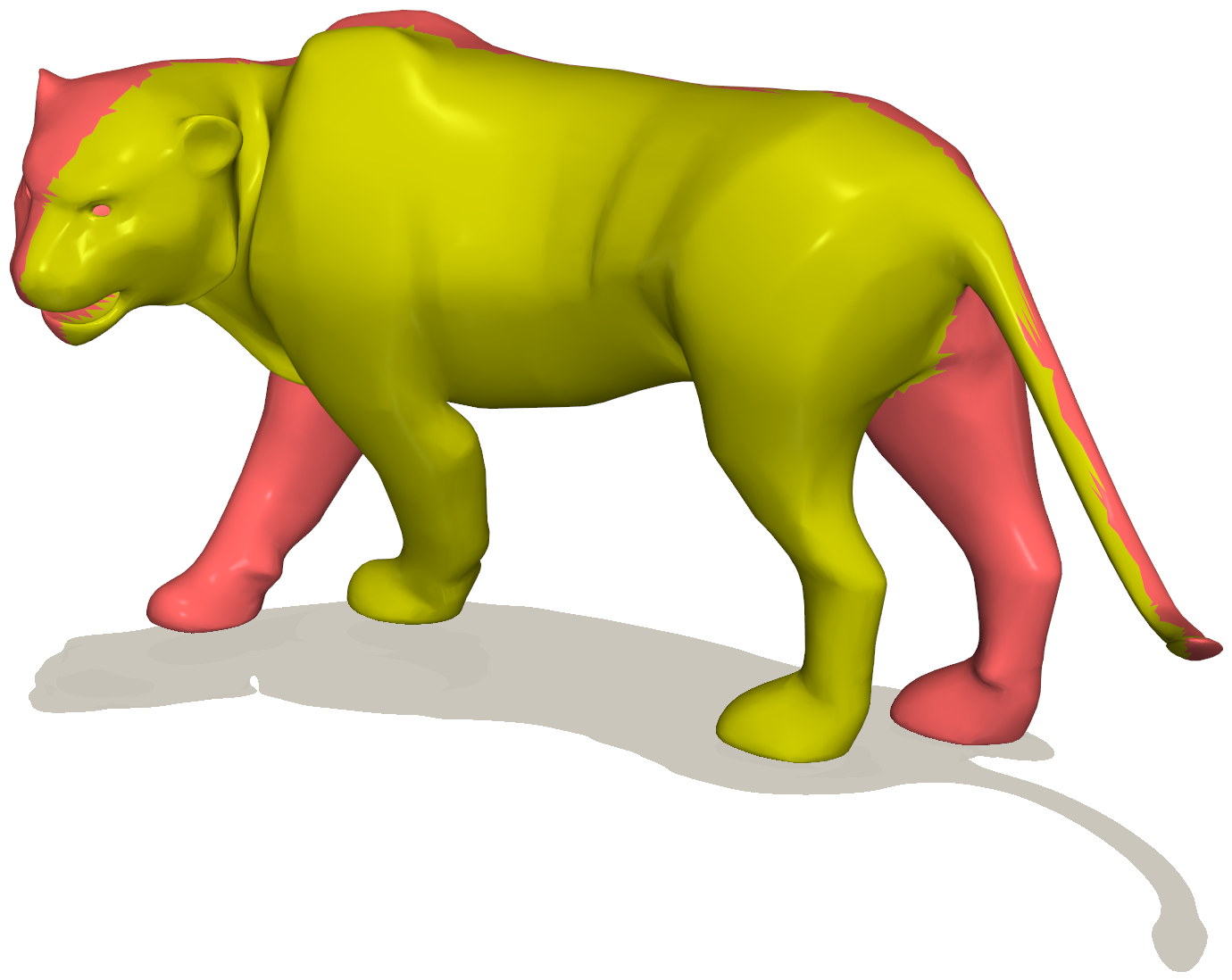}} &
        \adjustbox{valign=m}{\includegraphics[height=0.1\textheight]{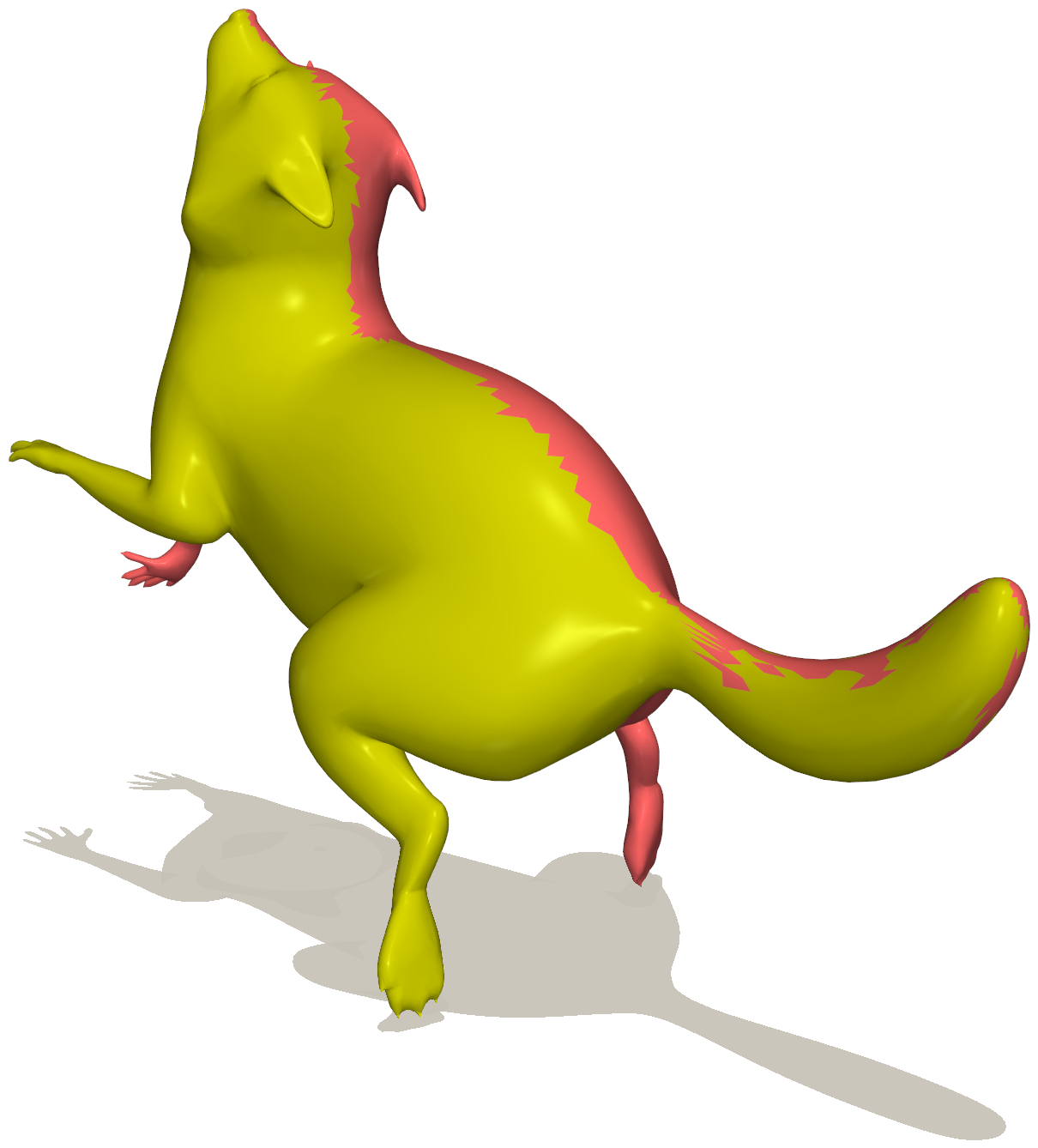}} &
        \adjustbox{valign=m}{\includegraphics[height=0.15\textheight]{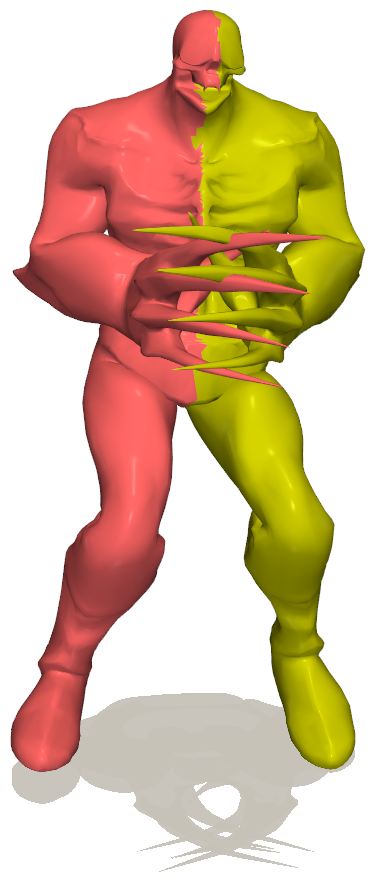}} &
        \adjustbox{valign=m}{\includegraphics[height=0.15\textheight]{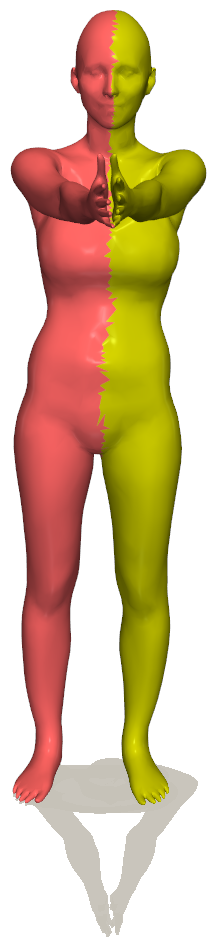}} \\

        \adjustbox{valign=m}{\includegraphics[height=0.08\textheight]{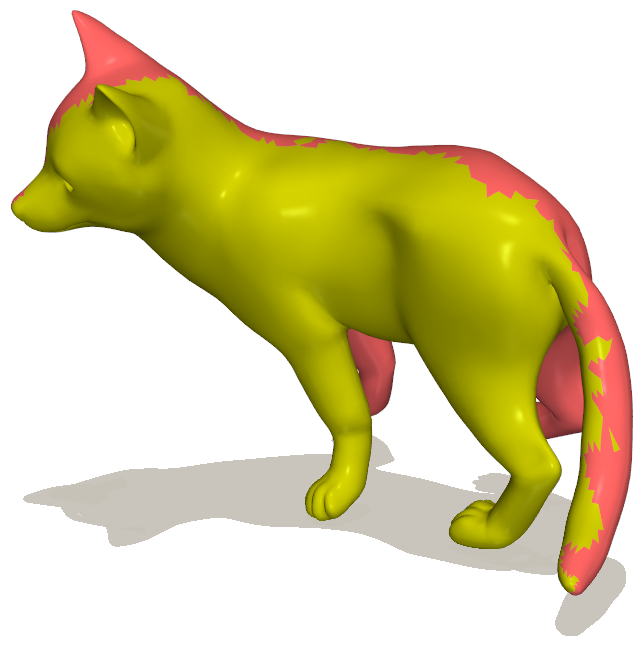}} &
        \adjustbox{valign=m}{\includegraphics[height=0.15\textheight]{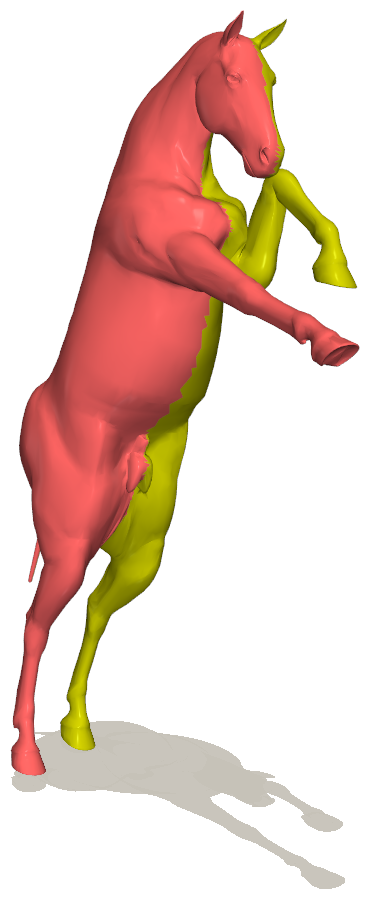}} &
        \adjustbox{valign=m}{\includegraphics[height=0.15\textheight]{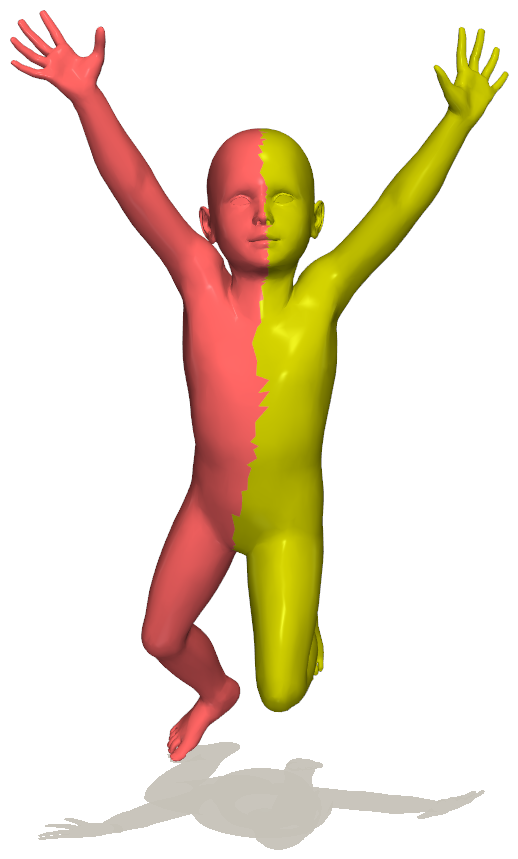}} &
        \adjustbox{valign=m}{\includegraphics[height=0.15\textheight]{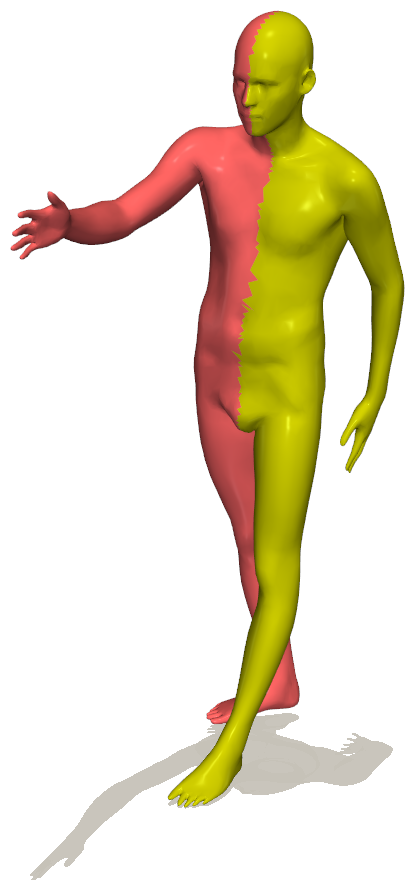}} \\
    \end{tabular}
    \caption{Our framework allows for the \textbf{propagation of custom annotations} throughout all shapes. We show the case of left/right annotations. The Left/Right annotations exhibit errors in thin parts of the shapes (e.g.~tails of animals). }
    \label{fig:additional_custom_annotations}
\end{figure}

We show more qualitative results of Left/Right annotation propagation using our framework in Fig.~\ref{fig:additional_custom_annotations}. We also include failure cases of the annotation propagation. This propagation relies on the manually created dense correspondence between templates in our framework; small errors caused by the manual annotation get amplified throughout the propagation and cause unwanted artefacts, especially in regions like the tails of animals. 

%% file: vis/tikz_vis/ablation_energies_partial_partial.tex
\newcommand{\pckLineWidth}{1pt}
\newcommand{\plotWidth}{1*\linewidth}
\newcommand{\plotHeight}{0.75*\linewidth}
\newcommand{\pckTitle}{\textsc{Partial-to-Partial - Energy Formulation}}

\pgfplotsset{%
    label style = {font=\footnotesize},
    tick label style = {font=\footnotesize},
    title style =  {font=\normalsize},
    legend style={  fill= gray!10,
                    fill opacity=0.6, 
                    font=\small,
                    draw=gray!20, %
                    text opacity=1}
}
\begin{tikzpicture}[scale=0.9, transform shape]
	\begin{axis}[
		width=\plotWidth,
		height=\plotHeight,
		grid=major,
		title=\pckTitle,
		legend style={
			at={(0.01,0.98)},
			anchor=north west,
			legend columns=1},
		legend cell align={left},
	ylabel={{\footnotesize
        $\%$ Samples $<$ Geo Dist Thresh}},
        xmin=0,
        xmax=1,
        xlabel=\footnotesize
        Geodesic Distance Threshold,
        ylabel near ticks,
        xtick={0, 0.25, 0.5, 0.75, 1, 1.25, 1.5},
        xticklabels={$0$, $0.25$, $0.5$, $0.75$, $1$, $1.25$,$1.5$},
        ymin=0,
        ymax=100,
        ytick={0, 20, 40, 60, 80, 100},
        yticklabels={$0$, $20$, $40$, $60$, $80$, $100$},
	]
\addplot [color=cPLOT1, solid, smooth, line width=\pckLineWidth] 
table[row sep=crcr]{%
0.0  0.05441521358679865\\
0.01  0.09806074948454338\\
0.02  0.28780546561142717\\
0.03  0.5815625952089105\\
0.04  0.9908103473929586\\
0.05  1.4968434924931096\\
0.06  2.067211291156819\\
0.07  2.708999057653202\\
0.08  3.4124290582908805\\
0.09  4.169990859944593\\
0.1  4.951926142684059\\
0.11  5.774672835613623\\
0.12  6.6142825765036815\\
0.13  7.447373828266153\\
0.14  8.321559902789488\\
0.15  9.146290483714404\\
0.16  9.969320589214735\\
0.17  10.800569659267238\\
0.18  11.611696436795455\\
0.19  12.412620361776147\\
0.2  13.167773156578361\\
0.21  13.922217419953661\\
0.22  14.675528033045907\\
0.23  15.432381303272708\\
0.24  16.15210752672935\\
0.25  16.89847453183786\\
0.26  17.60757278389083\\
0.27  18.302358701120188\\
0.28  19.00734747089707\\
0.29  19.6927807732912\\
0.3  20.347180399186605\\
0.31  21.01986013589633\\
0.32  21.67015027951565\\
0.33  22.274385880385726\\
0.34  22.86076648929763\\
0.35  23.46315990845774\\
0.36  24.03877083968059\\
0.37  24.61013058234198\\
0.38  25.180923499861834\\
0.39  25.732727775140468\\
0.4  26.26639364589016\\
0.41  26.80445241148671\\
0.42  27.324797891410473\\
0.43  27.83876658849203\\
0.44  28.33516370618619\\
0.45  28.795000602251715\\
0.46  29.26291475658403\\
0.47  29.750809497155245\\
0.48  30.191941163550307\\
0.49  30.60629034200812\\
0.5  31.013270793626052\\
0.51  31.41387446240178\\
0.52  31.785286636388754\\
0.53  32.12198077045708\\
0.54  32.49679389529322\\
0.55  32.825127358523986\\
0.56  33.14368308806337\\
0.57  33.44764307020838\\
0.58  33.75755471633944\\
0.59  34.04096728710402\\
0.6  34.30524950934199\\
0.61  34.60212417721788\\
0.62  34.87660925200337\\
0.63  35.133097628545315\\
0.64  35.34608217547489\\
0.65  35.582589965777935\\
0.66  35.814563154948736\\
0.67  36.04951217611257\\
0.68  36.250026569928515\\
0.69  36.464286473426526\\
0.7  36.683789509483695\\
0.71  36.89209774899565\\
0.72  37.10040598850762\\
0.73  37.28193174008233\\
0.74  37.47096792478231\\
0.75  37.65532780206466\\
0.76  37.805253051999124\\
0.77  37.97218305617946\\
0.78  38.130752389522236\\
0.79  38.29399803028263\\
0.8  38.46050291560682\\
0.81  38.59639924328844\\
0.82  38.74476572408369\\
0.83  38.87641086320384\\
0.84  39.00918965260704\\
0.85  39.127797813472014\\
0.86  39.25632541431375\\
0.87  39.38046012030863\\
0.88  39.50530335773043\\
0.89  39.618101560894736\\
0.9  39.7399689663235\\
0.91  39.865945854028354\\
0.92  39.97066679892587\\
0.93  40.06490147870509\\
0.94  40.15219255050057\\
0.95  40.241184097720655\\
0.96  40.336552427782934\\
0.97  40.42299326186613\\
0.98  40.49894783083104\\
0.99  40.58000382606971\\
1.0  40.657375457888435\\
1.01  40.72411911830349\\
1.02  40.78236040159561\\
1.03  40.84102680374388\\
1.04  40.899976618462915\\
1.05  40.95425012576433\\
1.06  40.99704542394978\\
1.07  41.019576723325564\\
1.08  41.04437532326746\\
1.09  41.0785265380446\\
1.1  41.12231378022772\\
1.11  41.16142471499323\\
1.12  41.20322806918101\\
1.13  41.24233900394652\\
1.14  41.274648037013684\\
1.15  41.30497318208549\\
1.16  41.339691222004156\\
1.17  41.363922996804526\\
1.18  41.386879415036454\\
1.19  41.40870218298532\\
1.2  41.421030629813586\\
1.21  41.438177090344844\\
1.22  41.45489843201995\\
1.23  41.467510291418975\\
1.24  41.47572925597115\\
1.25  41.49004159079476\\
1.26  41.494859604497755\\
1.27  41.50364539419146\\
1.28  41.51115582731672\\
1.29  41.51795772901507\\
1.3  41.52532645585495\\
1.31  41.53113641355562\\
1.32  41.533262007836356\\
1.33  41.537513196397825\\
1.34  41.54077244096162\\
1.35  41.54516533580847\\
1.36  41.548566286657646\\
1.37  41.55225065007758\\
1.38  41.55777719520749\\
1.39  41.56004449577361\\
1.4  41.56712981004272\\
1.41  41.56868857918193\\
1.42  41.56982222946499\\
1.43  41.57095587974805\\
1.44  41.57492365573875\\
1.45  41.57832460658793\\
1.46  41.57874972544407\\
1.47  41.58101702601019\\
1.48  41.58866916542083\\
1.49  41.59107817227233\\
};
\addlegendentry{\textcolor{black}{GC-PPSM (shot)(\textbf{26.02}) }}
\addplot [color=cPLOT2, solid, smooth, line width=\pckLineWidth] 
table[row sep=crcr]{%
0.0  0.11642088174636724\\
0.01  0.2166346801869757\\
0.02  0.6905568078528721\\
0.03  1.398131053176791\\
0.04  2.3357108224149363\\
0.05  3.4618329941100754\\
0.06  4.762046252314574\\
0.07  6.1362718576153705\\
0.08  7.483755775461666\\
0.09  8.829348866922288\\
0.1  10.177643138933872\\
0.11  11.482448404075\\
0.12  12.719859214469684\\
0.13  13.882987559712973\\
0.14  14.96562072453766\\
0.15  15.997606754031848\\
0.16  16.990290606509575\\
0.17  17.9190915389571\\
0.18  18.821015724922578\\
0.19  19.64663156032344\\
0.2  20.447126416600376\\
0.21  21.223175588891124\\
0.22  21.89941613982391\\
0.23  22.60901627056105\\
0.24  23.27458715831754\\
0.25  23.926246966236594\\
0.26  24.551975440866432\\
0.27  25.15609447108859\\
0.28  25.748598424941623\\
0.29  26.29018512540906\\
0.3  26.831501707821403\\
0.31  27.342700127090545\\
0.32  27.822429792941005\\
0.33  28.294190976166135\\
0.34  28.756633086490453\\
0.35  29.2101613009966\\
0.36  29.647752550252086\\
0.37  30.088720275196273\\
0.38  30.501865840465577\\
0.39  30.898399145346474\\
0.4  31.263733814863787\\
0.41  31.656350407945794\\
0.42  32.03127426841901\\
0.43  32.42780757329991\\
0.44  32.831769124695946\\
0.45  33.20588263100388\\
0.46  33.56986671024571\\
0.47  33.94371009849855\\
0.48  34.303642406913944\\
0.49  34.64007444453598\\
0.5  34.95381656552997\\
0.51  35.27471681498399\\
0.52  35.606691904696945\\
0.53  35.93231922011515\\
0.54  36.253219469569174\\
0.55  36.59991599328487\\
0.56  36.93931932951296\\
0.57  37.264271349793425\\
0.58  37.56180638748165\\
0.59  37.8605569564178\\
0.6  38.15214939689391\\
0.61  38.425373809623494\\
0.62  38.707512118171245\\
0.63  38.958316732327866\\
0.64  39.22181689507399\\
0.65  39.495311425858674\\
0.66  39.74935745667644\\
0.67  40.00434890068705\\
0.68  40.25461327873347\\
0.69  40.48029691376616\\
0.7  40.699092538393906\\
0.71  40.936526308823275\\
0.72  41.17706643688624\\
0.73  41.38289639486938\\
0.74  41.57454515495998\\
0.75  41.786857946265414\\
0.76  41.9691876334552\\
0.77  42.14746554981855\\
0.78  42.31939569188714\\
0.79  42.49267642423121\\
0.8  42.66730774685076\\
0.81  42.833835527817435\\
0.82  43.01657039208987\\
0.83  43.207138679960075\\
0.84  43.383255651882656\\
0.85  43.54640695716063\\
0.86  43.73238323809421\\
0.87  43.93969884538038\\
0.88  44.10744215759498\\
0.89  44.28842125450929\\
0.9  44.471426236836805\\
0.91  44.634982719197424\\
0.92  44.80866862862414\\
0.93  44.96061003461563\\
0.94  45.09094199619944\\
0.95  45.21384571126811\\
0.96  45.33013153398693\\
0.97  45.43331663103359\\
0.98  45.54555068292597\\
0.99  45.67709817575771\\
1.0  45.78190398113495\\
1.01  45.87901142194196\\
1.02  45.97665909885916\\
1.03  46.07930395979563\\
1.04  46.17776199087811\\
1.05  46.26757624419753\\
1.06  46.357795674599586\\
1.07  46.43599485154987\\
1.08  46.51784062224395\\
1.09  46.58320919157718\\
1.1  46.64047421925753\\
1.11  46.70557267053566\\
1.12  46.754734156563124\\
1.13  46.813890010629144\\
1.14  46.85535313208638\\
1.15  46.89627601743342\\
1.16  46.940035142358965\\
1.17  46.98609027075283\\
1.18  47.03943858663429\\
1.19  47.08468336086287\\
1.2  47.13789661771678\\
1.21  47.18989434332275\\
1.22  47.22338898215465\\
1.23  47.25715373904165\\
1.24  47.288352374405235\\
1.25  47.32319760351262\\
1.26  47.358583068730184\\
1.27  47.39194264853454\\
1.28  47.424491874173604\\
1.29  47.454610037316804\\
1.3  47.48540349559775\\
1.31  47.514441186520564\\
1.32  47.53983228369959\\
1.33  47.56144172810727\\
1.34  47.58062011001908\\
1.35  47.59871801971051\\
1.36  47.61262909954796\\
1.37  47.62708041549559\\
1.38  47.63599431131376\\
1.39  47.646528915462504\\
1.4  47.65895434599692\\
1.41  47.67110965847623\\
1.42  47.68015861332196\\
1.43  47.690288040388054\\
1.44  47.71243772090592\\
1.45  47.72553844657808\\
1.46  47.73188622087283\\
1.47  47.740800116691\\
1.48  47.75146977986729\\
1.49  47.75984343957526\\
};
\addlegendentry{\textcolor{black}{GC-PPSM (original energy)(\textbf{30.71} }}
\addplot [color=cPLOT3, solid, smooth, line width=\pckLineWidth] 
table[row sep=crcr]{%
0.0  0.0934718386832998\\
0.01  0.17937232077222331\\
0.02  0.5085198410841082\\
0.03  1.0045125285302483\\
0.04  1.7050695050535365\\
0.05  2.545352426000314\\
0.06  3.495351187051053\\
0.07  4.575577281525064\\
0.08  5.751615452175182\\
0.09  6.988775119696265\\
0.1  8.270261638551696\\
0.11  9.46529703120225\\
0.12  10.696674935505808\\
0.13  11.906027075171181\\
0.14  13.098171586212972\\
0.15  14.258516399558383\\
0.16  15.468006200252745\\
0.17  16.6496884730914\\
0.18  17.813612473190517\\
0.19  18.908568297766312\\
0.2  20.008892902472667\\
0.21  21.085952793279937\\
0.22  22.12804678272409\\
0.23  23.14302154945748\\
0.24  24.127022584668417\\
0.25  25.10331460225599\\
0.26  25.979664712797796\\
0.27  26.828069634454906\\
0.28  27.6292568231689\\
0.29  28.41640258692606\\
0.3  29.201758757306358\\
0.31  29.970044960092068\\
0.32  30.688222548325903\\
0.33  31.386989931472343\\
0.34  32.02821500450151\\
0.35  32.6876113333572\\
0.36  33.342602509285236\\
0.37  33.91898923766075\\
0.38  34.5253860703558\\
0.39  35.1092064942967\\
0.4  35.69784505425221\\
0.41  36.25372028930842\\
0.42  36.79954626924845\\
0.43  37.29664024492651\\
0.44  37.80034194999601\\
0.45  38.331300538805266\\
0.46  38.84023336297634\\
0.47  39.33402347395866\\
0.48  39.785276326983485\\
0.49  40.22000985652968\\
0.5  40.65254080961204\\
0.51  41.05987979438949\\
0.52  41.44712026893458\\
0.53  41.828991963349125\\
0.54  42.22724532021332\\
0.55  42.60994298080179\\
0.56  42.97130318189703\\
0.57  43.34408924839831\\
0.58  43.67502636208705\\
0.59  44.009405001500504\\
0.6  44.31556312997129\\
0.61  44.60933176583308\\
0.62  44.90681724947758\\
0.63  45.189297680962305\\
0.64  45.46819892569333\\
0.65  45.75976498509131\\
0.66  46.029718262938076\\
0.67  46.287970353320794\\
0.68  46.525435628326235\\
0.69  46.768682666549196\\
0.7  46.99912722907621\\
0.71  47.20493046741426\\
0.72  47.384853432302435\\
0.73  47.558581650886126\\
0.74  47.7117983761505\\
0.75  47.87492669550206\\
0.76  48.03750437073767\\
0.77  48.186178282045425\\
0.78  48.35247280506373\\
0.79  48.49811817373372\\
0.8  48.621462455707565\\
0.81  48.736547075942084\\
0.82  48.87310681669884\\
0.83  49.00760164202075\\
0.84  49.1478782305602\\
0.85  49.263651155939655\\
0.86  49.375982555594405\\
0.87  49.48335815820556\\
0.88  49.57958321746863\\
0.89  49.6770472259926\\
0.9  49.769142454386014\\
0.91  49.85284036001112\\
0.92  49.941081079592855\\
0.93  50.04198661384154\\
0.94  50.13256757091607\\
0.95  50.20910710303378\\
0.96  50.28027785502091\\
0.97  50.34291362321075\\
0.98  50.421931053850244\\
0.99  50.4990212300839\\
1.0  50.56716343943328\\
1.01  50.636544598043564\\
1.02  50.70799067208868\\
1.03  50.78191464465559\\
1.04  50.84812959959913\\
1.05  50.90814980823819\\
1.06  50.955367541181296\\
1.07  51.00162164692149\\
1.08  51.03700053137157\\
1.09  51.0766469077203\\
1.1  51.11354006348927\\
1.11  51.14410081192475\\
1.12  51.17328495007034\\
1.13  51.19737563014336\\
1.14  51.22146631021638\\
1.15  51.24294143073861\\
1.16  51.262489296854994\\
1.17  51.276943704898805\\
1.18  51.28919553647879\\
1.19  51.29952011365294\\
1.2  51.313010894493836\\
1.21  51.32429909887091\\
1.22  51.33600028633494\\
1.23  51.343709303958306\\
1.24  51.3570624237702\\
1.25  51.364909102422565\\
1.26  51.3741323913648\\
1.27  51.38087778178524\\
1.28  51.3851452736839\\
1.29  51.39037639278546\\
1.3  51.3935425964522\\
1.31  51.39629581703198\\
1.32  51.39877371555377\\
1.33  51.401113953046575\\
1.34  51.40331652951039\\
1.35  51.40606975009017\\
1.36  51.41171385227871\\
1.37  51.41474239491646\\
1.38  51.41845924269914\\
1.39  51.42135012430791\\
1.4  51.424516327974644\\
1.41  51.42740720958341\\
1.42  51.43029809119217\\
1.43  51.43332663382992\\
1.44  51.43910839704745\\
1.45  51.44489016026497\\
1.46  51.4481940249607\\
1.47  51.45287449994631\\
1.48  51.456729008757996\\
1.49  51.46003287345372\\
};
\addlegendentry{\textcolor{black}{GC-PPSM (XYZ)(\textbf{34.29}) }}
\addplot [color=cPLOT1, dashed, smooth, line width=\pckLineWidth]   
table[row sep=crcr]{%
0.0  0.022472715192532543\\
0.01  0.03778021684541703\\
0.02  0.12441416237025263\\
0.03  0.2305895993668557\\
0.04  0.3958780480655552\\
0.05  0.6036692407153489\\
0.06  0.838818521425617\\
0.07  1.0944863681812407\\
0.08  1.373441158940721\\
0.09  1.69196746993053\\
0.1  2.0015372639957794\\
0.11  2.347258817283798\\
0.12  2.7094277925605543\\
0.13  3.0715967678373106\\
0.14  3.4668234328538072\\
0.15  3.9035757672478093\\
0.16  4.337396877921046\\
0.17  4.797436156318905\\
0.18  5.275062777041353\\
0.19  5.731845140193917\\
0.2  6.223802187995662\\
0.21  6.710548171404936\\
0.22  7.218626949670889\\
0.23  7.746898602457668\\
0.24  8.273704643384065\\
0.25  8.798882226687816\\
0.26  9.318848745599093\\
0.27  9.835395503502813\\
0.28  10.388094020629302\\
0.29  10.903175166672638\\
0.3  11.405554343259325\\
0.31  11.930731926563075\\
0.32  12.415686606587437\\
0.33  12.972619113532808\\
0.34  13.491120020583706\\
0.35  14.016948986936512\\
0.36  14.53137874993079\\
0.37  15.032943697706155\\
0.38  15.532065959047546\\
0.39  16.014415106875674\\
0.4  16.462240953103677\\
0.41  16.9141379433883\\
0.42  17.383459430235245\\
0.43  17.839753256101016\\
0.44  18.296209927729052\\
0.45  18.742570162096673\\
0.46  19.174274277860466\\
0.47  19.596859030937438\\
0.48  20.014558411146467\\
0.49  20.39350049993649\\
0.5  20.75273825149248\\
0.51  21.135100101290064\\
0.52  21.48277580372526\\
0.53  21.842339246805782\\
0.54  22.204182530558008\\
0.55  22.54974123808376\\
0.56  22.889437498167986\\
0.57  23.216920326082356\\
0.58  23.535935174358958\\
0.59  23.870257524288448\\
0.6  24.184549845459372\\
0.61  24.481091978543443\\
0.62  24.775028579431275\\
0.63  25.064568344737964\\
0.64  25.352479652422005\\
0.65  25.607007578841777\\
0.66  25.86137265949928\\
0.67  26.124694257081348\\
0.68  26.347792951384026\\
0.69  26.58717622191318\\
0.7  26.806529463683766\\
0.71  27.032722227469474\\
0.72  27.24246756926644\\
0.73  27.43332280264071\\
0.74  27.624829419064028\\
0.75  27.795166086392936\\
0.76  27.9661541367709\\
0.77  28.11874061601295\\
0.78  28.257648051224766\\
0.79  28.399812401681874\\
0.8  28.523900872527598\\
0.81  28.648640726422375\\
0.82  28.775986112513397\\
0.83  28.90007458335912\\
0.84  29.029862655884102\\
0.85  29.139783545412794\\
0.86  29.25035581799055\\
0.87  29.350343116021094\\
0.88  29.442513817462928\\
0.89  29.53663866805194\\
0.9  29.614804633939013\\
0.91  29.69883304726761\\
0.92  29.776184784343357\\
0.93  29.86167880953234\\
0.94  29.937402088985436\\
0.95  30.020290581978188\\
0.96  30.08412612078596\\
0.97  30.143564824012586\\
0.98  30.194209856076913\\
0.99  30.24534342542804\\
1.0  30.284100716847046\\
1.01  30.31797263539811\\
1.02  30.3494018675152\\
1.03  30.371711736945468\\
1.04  30.405420809734267\\
1.05  30.44303818081742\\
1.06  30.487495073915692\\
1.07  30.515178853500696\\
1.08  30.540419946651724\\
1.09  30.56468396522917\\
1.1  30.58894798380662\\
1.11  30.61370053967086\\
1.12  30.633079185380357\\
1.13  30.65310921413892\\
1.14  30.67069655646351\\
1.15  30.6910522767466\\
1.16  30.709616693644783\\
1.17  30.723947120724077\\
1.18  30.748211139301524\\
1.19  30.761238800282698\\
1.2  30.7783376053205\\
1.21  30.790713883252618\\
1.22  30.805532847618704\\
1.23  30.81611782216591\\
1.24  30.825725722139534\\
1.25  30.83940476616977\\
1.26  30.849012666143388\\
1.27  30.855689342396243\\
1.28  30.862366018649094\\
1.29  30.865785779656658\\
1.3  30.870508306762336\\
1.31  30.87246245590951\\
1.32  30.8762079084416\\
1.33  30.88109328130954\\
1.34  30.88549011689069\\
1.35  30.88825849484919\\
1.36  30.890538335520894\\
1.37  30.89135256433222\\
1.38  30.89135256433222\\
1.39  30.89135256433222\\
1.4  30.89135256433222\\
1.41  30.891515410094485\\
1.42  30.893632405003924\\
1.43  30.895912245675632\\
1.44  30.89770354906054\\
1.45  30.90177469311716\\
1.46  30.90584583717378\\
1.47  30.917570732056838\\
1.48  30.921316184588925\\
1.49  30.9265272489814\\
};
\addlegendentry{\textcolor{black}{SM-COMB (shot) (\textbf{17.79})  }}
\addplot [color=cPLOT2, dashed, smooth, line width=\pckLineWidth]
table[row sep=crcr]{%
0.0  0.02337215963337789\\
0.01  0.05911781554324995\\
0.02  0.1816307046018713\\
0.03  0.3566927630322704\\
0.04  0.5800267328623258\\
0.05  0.8527019285850679\\
0.06  1.1569982814588506\\
0.07  1.4748902043154477\\
0.08  1.8271529501623067\\
0.09  2.19224746992553\\
0.1  2.565590987206416\\
0.11  2.909604735535612\\
0.12  3.2800458277639866\\
0.13  3.6576666030169944\\
0.14  4.035440137483292\\
0.15  4.438724460569028\\
0.16  4.839717395455414\\
0.17  5.24330723696773\\
0.18  5.65911781554325\\
0.19  6.064999045254917\\
0.2  6.491044491120871\\
0.21  6.899369868245178\\
0.22  7.309222837502387\\
0.23  7.718464769906435\\
0.24  8.14985678823754\\
0.25  8.544433836165744\\
0.26  8.934428107695245\\
0.27  9.336948634714531\\
0.28  9.741607790719877\\
0.29  10.149627649417607\\
0.3  10.532442237922474\\
0.31  10.927019285850678\\
0.32  11.314263891540959\\
0.33  11.689135000954746\\
0.34  12.074546496085544\\
0.35  12.441474126408249\\
0.36  12.855451594424288\\
0.37  13.248959327859462\\
0.38  13.615428680542296\\
0.39  14.004659156005347\\
0.4  14.385946152377315\\
0.41  14.754095856406341\\
0.42  15.14057666603017\\
0.43  15.49726942906244\\
0.44  15.881153332060341\\
0.45  16.26427343899179\\
0.46  16.613481000572847\\
0.47  16.9243459996181\\
0.48  17.24850105021959\\
0.49  17.57021195340844\\
0.5  17.862287569219017\\
0.51  18.146725224365095\\
0.52  18.413748329196107\\
0.53  18.693908726370058\\
0.54  18.975902234103494\\
0.55  19.265991980141305\\
0.56  19.538514416650756\\
0.57  19.81959136910445\\
0.58  20.08676723314875\\
0.59  20.36448348291006\\
0.6  20.618674813824708\\
0.61  20.894405193813252\\
0.62  21.15913691044491\\
0.63  21.41974412831774\\
0.64  21.651785373305326\\
0.65  21.8902425052511\\
0.66  22.095092610273056\\
0.67  22.291082680924195\\
0.68  22.483406530456367\\
0.69  22.646553370250146\\
0.7  22.810922283750237\\
0.71  22.969333587932024\\
0.72  23.118732098529694\\
0.73  23.279434790910827\\
0.74  23.41539049073897\\
0.75  23.543860989115906\\
0.76  23.683024632423145\\
0.77  23.83028451403475\\
0.78  23.9492839411877\\
0.79  24.076074088218448\\
0.8  24.20301699446248\\
0.81  24.32415505060149\\
0.82  24.457666603016996\\
0.83  24.586900897460378\\
0.84  24.703150658774106\\
0.85  24.82673286232576\\
0.86  24.932900515562345\\
0.87  25.047622684743175\\
0.88  25.15424861561963\\
0.89  25.24972312392591\\
0.9  25.32686652663739\\
0.91  25.4310483101012\\
0.92  25.527286614473937\\
0.93  25.614664884475847\\
0.94  25.689669658201257\\
0.95  25.789421424479663\\
0.96  25.888256635478328\\
0.97  25.97639870154669\\
0.98  26.05201451212526\\
0.99  26.15466870345618\\
1.0  26.23242314302081\\
1.01  26.302692381134236\\
1.02  26.359824326904718\\
1.03  26.40519381325186\\
1.04  26.45331296543823\\
1.05  26.49868245178537\\
1.06  26.539163643307234\\
1.07  26.5765896505633\\
1.08  26.61447393545923\\
1.09  26.644261982050793\\
1.1  26.665342753484822\\
1.11  26.687492839411874\\
1.12  26.707809814779456\\
1.13  26.731793011265992\\
1.14  26.747679969448157\\
1.15  26.765400038189803\\
1.16  26.78357838457132\\
1.17  26.80756158105786\\
1.18  26.824365094519763\\
1.19  26.84162688562154\\
1.2  26.856750047737254\\
1.21  26.878594615237734\\
1.22  26.89325949971358\\
1.23  26.908993698682455\\
1.24  26.921978231812105\\
1.25  26.93297689516899\\
1.26  26.946266946725224\\
1.27  26.95650181401566\\
1.28  26.965209089173193\\
1.29  26.972999809050986\\
1.3  26.978193622302843\\
1.31  26.986137101393926\\
1.32  26.990872637005918\\
1.33  26.994691617338173\\
1.34  26.999274393736872\\
1.35  27.004926484628605\\
1.36  27.0111896123735\\
1.37  27.018216536184838\\
1.38  27.02432690471644\\
1.39  27.028298644261984\\
1.4  27.03150658774107\\
1.41  27.033186939087262\\
1.42  27.035172808860036\\
1.43  27.035936604926487\\
1.44  27.039602826045446\\
1.45  27.04296352873783\\
1.46  27.04296352873783\\
1.47  27.04525491693718\\
1.48  27.049226656482716\\
1.49  27.049226656482716\\
    };
\addlegendentry{\textcolor{black}{SM-COMB (original energy) (\textbf{15.50})  }}
\addplot [color=cPLOT3, dashed, smooth, line width=\pckLineWidth]   
table[row sep=crcr]{%
0.0  0.054296667525597\\
0.01  0.1284451432748384\\
0.02  0.38336339589501417\\
0.03  0.7731687692930653\\
0.04  1.2893157831078204\\
0.05  1.9258883355507759\\
0.06  2.634768798742105\\
0.07  3.409120788392871\\
0.08  4.294169615952965\\
0.09  5.233778048895924\\
0.1  6.176410267197451\\
0.11  7.1387628247939885\\
0.12  8.115051088825668\\
0.13  9.092785511072313\\
0.14  10.070388464390327\\
0.15  11.051015203066907\\
0.16  12.018232111022881\\
0.17  12.986500770407922\\
0.18  13.921902197634612\\
0.19  14.832718935206854\\
0.2  15.71329781919341\\
0.21  16.59729489532444\\
0.22  17.450791180012516\\
0.23  18.267739102540506\\
0.24  19.070882787561988\\
0.25  19.839976020067418\\
0.26  20.608148970072413\\
0.27  21.35278898185203\\
0.28  22.057725377184354\\
0.29  22.73741973821907\\
0.3  23.444196698552265\\
0.31  24.07340699099175\\
0.32  24.716815927723644\\
0.33  25.331564638013454\\
0.34  25.921991596506082\\
0.35  26.535820024295457\\
0.36  27.146361728869\\
0.37  27.75585168201347\\
0.38  28.322877171209353\\
0.39  28.904627180412177\\
0.4  29.477042895682033\\
0.41  30.053271209882254\\
0.42  30.63186596479788\\
0.43  31.158530492903306\\
0.44  31.681382422078364\\
0.45  32.203839944467525\\
0.46  32.710126788634774\\
0.47  33.21785979101699\\
0.48  33.7007451658875\\
0.49  34.17258715075279\\
0.5  34.61708359846234\\
0.51  35.03186806830074\\
0.52  35.47255191707992\\
0.53  35.924410624792934\\
0.54  36.39822464358773\\
0.55  36.85560504630336\\
0.56  37.29037279329403\\
0.57  37.711073364920935\\
0.58  38.10179902081942\\
0.59  38.49646874457691\\
0.6  38.885485304403154\\
0.61  39.250180112432226\\
0.62  39.60961616331596\\
0.63  39.96011232705263\\
0.64  40.297461597925945\\
0.65  40.64296194237454\\
0.66  40.971897201815324\\
0.67  41.301752743756545\\
0.68  41.624640432480184\\
0.69  41.939114109771296\\
0.7  42.253324849205136\\
0.71  42.556097791847876\\
0.72  42.8411224291251\\
0.73  43.12312328104376\\
0.74  43.38277441509474\\
0.75  43.634931820213616\\
0.76  43.878675213899946\\
0.77  44.11729131936958\\
0.78  44.348413695907105\\
0.79  44.582954264589105\\
0.8  44.799352121119696\\
0.81  45.01745907372251\\
0.82  45.20243585630972\\
0.83  45.38833292139736\\
0.84  45.58619365899064\\
0.85  45.77432569586504\\
0.86  45.95851366488044\\
0.87  46.14112400675224\\
0.88  46.3034881336145\\
0.89  46.4864928822722\\
0.9  46.65661367592383\\
0.91  46.82292187064509\\
0.92  46.979106957861575\\
0.93  47.11820108435572\\
0.94  47.2657092222824\\
0.95  47.40454041091928\\
0.96  47.52351979133252\\
0.97  47.64591736389022\\
0.98  47.77922685752449\\
0.99  47.90543702901257\\
1.0  48.02178703085313\\
1.01  48.14392166555356\\
1.02  48.25212059381886\\
1.03  48.36110833565595\\
1.04  48.457475060344244\\
1.05  48.54279839502732\\
1.06  48.631145515068965\\
1.07  48.72356817189825\\
1.08  48.8170425801566\\
1.09  48.90078828769609\\
1.1  48.97901230023296\\
1.11  49.05145167990997\\
1.12  49.11981552279934\\
1.13  49.180291229970706\\
1.14  49.238137558569406\\
1.15  49.28835868930737\\
1.16  49.33424134540043\\
1.17  49.377363153992185\\
1.18  49.42061643151258\\
1.19  49.4604515168885\\
1.2  49.498446037263555\\
1.21  49.534205585851836\\
1.22  49.56746722479609\\
1.23  49.59823095409631\\
1.24  49.620054796249455\\
1.25  49.6422730451885\\
1.26  49.66646332805705\\
1.27  49.68210813056442\\
1.28  49.69748999521453\\
1.29  49.70787604057657\\
1.3  49.72128587129718\\
1.31  49.733906888445986\\
1.32  49.74836847059566\\
1.33  49.76046361202993\\
1.34  49.77676575918048\\
1.35  49.785574177398914\\
1.36  49.79753784990456\\
1.37  49.80910711562429\\
1.38  49.81949316098633\\
1.39  49.82646101420391\\
1.4  49.83211417813514\\
1.41  49.836978528494576\\
1.42  49.840265251710406\\
1.43  49.84355197492625\\
1.44  49.84815338742842\\
1.45  49.85341214457375\\
1.46  49.85525270957462\\
1.47  49.85893383957635\\
1.48  49.86314084529263\\
1.49  49.86655903743709\\
};
\addlegendentry{\textcolor{black}{SM-COMB (XYZ) (\textbf{30.03}} )}
\end{axis}
\end{tikzpicture}

%% file: supplementary/sec/procedural_generation.tex
\section{Details on the Procedural Data Generation}
\label{sec:procedure_data}

\subsection{Correspondence Generation}
\begin{figure*}[tbh!]
    \centering
    \begin{tabular}{@{}cc@{}}
    {\includegraphics[width=0.4\textwidth]{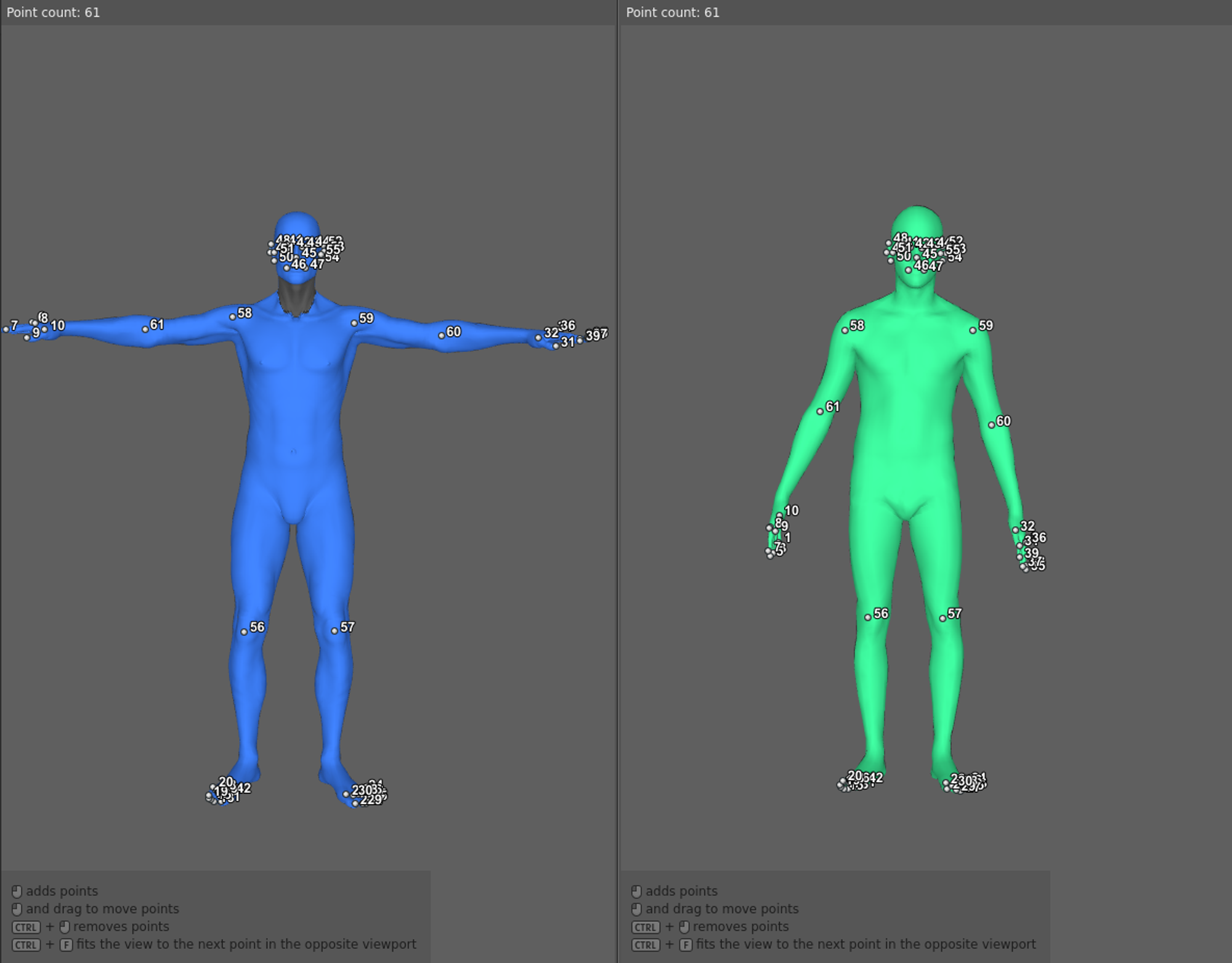}} & 
    {\includegraphics[width=0.4\textwidth]{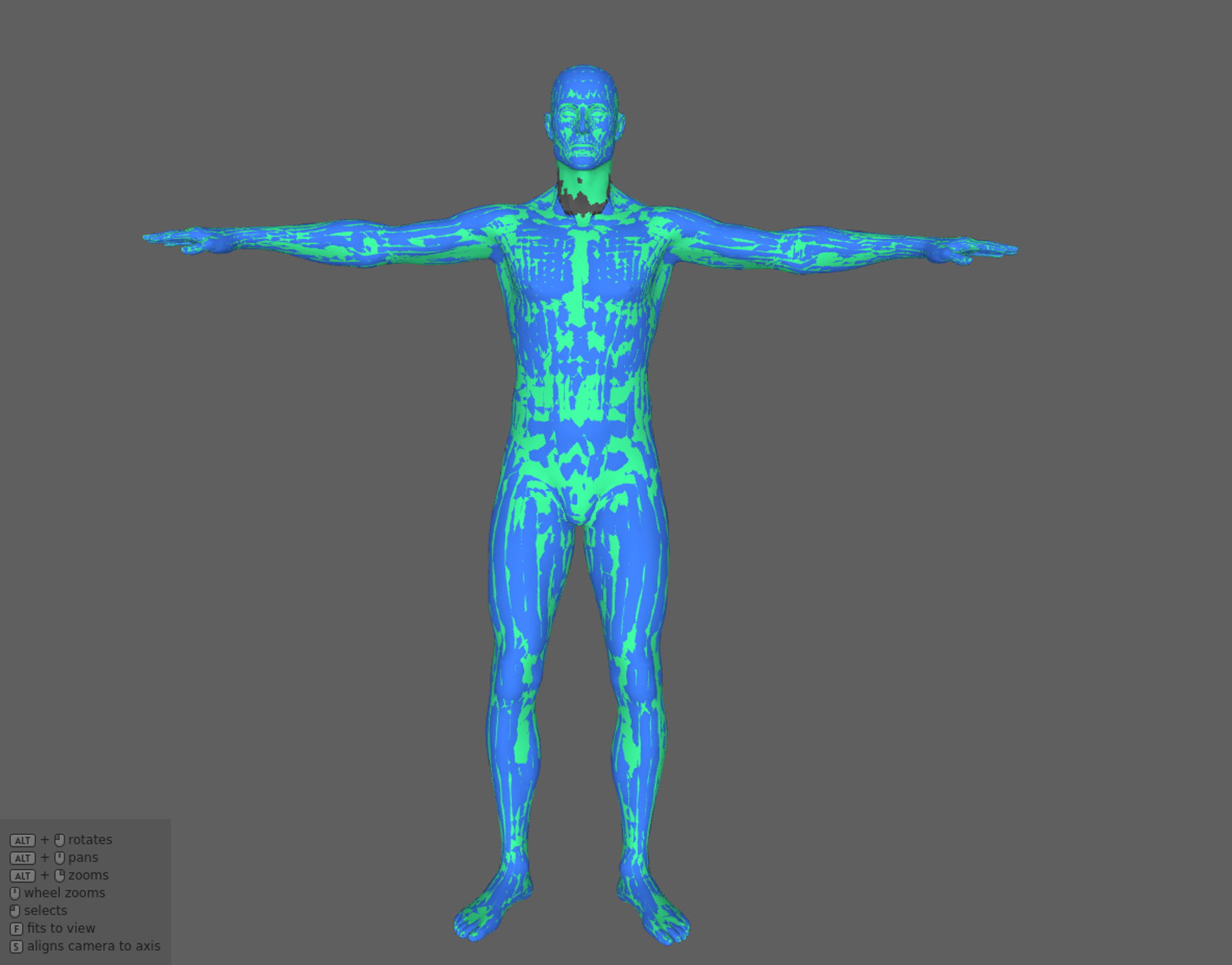}} \\
    (a) & (b) \\ 

    {\includegraphics[width=0.4\textwidth]{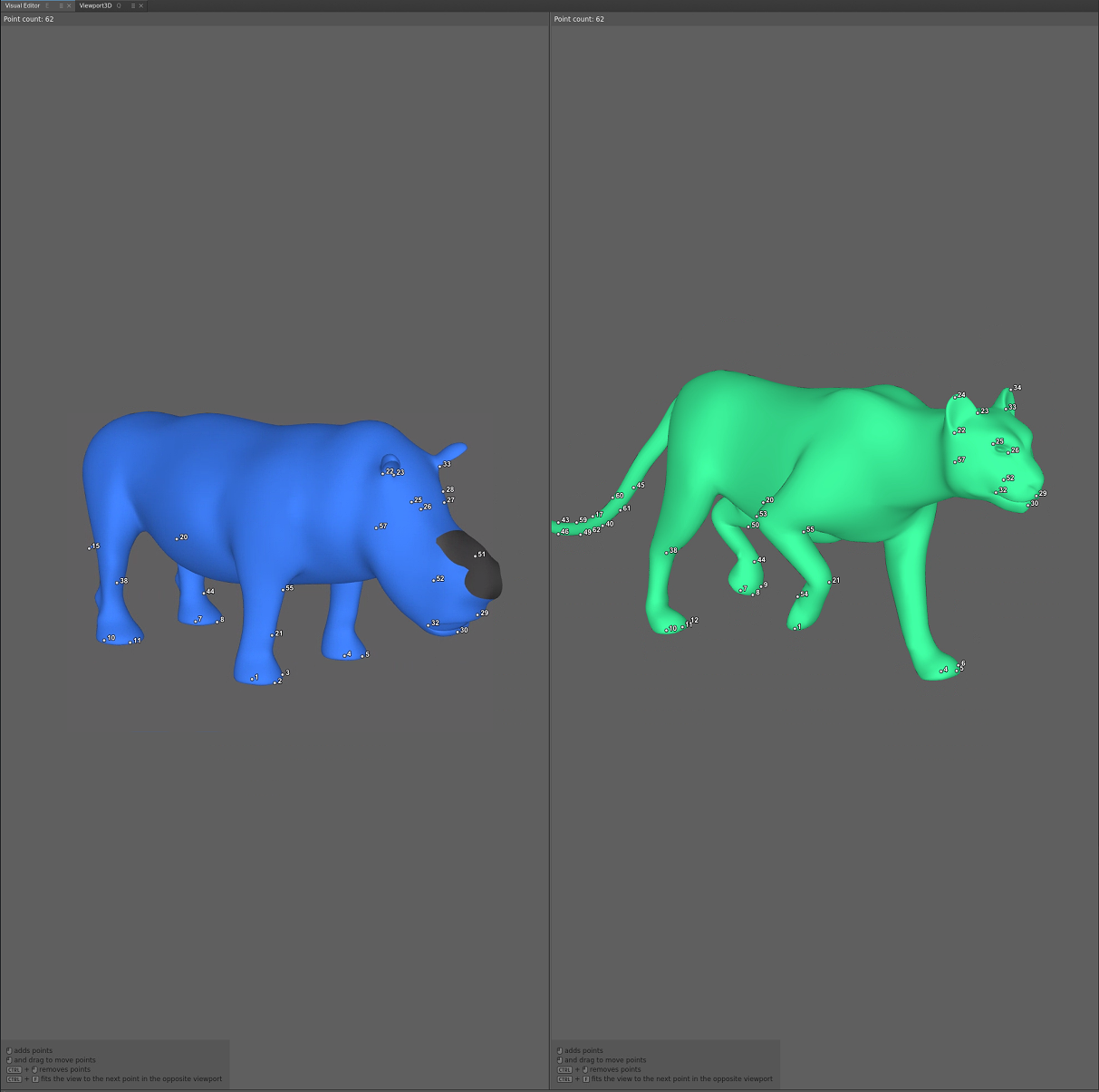}} & 
    {\includegraphics[width=0.4\textwidth]{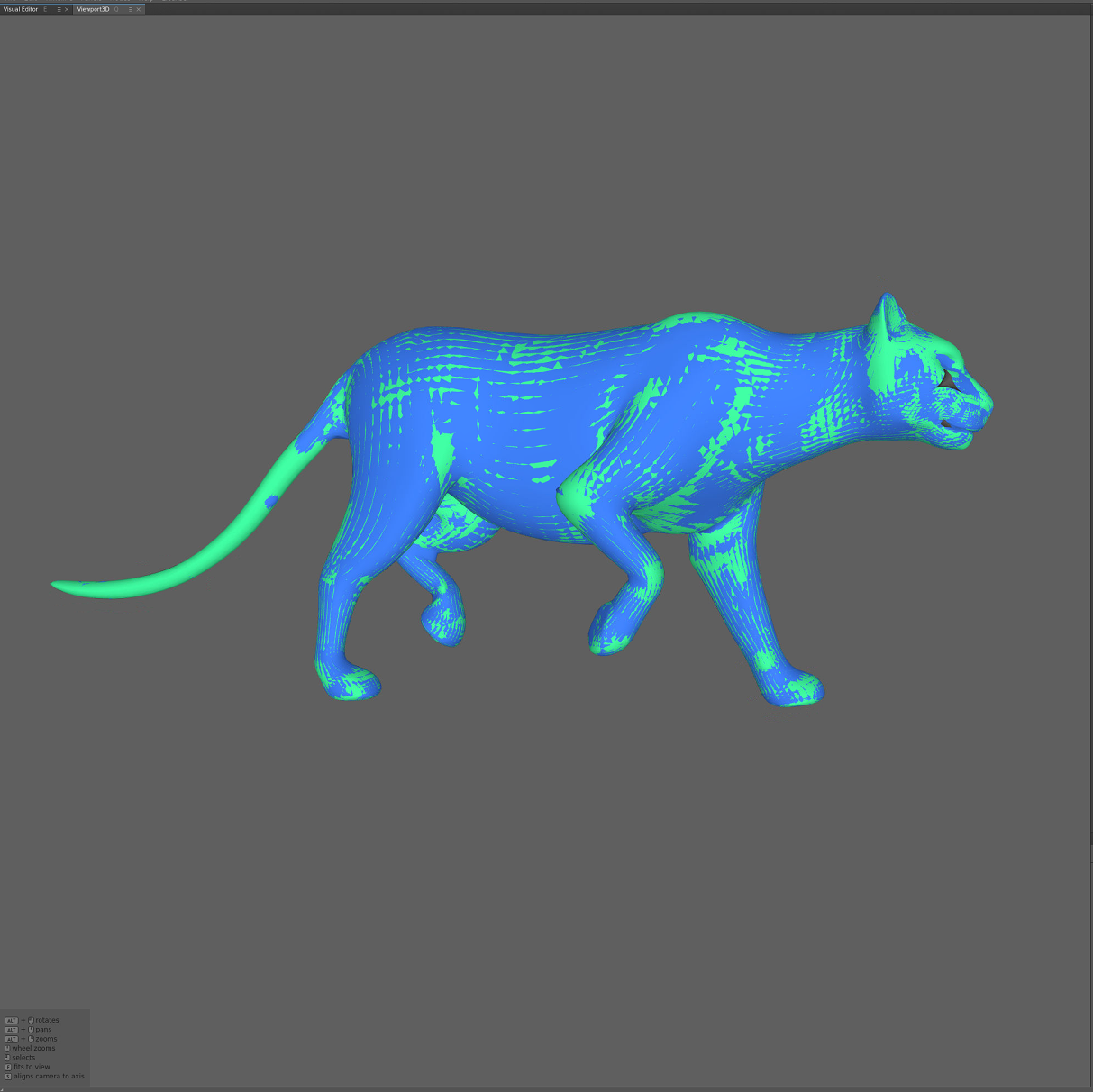}} \\
    (c) & (d) \\ 
    \end{tabular}

    \begin{tabular}{@{}c@{}}
        {\includegraphics[width=0.4\textwidth]{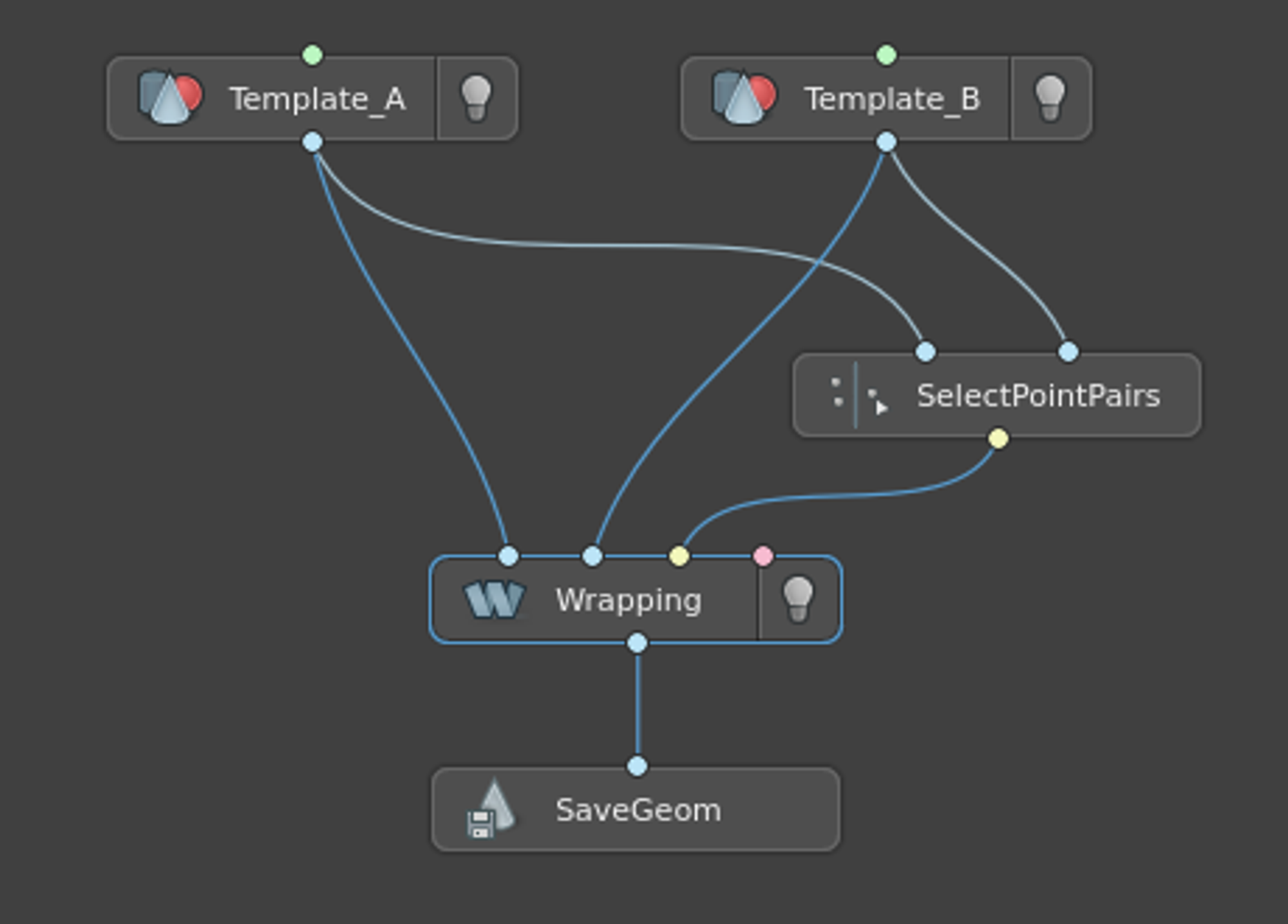}} \\
        (e) 
    \end{tabular}
\caption{\textbf{Usage of FaceForm} to generate the cross-category and cross-dataset ground-truth correspondences. \textbf{(a)} Given two templates, we manually place corresponding landmarks on both templates. \textbf{(b)} Overlap of the \textbf{template~A} (blue) and the \textbf{morphed template B} (green) on top of each other after running the "Wrapping" module. \textbf{(c)} and \textbf{(d)} We show a non-isometric deformation between a puma and rhino shape. Note that the rhino is missing its horn which is considered a missing semantic part and we only match the rhino shape without its horn to the puma. \textbf{(e)} The modules used to perform the morphing of \textbf{template~B} to \textbf{template~A}.}
\label{fig:face_form}
\end{figure*}

\begin{figure*}[tbh!]
    \centering
        \centering
    \begin{tabular}{@{}cc@{}}
        {\includegraphics[width=0.45\textwidth]{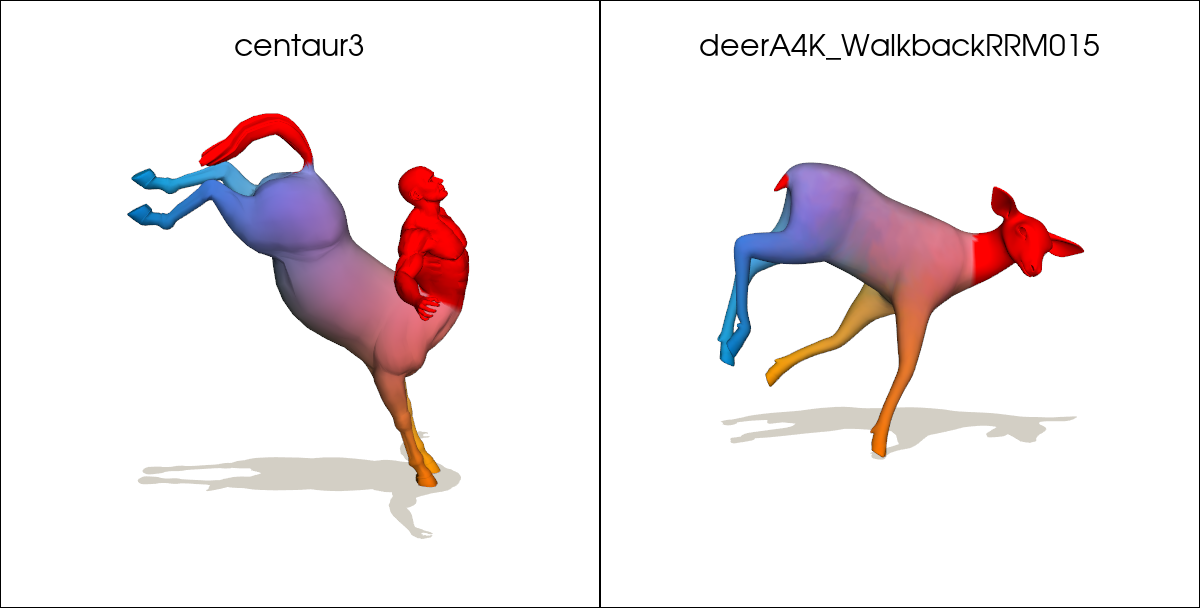}} & 
        {\includegraphics[width=0.45\textwidth]{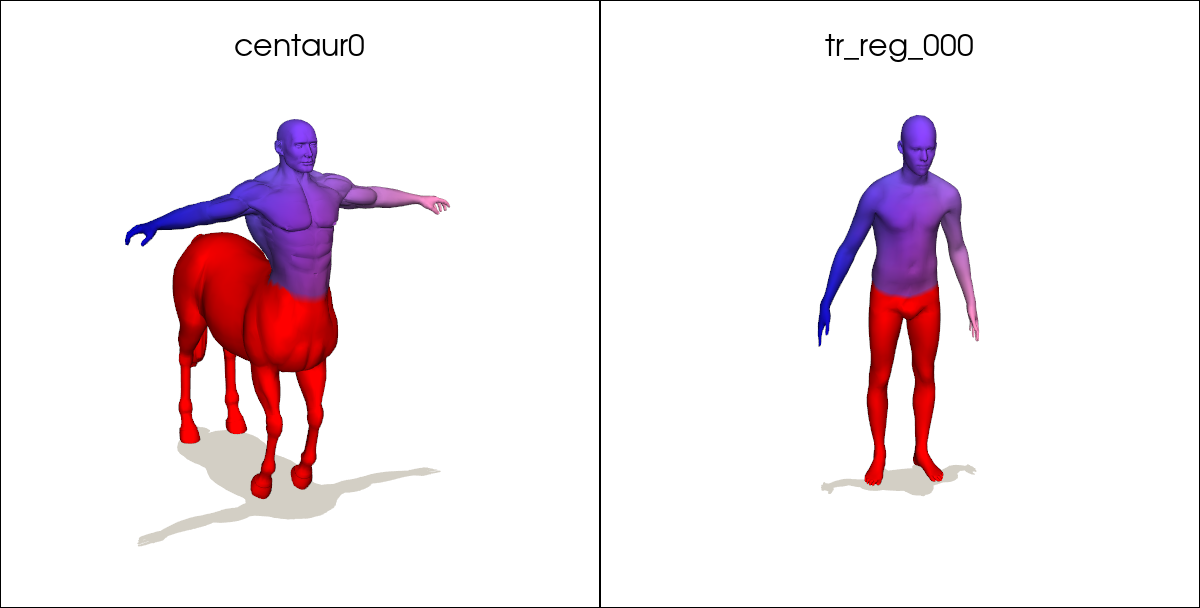}} \\
        (a) & (b) \\
        & \\

        {\includegraphics[width=0.45\textwidth]{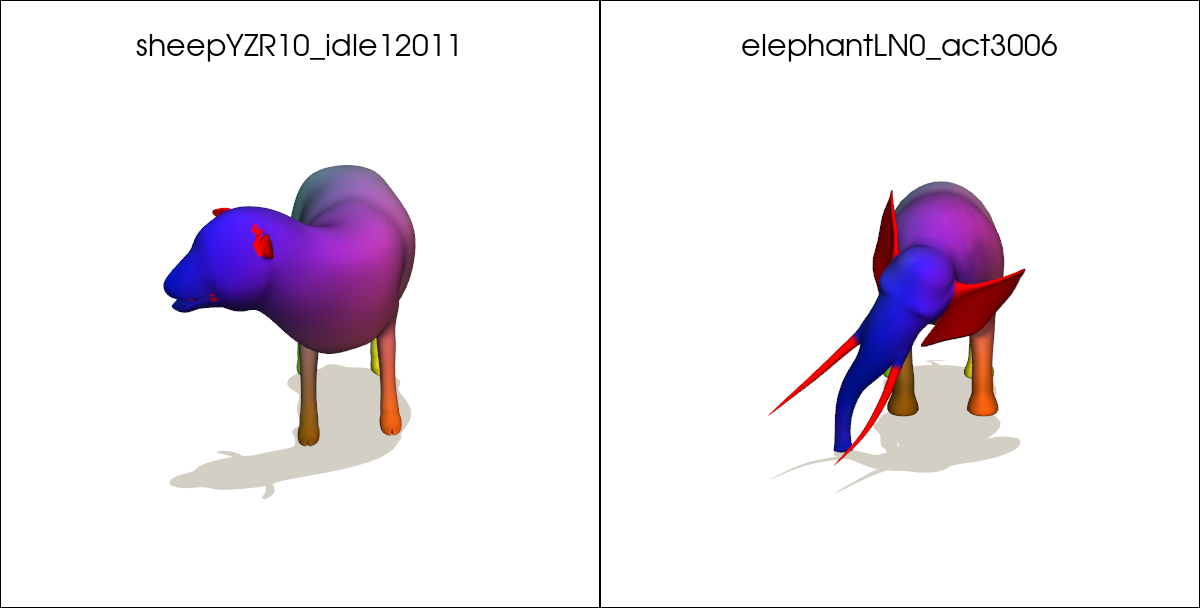}} & 
        {\includegraphics[width=0.45\textwidth]{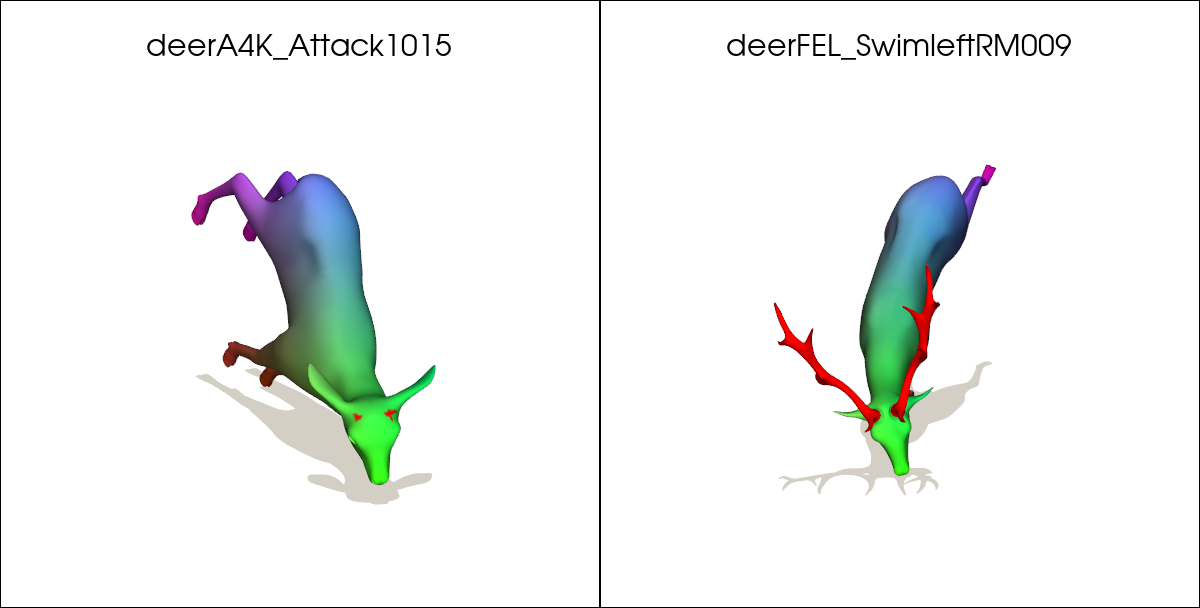}} \\
        (c) & (d) \\
        & \\

        {\includegraphics[width=0.45\textwidth]{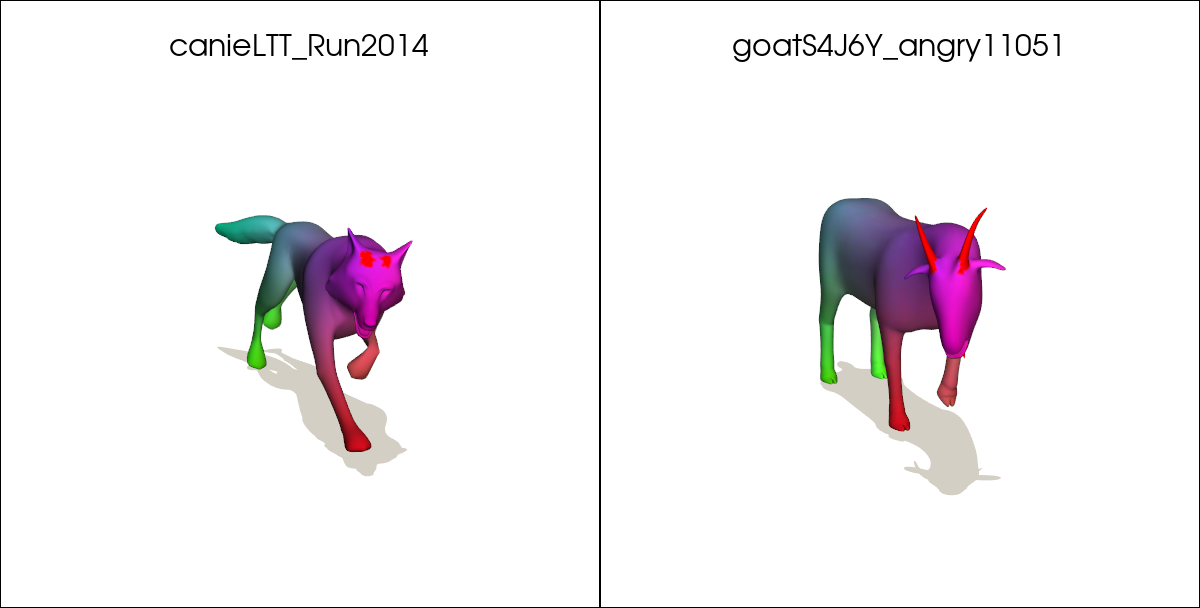}} & 
        {\includegraphics[width=0.45\textwidth]{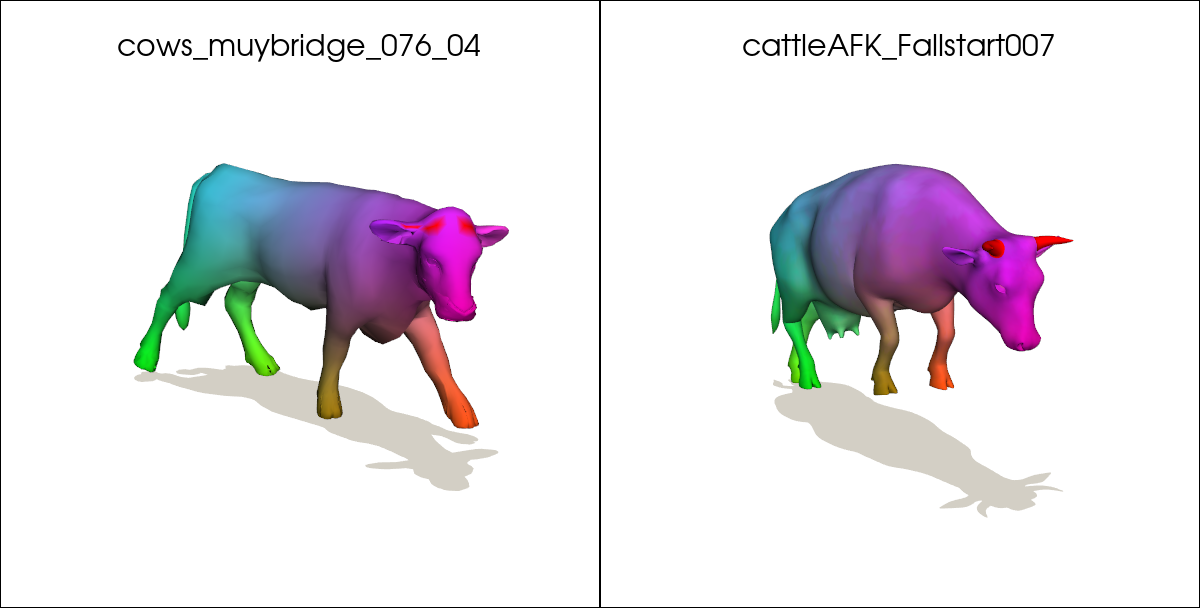}} \\
        (e) & (f) \\
        & \\

        {\includegraphics[width=0.45\textwidth]{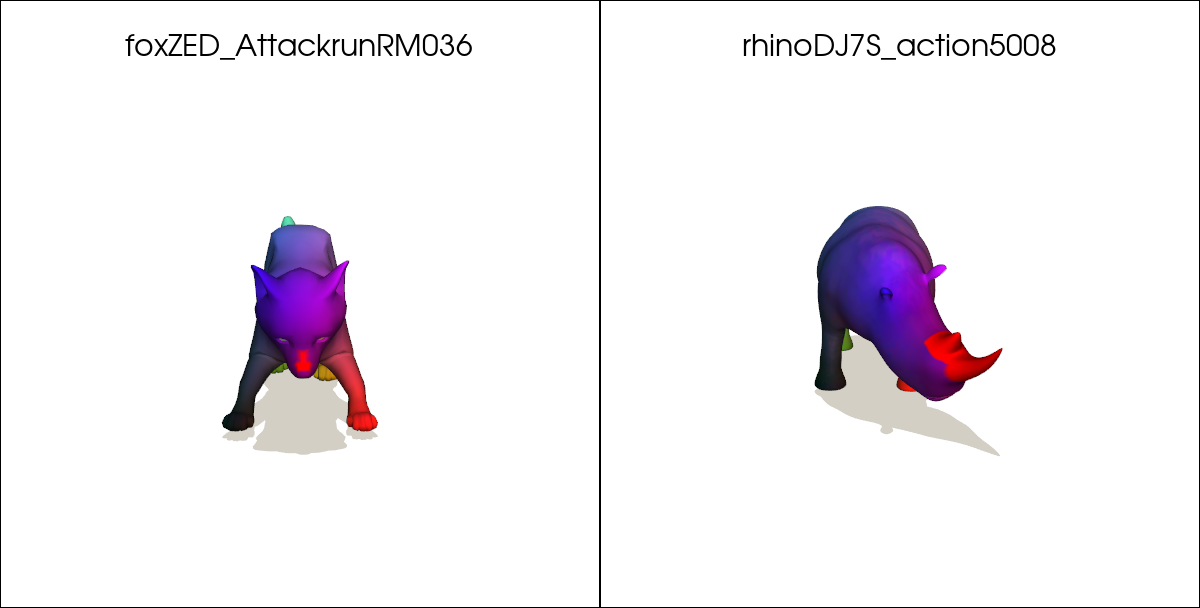}} & 
        {\includegraphics[width=0.45\textwidth]{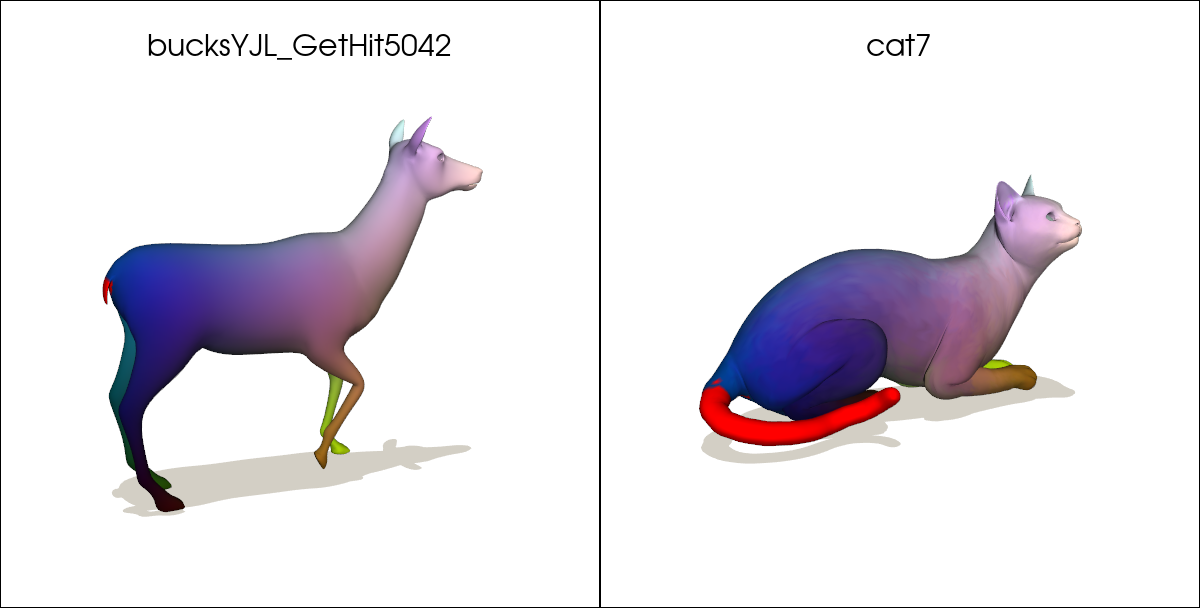}} \\
        (g) & (h) \\
        & \\

    \end{tabular}
    \caption{\textbf{Correspondence between pairs with semantic missing parts.} The colour is transferred from the left shape to the right shape. The red parts are the parts that cannot be semantically matched between the two shapes.
    \textbf{(a)} and \textbf{(b)} The centaur shape is a special case that can be matched to both humanoid and four-legged animals. \textbf{(c)} The elephant's tusks and ears cannot be matched to a sheep.  \textbf{(d), (e)} and \textbf{(f)} The horns are not always matched between shapes.  \textbf{(g)} The rhino's horn is not semantically matched to any other animal in \bm{}. \textbf{(h)} The cat's long tail is not matched to the buck's small fluffy tail.}
\label{fig:corres_vis_unmatchable}
\end{figure*}

This procedure is similar to the one used by \cite{magnet2022smooth}, and we use the FaceForm software with an academic license \cite{faceform}. The steps of our procedure can be summarised as follows: 
\begin{enumerate}
    \item We select a template for each dataset with cross-category correspondence and a template for each category in each dataset that does not have cross-category correspondences. The template should ideally be in a neutral A-or T-pose for humanoid shapes and in an "upright" standing position for four-legged animals. 
    \item We build a shape network connecting all templates in such a way that the semantic differences between two connected shapes are small. This makes the morphing step easier. For example, we try to connect all cats from all datasets together (see the interactive graph in our {\color[HTML]{0065bd} \href{https://nafieamrani.github.io/BeCoS/}{project page}}).
    \item For each edge in the shape network, we load the two connected templates $A$ and $B$ in FaceForm using the "LoadGeom" module (see Fig.~\ref{fig:face_form} \textbf{(c)}).
    \item Some template shapes do not completely match the other template as some parts are missing~(see Fig.~\ref{fig:corres_vis_unmatchable}). 
    For these shapes, we first cut the shapes so that we get a matchable and a semantic missing part.
    For these template shapes, we only load the matchable part in FaceForm.
    \item We connect the two modules to a "SelectPointPairs" module that allows manually selecting corresponding landmarks on the templates $A$ and $B$ (see Fig.~\ref{fig:face_form} \textbf{(a)}). Depending on the semantic difference between the two template shapes, we select between 15 and 95 landmarks. 
    \item Once the landmarks are placed manually on both shapes, we use the "Wrapping" module on FaceForm to morph template $A$ to template $B$. This process took around 25 hours of manual work. 
    \item We then save the morphed template $\hat{A}$ (see Fig.~\ref{fig:face_form} \textbf{(b)}).
    \item To get the correspondences between the two shapes, we find for every vertex $\hat{A}$ the closest triangle of template shape $B$.
    We then project every vertex of $\hat{A}$ on the closest triangle and return the barycentric coordinates. The correspondences of semantic missing parts are set to~$-1$.
    \item Once steps [3-7] are done for all edges in the tree, we are able to generate the correspondences between any two input shapes. 
\end{enumerate}

\subsection{Correspondence Computation between a Shape Pair}
Given any two shapes in the dataset, we compute the ground-truth correspondences between them using the following steps:
First, we extract the path between those two shapes over the given template shapes.
In the interactive shape network on our {\color[HTML]{0065bd} \href{https://becos-authors.github.io/BeCoS/embedded_page.html}{project page}}, we show the chosen templates and their inter-connectivity. 
We then extract the correspondences between any two shapes in the shape network either through the given datasets or by using our manually annotated correspondences. 
We then propagate the correspondences along the path. 

\subsection{Overlap Region for the P2P Setting}
\label{sec:p2p_overlapping_region}
For the overlap computation, we rigidly align the shapes $\mathcal{X}$ and $\mathcal{Y}$ using the given correspondences to ensure that both partial shapes are semantically similar and perform the in the main paper described partiality generation on both shapes.
Subsequently, we quantify the overlap as the percentage of vertices from $\hat{\mathcal{X}}$ that are mapped to $\hat{\mathcal{Y}}$ and vice-versa ($\hat{\mathcal{X}}$ and $\hat{\mathcal{Y}}$ are the partial shape generated using ray casting from shapes $\mathcal{X}$ and $\mathcal{Y}$, respectively).
If both overlaps do not fall within a predefined range, the process iterates with new random camera poses until the overlap meets the specified criteria, or until a maximum of $m$ iterations is reached. 
In the latter case, the partial shapes whose overlap is the closest to the desired range are selected. 

In the \bm{} default configuration for the partial-to-partial setting, shapes have an overlap between 10\% and 90\%. 
We show a histogram of the distribution of the overlapping region of each shape pair going from shape $\mathcal{X}$ to shape $\mathcal{Y}$ and vice-versa (see Fig.~\ref{fig:p2p_overlapping}).

\begin{figure}[!ht]%
    \centering
    \input{vis/tikz_vis/histogram_overlap}
\caption{Partial-to-Partial \textbf{histogram of overlapping region} of \bm{}. The pairs that are below 10\% and the ones above 90\% are the ones that reached the maximum number of iterations $m=10$. Our framework produces partial-to-partial shapes, creating this histogram when the predefined parameters discussed in Sec.~\ref{sec:benchmark} of the main paper are used.}
\label{fig:p2p_overlapping}
\end{figure}

%% file: vis/tikz_vis/histogram_overlap.tex
\newcommand{\pckLineWidth}{3pt}
\newcommand{\plotWidth}{1*\linewidth}
\newcommand{\plotHeight}{0.6*\linewidth}
\newcommand{\pckTitle}{\textsc{Histogram of Overlapping Regions - P2P}}

\pgfplotsset{%
    label style = {font=\footnotesize},
    tick label style = {font=\footnotesize},
    title style =  {font=\small},
}
\pgfkeys{/pgf/number format/.cd,1000 sep={}}

\begin{tikzpicture}[scale=1.0, transform shape]
\begin{axis}[
    width=\plotWidth,
    height=\plotHeight,
    grid=major,
    title=\pckTitle,
    ymin=0, ymax=4000,
    xmin=-5, xmax=105, %
    xtick={0, 10, 20, 30, 40, 50, 60, 70, 80, 90, 100}, %
    area style,
    ylabel={{\footnotesize Number of Shapes}},
    xlabel=\footnotesize Percentage of Overlapping Region (\%),
    ylabel near ticks,
    nodes near coords,
    every node near coord/.append style={font=\footnotesize, anchor=south, text=black}, %
]
\addplot+[
    ybar,
    mark=yes,
    bar width=7,
    draw opacity=3, 
] plot coordinates { (5, 20) (15, 1850) (25, 2391) (35, 2321) (45, 2828) (55, 3163) (65, 3391) (75, 3247) (85, 1715) (95, 2)};
\end{axis}
\end{tikzpicture}

%% file: supplementary/sec/details_benchmark.tex
\section{Details about the \bm{} Benchmark and the Framework}
\label{sec:benchmark_details}

As mentioned in the main paper, we use seven different full datasets in our benchmark. 
We use each dataset entirely with two exceptions: 
For TOSCA we exclude the gorilla as it is excluded from most shape matching analyses due to its difficulty. 
The original dataset~\cite{li20214dcomplete} of DT4D provides animated meshes for each category with a lot of categories.
Some categories in the original dataset are difficult to semantically match to other categories, e.g.~dragons.
Therefore, we use the same categories that were selected from DT4D-Matching~\cite{magnet2022smooth}.
This dataset provides its own cross-dataset correspondences but only for a version of the dataset where each mesh is around 8000 vertices.
Since the shapes are only re-meshed and not in the original resolution, we use the original animated files but extract the same categories and frames from the animation as DT4D-Matching.
This ensures that we have higher resolution input shapes and less error propagation along the correspondence paths, as the shapes in the same category have the same triangulation.
We end up with 39 categories for four-legged animals and eight categories of humanoid shapes. 

%% file: supplementary/sec/further_distribution.tex
\section{Hosting, Licences, Maintenance Plan \& Long-Term Preservation of \bm{}}
\label{sec:further_distribution_information}

Our pipeline is based on existing 3D shape datasets. 
The original datasets have to be downloaded and can only be used with the corresponding licences from the original datasets.
We ensure this correct usage by letting the user download the data directly from the corresponding websites. 
Our code for generating the (partial) shape pairs will be available on GitHub with a \href{https://creativecommons.org/licenses/by-nc-sa/4.0/}{CC BY-NC-SA 4.0} (open source for non-commercial use) licence.

Our main product is a publicly accessible framework, which we intend to permanently host on our GitHub repository.
Since the \bm{} benchmark is a specific instantiation of this universal framework, its accessibility depends on the availability of certain 3D shape datasets.
The biggest risk regarding accessibility is therefore related to the \bm{} benchmark in case some of the involved 3D shape datasets become unavailable.
However, given that the 3D shape datasets used in our framework are well-established, we anticipate their continued public availability.
In the event that a dataset is taken down by the originators, our framework's functionality will remain intact for the remaining datasets. %

\section{Compute and Resources}
\label{sec:compute_info}
We use an internal cluster to compute the given results with the different models. 
For SM-COMB and GC-PPSM, we use an Intel Xeon E5-2697 with 16 cores and up to 36GB of RAM per run. 
For the learning based methods, we use an Intel Xeon E5-2697 with 16 cores and one Nvidia Titan GPU with 12G VRAM. 
As PFM's precompiled code runs only on Windows, we use an Intel i7-14700KF with 20 cores and up to 32GB of RAM. 
We present the compute time per method used in our paper in Table.~\ref{tab:method_compute_time}.

\input{supplementary/tab/tab_compute_time}

\section{All Framework Options}
In Table~\ref{tab:general_options}, we provide a full list of the general options available in our framework.
We invite the reader to check our GitHub repository for further information.
\input{supplementary/tab/configurations}

%% file: supplementary/tab/tab_compute_time.tex
\begin{table}[tbh!]
\centering
\footnotesize
\begin{tabular}{lcc}
\toprule
\multirow{2}{*}{\textbf{Method}}     & \textbf{Training} & \textbf{Evaluation time}  \\
  &  \textbf{time} & \textbf{(per shape pair)}  \\
\midrule
DPFM~\cite{attaiki2021dpfm} &8-20h & 1.1s\\
SM-COMB~\cite{roetzer2022scalable} & N/A & 0.08h \\
GC-PPSM~\cite{ehm2024partial} & N/A & 2.5h \\
PFM~\cite{rodola2017partial} & N/A & 0.2h\\ 
ULRSSM~\cite{cao2023unsupervised} &16-86h & 20s\\ 
Smooth Shells~\cite{eisenberger2020} & N/A & 0.2h\\ 
GeomFMaps~\cite{donati2020deepGeoMaps} &3-7h & 1.1s\\ 
\bottomrule
\end{tabular}
\caption{A breakdown of the \textbf{compute time} per method presented as a baseline in our work. Axiomatic methods do not require any training time. We show the training time for the entire training data and the evaluation time per shape pair.}
\label{tab:method_compute_time}
\end{table}

%% file: supplementary/tab/configurations.tex
\definecolor{light-gray}{gray}{0.95}
\newcommand{\code}[1]{\colorbox{light-gray}{\texttt{#1}}}
\begin{table*}[tbh!]
\caption{\textbf{All options in the main configuration file.} The default values of these options are stored on our framework code and allow the generation of the \bm{} benchmark.} 
\label{tab:general_options}
\centering

    \begin{tabular}{@{}l|l@{}}
        \toprule
        \textbf{Option}                       & \textbf{Description}                                        \\
        \midrule
        \code{data\_dir}                      & Path to data directory.                                     \\
        \midrule
        \code{datasets}                       & True if the dataset is used.                                \\
        \midrule
        \code{combinations}                   & Type of shapes to generate, possible values:                \\
                                              & human, four-legged, human\_centaur,                         \\
                                              &  four-legged\_centaur, or all.                              \\

        \midrule
        \code{setting}                        & The setting to generate, possible values:                    \\
                                              & full\_full/partial\_full/partial\_partial.                  \\

        \midrule
        \code{remesh}                         & True if the shapes should be re-meshed after        \\
                                              & loading.                 \\
        \midrule
        \code{cam\_pos\_regime}               & Specifies how far the pair of camera poses are             \\
                                              & apart, possible values: high/medium/low.                    \\
        \midrule
        \code{store\_vis}                     & If True, store visualization of each pair        \\
                                              & of shapes.                     \\
        \midrule
        \code{show\_output}                   & If True, an open3d visualization is shown during            \\
                                              & generation.                                                 \\
        \midrule
        \code{original\_settings}             & If True, multiple assertions are activated to         \\
                                              & ensure that \bm{} is generated.                                    \\
        \midrule
        \code{use\_precompute\_remeshing}     & If True, use the cached re-meshed shapes                    \\
                                               & if available.                                              \\
        \midrule
        \code{update\_precomputed\_remeshed}  & If True, update the caching directory for             \\
                                              & re-meshed  meshes.                                                     \\
        \midrule
        \code{use\_precomputed\_partial\_raycasting} & If True, use the cached triangle IDs of each         \\
                                              &  partial shape if available.                                \\
        \midrule
        \code{update\_precomputed\_raycasting} & If True, update the caching directory for triangle        \\
                                               & IDs.                                                       \\
        \midrule
        \code{one\_axis\_rotation}            & If True, perform a random rotation around the               \\
                                              & z-axis after generating the pair of shapes.                        \\
        \midrule
        \code{n\_cam\_pos}                    & The number of camera poses to sample to get                 \\
                                              & within the desired range of overlap.                        \\
        \midrule
        \code{min\_overlap}                   & Minimum overalp between partial shapes desired.             \\
        \midrule
        \code{max\_overlap}                   & Maximum overalp between partial shapes desired.              \\
        \bottomrule
    \end{tabular}

\end{table*}